# EYE IMAGE-BASED ALGORITHMS TO ESTIMATE PERCENTAGE CLOSURE OF EYE AND SACCADIC RATIO FOR ALERTNESS DETECTION

*Supratim Gupta*

# EYE IMAGE-BASED ALGORITHMS TO ESTIMATE PERCENTAGE CLOSURE OF EYE AND SACCADIC RATIO FOR ALERTNESS DETECTION

*Thesis submitted to the*
*Indian Institute of Technology, Kharagpur*
*for award of the degree*
*of*

## Doctor of Philosophy

by

## Supratim Gupta

Under the guidance of

Dr. Aurobinda Routray and Dr. Anirban Mukherjee

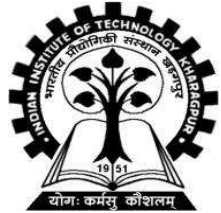

DEPARTMENT OF ELECTRICAL ENGINEERING
INDIAN INSTITUTE OF TECHNOLOGY, KHARAGPUR
WEST BENGAL - 721 302, INDIA
MAY 2013

Uncertainties around us set a quest for the truth,
Uncertainties around steady light stimulate the vision.
Research through uncertainties is a journey from belief to realization.

*Dedicated to*

*my parents, wife and siblings*

# Certificate

This is to certify that the thesis entitled "**Eye Image-Based Algorithms To Estimate Percentage Closure Of Eye And Saccadic Ratio For Alertness Detection**" submitted by **Sri Supratim Gupta** to Indian Institute of Technology, Kharagpur, is a record of bona fide research work carried under our supervision and we consider it worthy of consideration for the award of the degree of Doctor of Philosophy of the Institute.

| | |
|---|---|
| Dr. Anirban Mukherjee | Dr. Aurobinda Routray |
| Department of Electrical Engineering, | Department of Electrical Engineering, |
| Indian Institute of Technology, | Indian Institute of Technology, |
| Kharagpur -721 302, India. | Kharagpur -721 302, India. |

# Declaration

I certify that

a. The work contained in this thesis is original and has been done by me under the guidance of my supervisor(s).
b. The work has not been submitted to any other Institute for any degree or diploma.
c. I have followed the guidelines provided by the Institute in preparing the thesis.
d. I have conformed to the norms and guidelines given in the Ethical Code of Conduct of the Institute.
e. Whenever I have used materials (data, theoretical analysis, figures, and text) from other sources, I have given due credit to them by citing them in the text of the thesis and giving their details in the references. Further, I have taken permission from the copyright owners of the sources, whenever necessary.

Supratim Gupta

# Acknowledgments

It gives me immense pleasure to convey my deep sense of gratitude to my supervisor Dr. Aurobinda Routray for his guidance, support and for giving freedom to work independently throughout my research work. I am also grateful to my co-supervisor Dr. Anirban Mukherjee for his continuous support and encouragement.

It is a privilege to thank Prof. J. A. Stern, Washington University, DC USA, for his valuable suggestion and support that drew my attention on saccadic eye movement for alertness detection. I am also thankful to Dr. D.F. Dinges, USA, Dr. D. Tweed, Canada, Dr. T. Haslewanter, Austria, and Dr. K. R. Rao, Arlington, Texus for providing their valuable research articles. I also offer my sincere thanks to Dr. S. Sen, Dr. P. K. Dutta, Dr. C. S. Kumar, Dr. A. K. Sinha, Dr. J. Pal, Dr. A. K. Pradhan, and Dr. N. Chakraborty for giving constructive suggestions and encouragement.

I would like to thank Department of Information Technology, New Delhi, Indian Institute of Technology, Kharagpur and CRRI, New Delhi for providing financial support and various resources to carry out my research work.

Nothing would have been possible without the support of my friends specially A. Mitra, H. Mishra, M. Bhagat, Vignesh R., S. Ghorai, T. Pradhan, A. Dasgupta, A. Samanta, A. Mukherjee, A. Rajaguru, M. Midya, S. Kar, S. L. Happy, the subjects in the experiments and other team members. I am thankful to Dr. B. K. P. Nayak for helping me in validating the results of PERCLOS. Mr. M. Mukhi, Mr. A. Nandi, Mr. P. Das, Mr. S. K. Kundu are always available to provide all possible help at right time.

It would have been impossible to walk through this course without the endless support of my beloved wife, Mrs. Shakuntala Gupta for designing and collecting the experimental data and for keeping patience whenever I had lost hope. I express my gratitude to my parents- Mr. Nilotpal Gupta & Mrs. Dhira Gupta, my siblings and other family members who supported me throughout the work.

Finally, I am grateful to GOD and my Gurus whose blessings help me to shape the research work into reality.

<div align="right">Supratim Gupta</div>

# Abstract

*In the area of vehicular safety much attention has been drawn towards detection of the level of alertness in drivers (operators) and subsequent prevention of road casualties. Literature has suggested that the temporal pattern of ocular movements may provide prominent indication of alertness level. The eye images captured using a camera, can be analyzed to find the pattern of eyelid and pupil (iris) motion to determine the vigilance level. The eyelid motion can be measured by the index called Percentage Closure of Eyes (PERCLOS). It is the proportion of time within a predefined time period for which the eyelid is fully or partially closed and is related to the level of drowsiness. This index is increased with the diminished level of alertness. On the other hand, the iris motion can be characterized by Saccadic Ratio (SR). Saccade is a quick and simultaneous movement of iris between two points of fixation and the SR is the ratio of saccadic peak velocity to the standard saccadic duration. This ratio degrades during the lower states of alertness. It is found that in-vehicle measurement of PERCLOS and SR may pose several challenges. Among these, the variation of illumination, relative motion between camera and the driver's face due to free movement are the prevalent factors. Therefore an algorithm that determines the indices in presence of these constraints is essential. The current research work has developed two novel algorithms for image-based measurement of PERCLOS and SR. The PERCLOS is estimated by correlation filter-based technique. An innovative combination of gray scale and Near Infrared (NIR) sensitive camera with passive NIR illuminator helps to achieve higher accuracy than the existing art through day and night. The estimation of SR needs detection of iris and two eye corner positions. Two novel techniques have been developed for the detection of iris center and eye corners. A novel index called Form Factor (FF) has been introduced to find the pupil position and a local form factor-based method for edge detection has been developed. The saccadic velocity profile can be estimated online from the temporal information of the iris positions using standard tracking algorithm such as Kalman or Extended Kalman filter. Experimental results indicate that the estimation of both SR and PERCLOS can predict the operator's level of alertness from onset of diminished alertness to fatigue.*

# Contents

# List of Symbols

| | |
|---|---|
| $I$ | 2-D Image |
| $N$ | Number of pixels in an image or subimage |
| $\alpha$ | Brightness distortion factor |
| $\lambda$ | Color distortion factor |
| $N_{lh}$ | Number of LBP histogram |
| $\mathbf{h}$ | LBP histogram |
| $N_h$ | Number of histogram bins |
| $P(.)$ | Probability distribution function |
| $N_{cd}$ | Dimension of color feature space |
| $Th$ | Threshold value |
| $L_{cb}$ | Length of codebook |
| $N_s$ | Number of sample values |
| $N_d$ | Dimension of feature space |
| $N_{fe}$ | Number of features |
| $W_I$ | Width of an image |
| $H_I$ | Height of an image |
| $W_B$ | Width of base window |
| $H_B$ | Height of base window |
| $n_B$ | Number of pixels in base window |
| $tpr$ | True positive rate |
| $fpr$ | False positive rate |
| $(.)^T$ | Transpose of a vector or matrix |
| $tp$ | True positive |
| $fp$ | False positive |
| $B_{LBP}$ | Number of bits representing LBP |
| $t_c$ | Constant computation time |
| $k_{mh}$ | Number of model histogram |
| $P3$ | PERCLOS computed as 3 min moving average |
| $\mathbf{\Gamma}$ | Lexicographically order image vector |
| $\mathbf{\Psi}$ | Mean of image vector |
| $\mathbf{\Phi}$ | Mean subtracted image vector |
| $\mathbf{C}$ | Covariance matrix |
| $K_E$ | Number of principal components |
| $a, b, c, d$ | Non-negative optimal trade-off parameters |
| $\mathbf{h}_f$ | Correlation filter vector |
| $\mathbf{C}_n$ | Noise power spectral density |
| $\sigma$ | Standard deviation |
| $\mu$ | Mean |
| $\mathbf{D}_x$ | Diagonal average power spectral density |
| $\mathbf{S}_x$ | Similarity matrix |
| $E(.)$ | expectation operation |
| $G(i,j)$ | 2-D Correlation surface |
| $\odot$ | Circular correlation |
| $\otimes$ | Circular convolution |
| $\mathbf{ln}$ | Natural logarithm |
| $H$ | Entropy |
| $H_{JE}$ | Joint entropy |
| $H_{ME}$ | Marginal entropy |
| $\hat{I}$ | Symmetrically extended image |
| $\hat{H}_f$ | Symmetrically extended 2-D correlation filter |
| $\rho$ | Correlation coefficient |
| $F$ | Form Factor |
| $F_s$ | Form Factor of noise free signal |
| $F_g$ | Form Factor of noisy signal |
| $F_r$ | Frame rate of capturing image |
| $\Delta$ | Small change |
| $\alpha$ | Edge strength index |
| $\alpha_s$ | Edge strength index of noise free signal |
| $\alpha_g$ | Edge strength index of noisy signal |
| $\Delta_w^p(.)$ | Baddeley error metric |
| $\eta(k)$ | Uncertainty in measurement at $k^{th}$ instant |
| $\zeta(k)$ | Uncertainty in state prediction at $k^{th}$ instant |
| $\mathbf{K}(k)$ | Kalman gain at $k^{th}$ instant |

**J**     Jacobian matrix

# List of Abbreviations

| | |
|---|---|
| ACE | Average Correlation Energy |
| ACH | Average Correlation Height |
| ACS | Autocorrelation Sequence |
| APE | Absolute Percentage Error |
| ASM | Average Similarity Measure |
| AUC | Area Under the Curve |
| AVT | Auditory Vigilance Test |
| BEM | Baddeley Error Metric |
| BSDS | Berkley Segmentation Data Set |
| BUN | Blood Urea Nitrogen |
| CB | Code Book model |
| CCD | Charge Coupled Device |
| CHT | Circular Hough Transform |
| COM | Center of Mass |
| CRRI | Central Road Research Institute |
| DCT | Discrete Cosine Transform |
| DFF | Directional Form Factor |
| DFT | Discrete Fourier Transform |
| ECG | Electrocardiogram |
| EEG | Electroencephalogram |
| EKF | Extended Kalman Filter |
| EMG | Electromyogram |
| EOG | Electrooculogram |
| ESD | Eigen Space Decomposition |
| ESI | Edge Strength Index |
| FF | Form Factor |
| FIR | Finite Impulse Response |
| FOM | Figure of Merit |
| FPR | False Positive Rate |
| FR | Fisher Ratio |
| GUI | Graphical User Interface |
| HCL | Haar-like feature based Classifier |
| HIL | High Illumination Level |
| HLF | Haar-Like Features |
| i.i.d. | Independently and Identically Distributed |
| IITKGP | Indian Institute of Technology, Kharagpur |
| IQM | Image Quality Metric |
| IR | Infrared |
| KDE | Kernel Density Estimation |
| KF | Kalman Filter |
| LBP | Local Binary Pattern |
| LED | Light Emitting Diode |
| LFF | Local Form Factor |
| LIL | Low Illumination Level |
| LS | Local Statistics |
| MACE | Minimum Average Correlation Energy |
| MC | Multi-Colour space |
| MDL | Minimum Description Length |
| MI | Mutual Information |
| MNRL | Maximum Negative Run-Length |
| MSE | Mean Square Error |
| NIR | Near Infra-Red |
| NMS | Non-Minimum Suppression |
| NTVR | Noise-to-Total Variance Ratio |
| ONV | Output Noise Variance |
| OT-MACH | Optimal Trade-off Maximum Average Correlation Height |
| PC | Phase Congruency |
| PCA | Principal Component Analysis |
| PCE | Peak Correlation Energy |
| pdf | Probability Density Function |
| PERCLOS | Percentage Closure of Eyes |
| PSNR | Peak Signal-to-Noise Ratio |
| PSR | Peak-to-Sidelobe Ratio |
| PSV | Peak Saccadic Velocity |
| R.M.S. | Root Mean Square |
| RAM | Random Access Memory |
| RBS | Random Blood Sugar |
| ROC | Receiver's Operating Characteristics |
| SCD | Saccadic Duration |
| s.d. | Standard Deviation |
| SMG | Single Mode Gaussian model |
| SNR | Signal-to-Noise Ratio |
| SR | Saccadic Ratio |
| SSD | Subspace Decomposition |
| STVR | Signal-to-Total-Variance Ratio |
| SUSAN | Smallest Univalue Segment Assimilating Nucleus |
| SVM | Support Vector Machine |
| TEX | Texture |
| TPR | True Positive Rate |
| TV | Total Variance |
| TVNR | Total-Variance-to-Noise Ratio |
| VRT | Visual Response Test |
| WMV | Windows Media Video |
| XML | EXtensible Markup Language |

# Chapter 1

# Ocular Features and Human Alertness

## 1.1 Introduction

In the area of vehicular safety considerable research has been carried out to detect the state of alertness or drowsiness or fatigue in human drivers (operators) for prevention of road accidents. In the context of human subject the term fatigue has been synonymously used with drowsiness and/or diminished level of alertness. However, there are subtle distinctions among these states. The fatigue in human drivers occurs at extreme level of drowsiness leading to sleepiness while drowsiness itself is a higher level of diminished alertness.

A brief survey on the physical manifestation of reduced alertness has been presented to get possible indices of alertness. The state-of-the-art in measurement of such indices is described in the following sections. This also highlights the research issues and motivation for the development of non-invasive and robust algorithms to estimate human alertness through ocular parameters.

## 1.2 Causes and Manifestation of Diminished Alertness

The alertness level in human drivers can be affected by various intrinsic and extrinsic factors such as sleep deprivation, irregular working hours, prolonged driving etc. [1–6].

It is also dependent on environmental stimuli and psychological factors like anxiety, mood etc. [2]. Reduced level of alertness may be manifested as changes in the physiological states in human subject. Some of the major changes can be attributed to -

- Variations in the level of bio-chemicals such as creatinine, urea, Random Blood Sugar (RBS) etc.
- Variations in physiological signals such as Electroencephalogram (EEG), Electrooculogram (EOG), Electrocardiogram (ECG), Electromyogram (EMG).
- Variations in the voice response in oral interaction
- Temporal variations in facial expression, body movement, and movement of eyelid and pupil (or iris)

## 1.3 Measurement of Alertness Level

The changes in the above physiological parameters can be used to assess the level of alertness. The following literature review discusses the state-of-the-art of the research in assessing alertness from these signals. There are several methods based on different types of indices for measurement of alertness level.

### 1.3.1 Intrusive Measures of Alertness

**Self Assessment-** A self-assessment of the level of alertness by the drivers is often unreliable [7]. The drivers' may feel it at much reduced level of alertness when the performance of individual is severely affected. The subjective assessment of alertness level can be improved by a psychometric method based on a set of questions [8, 9]. The objective score obtained from the response to this questionnaire can be used to assess the alertness level.

**Psychomotor Response Test based Assessment-** The degradation of performance due to reduced level of alertness can also be determined from different psychomotor response tests such as action judgment test, speed anticipation reaction test, complex reaction time test [10]. Pang *et al.* [11] have considered Auditory Vigilance Test (AVT) and Visual Response Test (VRT) for detection of mental fatigue.

**Physiological Signal based Assessment-** Several drowsiness or fatigue detection technologies have been evolved for more accurate measurement of alertness. These are based on many physiological signals like EEG [2, 12–15], EOG [16–18], EMG [19–21], ECG [3]. It has also been reported that skin conductance can be used for assessing the vigilance level of human subject [12]. However, acquisition of these physiological signals needs physical contacts of electrodes with the driver introducing discomfort during driving. Moreover, a small displacement of the electrodes may produce error during analysis of these signals for detection of alertness.

**Steering Grip Pressure based Assessment-** A few investigations have been conducted to extract the state of alertness using steering grip pressure [22–24]. Here, the pressure sensor can be distributed either on the steering wheel or in the gloves to be worn during driving. However, in both the cases the wide inter personal variability in the temporal pattern of such signal exists due to several factors including drowsiness in human operator which may increase false alarm rate.

**Voice Response based Assessment-** Harrison *et al.* [25] have reported that the voice response to some verbal stimulus is affected by sleep deprivation which is one of the causes for induction of drowsiness in human subjects. These effects can be observed as slurred speech, repetitive speech, slow or mumbled speech, flattened voice etc. Several other researches have also shown that some speech related features such as pitch, duration of speech, intensity, glottal source properties and vocal tract spectrum are considerably affected by fatigue [26]. However, the method requires oral interaction with the subject which may alter the state of alertness.

**Blood Bio-chemical based Assessment-** Literature suggests that a few blood bio-chemical particularly the neurotransmitters and their metabolic products, blood glucose, serum urea, and creatinine can reliably reflect the diminished level of drowsiness [27]. Significant changes in glucose metabolism are observed with gradual increase in sleep deprivation. Methods based on these indices are invasive in nature and are reported to produce most accurate measurement of alertness.

All of the above methods of measurement of alertness need either physical contact or some sort of interaction with the human subject. This makes them unsuitable for use in driver monitoring during vehicle driving as well as other real-time operations.

### 1.3.2 Ocular Measure: A Non-intrusive Method for Alertness measurement

The EEG based or bio-chemical index based assessment of alertness can be considered to produce most accurate estimate of drowsiness [27, 28]. However, these methods are invasive resulting in discomfort to the operator. In some of the literatures [29–35] it has been found that the facial features (like eyes) that exhibit prominent indication of the state of alertness of human operator may be utilized for the purpose. Root cause analysis of variation in state of alertness in human subject reveals that most of the information intake takes place through eye during driving. Thus, reduction in alertness level is reflected critically on the eyes [33, 34]. Therefore, ocular features can be tracked and analyzed to estimate the level of alertness in human operator. Literatures have suggested several ocular features like eye blink frequency, PERCLOS, saccadic eye velocity etc. [32, 33]. The eye blink frequency might be poor indicator of alertness in human subjects as it may be affected by several factors apart from drowsiness [36]. Moreover, considerable inter personal variations in blink rate exists [37]. On the other hand, the temporal variation of eyelid position may indicate the state of drowsiness. A considerable research on such phenomenon has been carried out by measuring percentage closure of iris by eyelid over a predefined time interval [38–43]. Smith *et al.* [40] have used color predicates to determine the temporal variation of eyelid position over iris. This method may not be feasible during night drive and the method may be computationally expensive compared to gray scale or monochrome image processing. However, most of these methods except that in [40] are based on active Near Infra-Red (NIR) illumination which relies on the retinal reflectance characteristics [43]. In these methods the image frame is searched for the existence of iris only to measure PERCLOS. Research in [7, 44] suggest that these methods may fail due to inappropriate retinal reflection during day time, relative motion between subjects and the camera (due to their free movement). It has been observed that the level of alertness declines much earlier than the inception of drowsiness. Though PERCLOS may be a good indicator of drowsiness (a higher stage of diminished alertness) it fails to identify the inception of reduced alertness level [33].

The other ocular parameters such as temporal pattern of the pupil (or iris) move-ment and its variation in size may provide an indication of the onset of diminished alertness. Cognitive aspect of eye indicates that pupil dilates during information processing to capture more light. The dilation may be due to other factors like change in ambient light apart from reduced level of alertness [32]. However, sudden change in ambient light occurs only for small durations particularly at night. It has been reported that saccadic profile can be used to monitor the vigilance level of the driver [38, 45–50]. The saccadic peak velocity decreases with increase in sleep deprivation. The eye saccade occurs when the subject makes a sudden and quick jump from one point of interest to another. The peak saccadic velocity may reach about $600°/sec$. It has been found by the researchers that a typical saccade may last 20 ms and the refractory period of saccade is about 200 ms [51–53]. Thus from the literature it is evident that a saccadic eye movement can be simulated by a step change of small duration ($\leq 20$ ms). Moreover, the time interval between two consecutive saccades must be at least 200 ms. This indicates that a high speed ($\geq 60$ fps) recording of saccadic movement is essential. An image based measurement of SR has been developed in [54]. The saccadic ratio is defined as the ratio between peak saccadic velocity and standard saccadic duration. In [54] a high speed fixed camera projective geometry has been utilized. However, in the context of driving or any operations where operators move freely the fixed projective geometry of vision system may not be valid.

From the above literature review it is apparent that the state-of-the-art for detection of alertness has been utilizing PERCLOS as the only index. It can be noted that PERCLOS cannot used to measure the decline in alertness level at an early stage. Researches show that SR can also be used as another ocular index for vigilance level. However, measurement of SR when the operator has reached drowsiness may be difficult to perform due to partial occlusion by longer eyelid closure. Other ocular features such as eye blink rate, pupil dilation may not exhibit prominent correlation with vigilance level. Therefore, image based algorithm for determination of level of alertness in driver or any operator in general is yet to achieve maturity. Therefore, the thesis has focused on development of image processing algorithms to estimate PERCLOS and SR for determination of alertness level from eye images.

### 1.3.3 Validation of PERCLOS and SR Measurement

The advantage of these ocular features is their non-intrusiveness which does not disturb the operators during their operation. However, being an indirect method of measurement, any inference drawn from such measurements needs a validation by more direct techniques. A complete database that enlists image database of human subjects with varying level of alertness along with the benchmark data like bio-marker, EEG parameter etc. for PERCLOS and/or SR measurement is unavailable. Therefore, it is necessary to conduct experiments for the purpose.

The literature survey indicates that the bio-chemical indicator can be used to validate the measurement of drowsiness index such as PERCLOS. In Experiment II bio-indicators like creatinine, urea, random blood sugar has been recorded along with facial image sequence. Details of the procedure have been elaborated in Chapter 2.

On the other hand measurement of higher range of alertness index like SR can be judged by subjective assessment based on a few standard questionnaires when other alertness measurement cues are unavailable. Standard questionnaire like Stanford sleepiness scale, Epworth scale [8, 9] have been modified to suit for detection of alertness level in Indian context. The procedures for generating such data along with the facial image sequence for estimation of SR are detailed in Experiment III of Chapter 2.

## 1.4 Motivation, Objective and Scope of the Thesis

Discussion in earlier section reveals that PERCLOS and SR can be measured in various ways. However, image based measurement is most non-intrusive. This allows the operators to carry out their activities without any disturbance. Thus this mode of measurement is desirable from operator's point of view. Literature survey indicates most of the technologies measure PERCLOS as prime index for alertness under active NIR illumination while a few have accomplished the task by color image processing. However, PERCLOS is indicative of highly reduced level of alertness which is almost drowsiness. It does not indicate the onset of decline in the level of alertness. Literature suggests that SR can be used to measure loss of attention at an early stage. The state-of-the-art for measuring SR uses fixed camera-object geometric information which may not be true under free movement condition of operator. Moreover, its measurement during drowsy state may not be effective due to longer partial occlusion by eyelid. This has led to decide the state of alertness from a combined measurement of PERCLOS and SR.

It is evident that measurements of PERCLOS and SR by image processing are required to be computationally inexpensive and robust against possible odds in driving scenario. A few basic steps are essential to get the eye image. These are extraction of face and eye from the image sequence. Further the eye image can be processed temporally to determine eyelid and pupil (or iris) motion by the eye image processing. The state-of-the-arts in these contexts are elaborated in relevant chapters for convenience. The important research issues in such context are highlighted in the following subsection.

## 1.4.1 Research Issues

1. Experimental data that includes multi-modal measurement of alertness level in human driver based on ocular feature, EEG parameters, bio-markers, psychometric questionnaire etc. are necessary for development and validation of algorithms.
2. The measurement of PERCLOS is required to be effective both during day and night. No single algorithm has been found to be accurate under varying level of illumination.
3. The relative motion that remains between camera and the operator due to operator's free movement and the vehicle jerking puts constraint on the measurement of both PERCLOS and SR.
    - Change in reflection angle of eye with respect to camera under such motion may provide inaccurate PERCLOS value.
    - It is found in the literature that the measurement of SR is dependent on the 3-Dimensional space parameters like distance between camera and the operator. So, any variation in these parameters may result in incorrect SR estimation.
4. Variation in light intensity throughout day and night and that in orientation of light sources on road put difficulties in the use of intensity-based corner detection and popular canny edge detection based method that assumes edges as step change of intensity level.
5. Partial occlusion of iris by eyelid pose difficulties in finding the iris center (and

    thus the SR measurement) using popular circle detection technique may not be effective.
6. Detection of iris center may be ineffective under noise. Methods based on digital low-pass filter have been used to overcome this problem. However, application of such filters may change the position of image features such as edges, corner points etc.

 Thus the aim of this thesis is set to overcome the challenges in the above mentioned research issues.

### 1.4.2 Contributions of the Thesis

The major contributions of this thesis can be summarized as mentioned below.
1. Experiments for simulated and actual driving have been designed and conducted to record some of the measurement cues described in the above literature review. This facilitates to validate the image based measurement of PERCLOS and SR with more direct measurements such as bio-markers, EEG parameters etc.
2. Image database considering all the required parameter variations like level of illumination, sleep deprivation etc. is unavailable. This leads to generate an extensive database of facial image sequence of human drivers with varying alertness level.
3. A new correlation filters based classification of eye image sequence has been developed and tested for measuring PERCLOS.
4. A new feature called Form Factor (FF) has been introduced and applied to find the center of iris under partial occlusion by eyelid.
5. A new method based on radial Form Factor has been devised to measure the pupil diameter which may provide the state of the operator's level of alertness.
6. A new method of edge detection based on computation of local FF and subsequent detection of corner from the edge map has been developed.
7. Development of method for blind recovery of FF under zero-mean additive independent and identically distributed (i.i.d.) noise. An image quality metric called Signal-to-Total Variance ratio (STVR) instead of the popular signal-to-noise ratio (SNR) or peak-SNR (PSNR) has been proposed and analyzed for accurate estimation of FF without noise.
8. An Extended Kalman Filter (EKF) based estimation method using a simple model has been proposed to improve the accuracy in SR estimation for iris position information.

## 1.5 Organization of the Thesis

The thesis consists of seven chapters as follows:

- **Chapter 1:** *Ocular Features and Human Alertness-* In this chapter the terms alertness, drowsiness, and fatigue have been described. Review on state-of-the-art for measurement of level of alertness by various cues including ocular features have been discussed. The limitations of the image based measurement of ocular indices for alertness such as PERCLOS and SR are highlighted.
- **Chapter 2:** *Experiment Design, Database Generation and Collection-* Experimental procedure for generation of facial image sequence with varying illumination, orientation of driver's face and sleep deprivation have been described. The specifications of the experimental data as well as a few standard external database have been summarized.
- **Chapter 3:** *Detection of Face and Eye-* Several object detection methods are studied to find real-time and robust face and eye detectors. Statistical analysis of these methods is carried out on the in-house facial image database to judge the detection accuracy. The upper bounds of the execution time as well as memory consumption of these algorithms have been formulated and compared with the experimental outcome.
- **Chapter 4:** *PERCLOS: Estimation of Eyelid Motion-* A new correlation filter based method for eye image classification into three states with varying eyelid position has been formulated. A method of synthesis and testing of such filter through Discrete Cosine Transform (DCT) has been proposed to reduce the classification inaccuracy due to high correlation between classes. The experimental results of both DCT and Discrete Fourier Transform (DFT) based correlation filters are compared to elucidate the effectiveness of the proposed method.

Ineffectiveness of PERCLOS for quantifying alertness level during lower level of drowsiness has been demonstrated statistically on the data set generated in Experiment II. The evaluation aids to decide a threshold for PERCLOS value that distinguishes between the drowsy and higher alertness states. A real time system based on the proposed algorithm and the threshold for PERCLOS has been developed to monitor drowsiness on-board.

- **Chapter 5:** *Form Factor: A New Image Feature for Pupil Motion Detection*- In this chapter, the feature called Form Factor (FF) has been defined and its different versions are described. A recovery scheme of this feature under zero-mean independent and identically distributed (i.i.d) additive noise using blind estimation of either SNR or STVR has been proposed. Several blind estimation techniques for SNR and STVR are analyzed mathematically and verified by experimental results.

  The proposed feature has been applied in two applications that are elaborated as case studies. An edge detection method based on local computation of FF has been proposed in Case Study II. The algorithm has been applied on different image data. Qualitative results are presented to exhibit the efficacy of the method over that produced by popular Canny edge detector. Performance indices for quantitative evaluation of both the proposed and Canny edge operator have been studied. The numerical outcomes of the proposed and the Canny edge operator have been produced and compared with that obtained by human visual system.

  In Case Study I, pupil center detection and diameter measurement methods have been proposed using directional version of FF. The results of pupil center detection are compared with that produced by popular Circular Hough Transform (CHT) method to demonstrate its effectiveness. The performance of the pupil diameter measurement has been judged statically over various data sets.

- **Chapter 6:** *Detection and Interpretation of Saccadic Eye Movement from Image Sequence Analysis*- An eye corner detection scheme from eye edge map given by local FF based edge detector has been developed to produce relative position of pupil center. A simple saccadic model for indirect measurement of Peak Saccadic Velocity (PSV) and the Saccadic Duration (SCD) from the temporal information of the pupil position has been proposed. Real-time estimations of the SV and SCD are carried out using Kalman filter (KF) and EKF. Experimental results from these methods are presented and compared with that obtained by low order $(5-7)$ Finite Impulse Response (FIR) filter based method. The SR is computed for the database generated in Experiment III and validated by questionnaire based subjective evaluation of alertness level.
- **Chapter 7:** *Conclusion and Future Scope-* In this chapter conclusion of this thesis and the research directions for future work have been compiled.

# Chapter 2

# Experiment Design, Database Generation and Collection

## 2.1 Introduction

The aim of the thesis is to develop image processing techniques for measuring ocular features to estimate the level of alertness of human drivers. Although the developed algorithms are generic in nature, our experiments are primarily concerned with human vehicle drivers (operators). In practice, the alertness level of vehicle operator is diminished due to several extrinsic and intrinsic factors like ambient conditions, prolonged work load, wakefulness, sleep deprivation etc. A database containing sequence of facial images of human subjects with wide variation in the level of alertness is required for testing the developed algorithms. Other factors that affect the performance of the algorithms are also to be considered. Thus the database should cover possible ranges of illumination variations, face orientations etc. However, an extensive database considering the variations in on-road conditions along with ample variations of alertness level in vehicle operator is not available. Therefore, several experiments under simulated and controlled environment as well as actual on-road driving conditions were conducted to generate video of operator's face along with other data sets like bio-marker, EEG signal, subjective assessment based on questionnaire etc. Three different experiments were conducted. Experiment I was a pilot phase experiment where data were collected from 21 drivers during on-board driving. Second experiment was more extensive where the subjects were kept sleep deprived for 36 hours. Following sections elaborate these experiments.

## 2.2　Experiment-I: On-Road Driving

A preliminary experiment was conducted to explore the variation in the alertness level of the vehicle operators at two sessions- (a) pre-lunch and (b) post-lunch. The aim of the experiment was to record the videos of operators' face in daylight for measurement of drowsiness index (such as PERCLOS). This was conducted at Central Road Research Institute (CRRI), New Delhi, India.

### 2.2.1　Experimental Procedure

The experiment was conducted on 21 healthy male professional drivers aged between $25 - 35$ years without any sleep disorder during summer. They were instructed to refrain from taking any type of medicine or stimuli like tea, coffee or alcohol during the experiment. Apart from recording of the videos, they were engaged in their professional examination with various tasks at CRRI, New Delhi, India. The following tasks were executed between two recording sessions of the video for each of the operators.

1. Computerized subjective assessment
2. Psychomotor tests [10]
    - Complex Reaction Time Test
    - Action Judgment Test
    - Speed Distance Judgment Test
3. Glare and Vision test
4. Electroencephalogram (EEG) data collection

All the tasks are designed to simulate different types of driving activities [10] and can be used to detect the psychomotor vigilance level of the operator.

### 2.2.2　Collection of Data on Actual Road Driving

The videos of the operator's face were captured with a color video camera at 30 fps. The camera was placed in front of the driver on the steering base. Although, the steering bar might be an obstruction for the recording for reasonably small interval, it may not put any constraint on the measurement of drowsiness index. The recording was carried out for 20 minutes at each of the session. The video clip was saved in Windows Media Video (WMV) format with frame resolution of $240 \times 320$. Some snapshots are shown in Figure 2.1. Each of the operators had to appear for the tasks mentioned above before lunch. After lunch, they had to go through another session of video recording while driving on road. In these experiments two stages of alertness level have been captured.

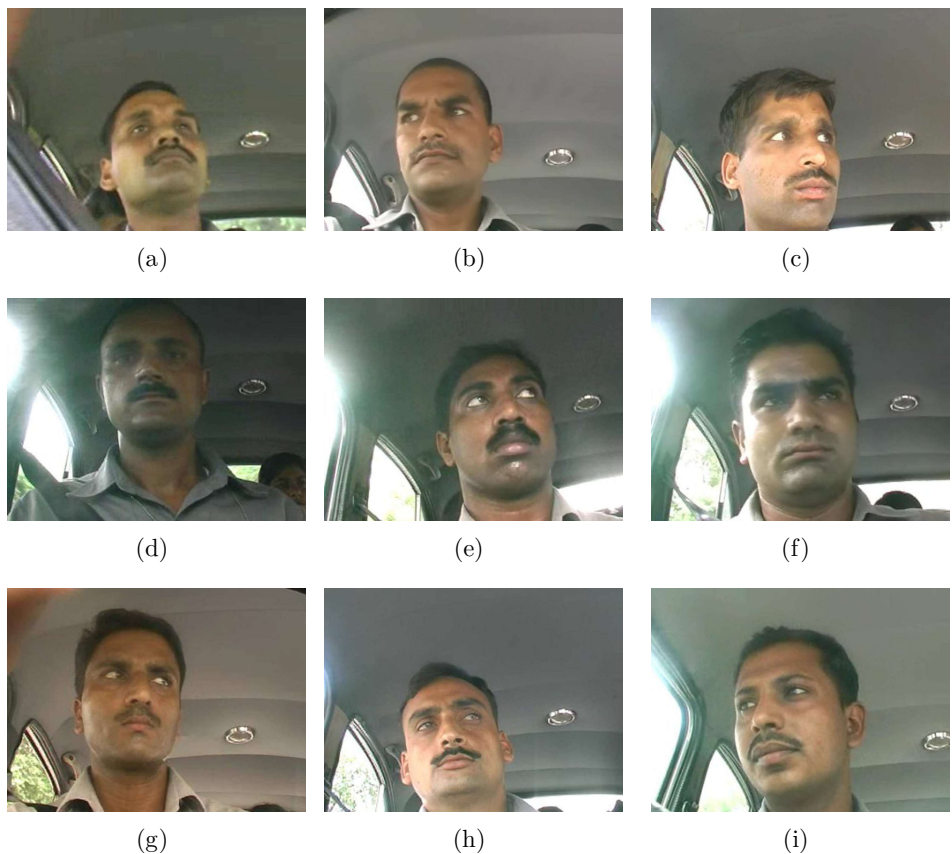

Figure 2.1: A Few Frames of Videos of the Vehicle Operators' Face Recorded on Road Driving

The database was carefully studied, eyes were extracted from the frames and PERCLOS values were measured. It was observed that PERCLOS values for the two stages were not significantly different. The reduced levels of alertness after lunch is primar-ily due to the food intake which may not produce PERCLOS values that distinguish between the two stages. Therefore, an extensive experiment was conducted under simulated and controlled environments with a greater number observation stages covering wide variations in level of drowsiness.

## 2.3 Experiment-II: Simulated Driving for PERCLOS Measurement

Experiments on road with vehicle operators at lower levels of alertness may be fatal while in simulated driving, drowsiness can be induced to achieve larger variations in alertness level without any risk. A few experiments with sleep deprivation along with different activities under simulated driving in controlled environment have been reported [39,55,56]. Literature survey indicates that an experiment can be conducted to induce drowsiness by prolonged activities with sleep deprivation in vehicle operator. Thus an extensive experiment was designed to generate different modes of data along with video recording of operator's face at various levels of alertness. These are mentioned below.

1. Recording of operator's face under daylight and near infrared (NIR) light
2. Collection of blood samples at 8hr interval
3. Collection of psychomotor response data
4. Recording of EEG data
5. Recording of utterance of specific words
6. Collection of questionnaire based subjective assessment data

All the data were collected in a sequence with small time gaps between two operations so that correlation can be determined between different modes of data. The experiment was conducted at Indian Institute of Technology, Kharagpur, India.

### 2.3.1 Procedure for Driving Simulation

The video recording of operator's face was a part of the extensive experiment conducted to induce drowsiness leading to fatigue. It was conducted on twelve healthy male vehicle drivers without any sleep disorder and aged between $19-45$ years. They were instructed to refrain from taking of any type of medicine or stimuli like tea, coffee or alcohol during the experiment. The whole experiment was executed in two phases with 6 subjects in each phase.

Each phase of the experiment consisted of a number of identical sessions. Each of these sessions started with examination of the subjects by a physician. The operators who were observed to be fit were asked to do some tasks to reduce their level of alertness. These are as follows-

1. Exercises on a tread mill for $2-5$ min to induce physical fatigue
2. Simulated driving for about 15 min to induce physical, visual, and mental fatigue
3. Auditory and visual tasks for 15 min to generate mental and visual fatigue
4. Computerized game related to driving for about 20 min

A single session (or stage) of the experiment spanned about 3 hr. The subjects were allowed to read books or newspapers in between the stages in a room monitored by CCTV camera. These activities were repeated 12 times up to 36 hr. The schedules of each stage in the experiment for two subjects are shown in Table 2.1.

### 2.3.2 Collection of Data

Although the experiment consisted of several other modes of data collection, the recording of video under simulated driving conditions and the collection of blood samples are elaborated here. The blood sample analysis may provide the state of alertness in human drivers as it prominently reflects the changes [27]. The results of the image based algorithms can be correlated with the blood sample analysis.

#### 2.3.2.1 Video Recording with Low Speed Camera

A few movie clips of road scenes that generate the feeling of vehicle driving were combined in a sequence and played in front of the drivers. Two such movie clips were made to simulate the day as well as night driving conditions. A soothing and slow music played along with the videos. These movies are played using a simulated car driving system as shown in Figure 2.2. Each of the subjects was instructed to follow the road in the movie clip while they drive for 15 min at a stretch.

Table 2.1: Commencement Time of Stages for Two Subjects

| Vehicle Operator | Day No. | Stage No. | Starting time |
|---|---|---|---|
| Subject 1 | 1 | 1 | 9 : 00am |
|  |  | 2 | 12 : 15pm |
|  |  | 3 | 3 : 30pm |
|  |  | 4 | 6 : 45pm |
|  |  | 5 | 10 : 00pm |
|  | 2 | 6 | 1 : 15am |
|  |  | 7 | 4 : 30am |
|  |  | 8 | 7 : 45am |
|  |  | 9 | 11 : 00am |
|  |  | 10 | 2 : 15pm |
|  |  | 11 | 5 : 30pm |
|  |  | 12 | 8 : 45pm |
| Subject 2 | 1 | 1 | 9 : 30am |
|  |  | 2 | 12 : 45pm |
|  |  | 3 | 4 : 00pm |
|  |  | 4 | 7 : 15pm |
|  |  | 5 | 10 : 30pm |
|  | 2 | 6 | 1 : 45am |
|  |  | 7 | 5 : 00am |
|  |  | 8 | 8 : 15am |
|  |  | 9 | 11 : 30am |
|  |  | 10 | 2 : 45pm |
|  |  | 11 | 6 : 00pm |
|  |  | 12 | 9 : 15pm |

The videos of faces of the drivers were recorded using a Charge Coupled Device (CCD) camera with a facility to switch over into night capture mode. Thus, the recordings can provide the gray scale as well as NIR monochrome frames. The camera was placed behind the steering in front of the driver. It was at a distance around 60 cm from the subject. The video was captured with 30 fps speed under day as well as night driving mode with a resolution of $480 \times 720$ and saved in WMV format in memory of the camera. A few frames from these videos are shown in Figure 2.3 and 2.4 recorded under daylight and night driving condition.

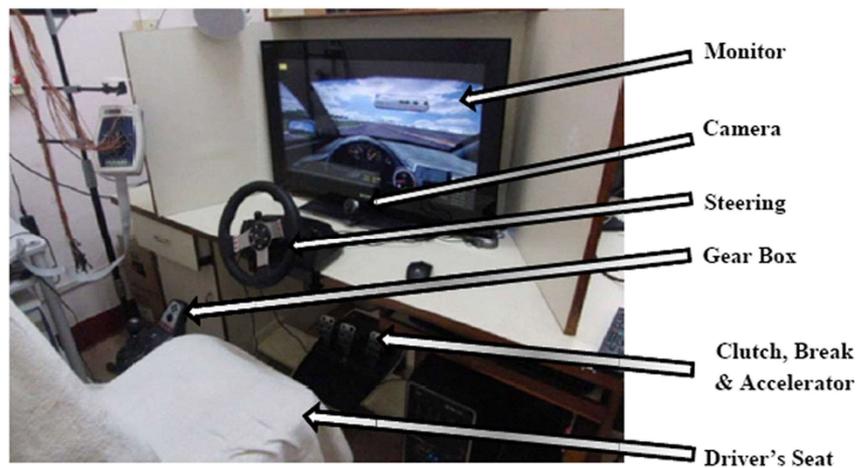

Figure 2.2: Driving Simulator for Experiment

#### 2.3.2.2 Collection of Blood Sample

The blood samples were collected from each of the subjects in 8hr intervals for five stages (for subject 1: Day 1: 08.30 hr, 16.30 hr; Day 2: 00.30 hr, 08.30 hr and 16.30 hr). The blood samples were collected by a medical professional from alternate arms at alternate stages. The anterior region of the elbow was sterilized by applying a swab soaked with 70% alcohol. 5 ml of blood was collected by puncturing the brachial vein through 5 ml disposable syringes fitted with 22 G needles (BD biosciences, India Ltd.). 2 ml of blood was collected in a tube containing clot inhibitor (Sodium heparin, 72 USP units, Crest Diagnostics) to obtain whole blood for glucose estimation. The remaining 3 ml of blood was collected in tubes containing clot activator and coated with silicon (Crest Diagnostics) to obtain serum for serum urea and creatinine estimation. The samples were vortexed 8 times for proper mixing followed by preserving in refrigerators at $2-8°\ C$. The samples were analyzed for measuring the level of blood glucose, serum urea and serum creatinine that varies with the alertness level.

## 2.4 Experiment-III:Simulated Driving for Saccadic Ratio Measurement

The video data collected in the earlier experiments have been recorded with 30 fps speed and can easily be used for measurement of slow eyelid closure index (e.g. PER-

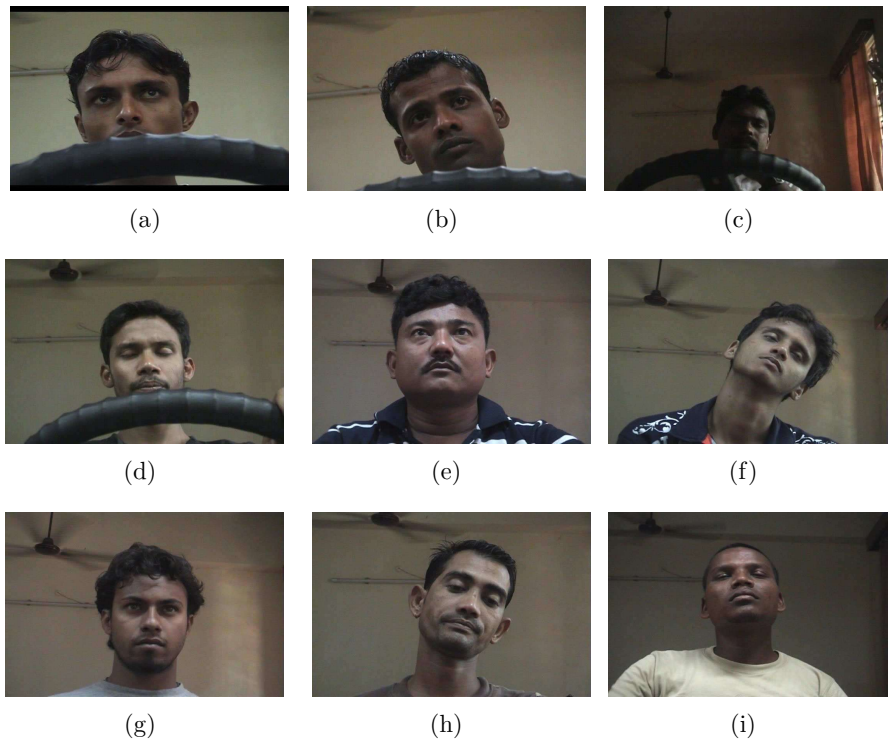

Figure 2.3: A Few Frames of Videos of the Vehicle Operators' Face for Various Alertness Level Recorded under Daylight Condition

CLOS). However, rapid movement of iris like saccade cannot be determined from the video with low frame rate [53].

### 2.4.1   Procedure for Driving Simulation

This experiment has been conducted on 3 healthy male vehicle drivers aged between $19-24$ years without any sleep disorder. They were instructed to refrain from taking any type of medicine or stimuli like tea, coffee or alcohol during the experiment. In this experiment, the following modes of data were collected.

1. Recording of operator's face with high speed camera
2. Collection of questionnaire based subjective assessment data

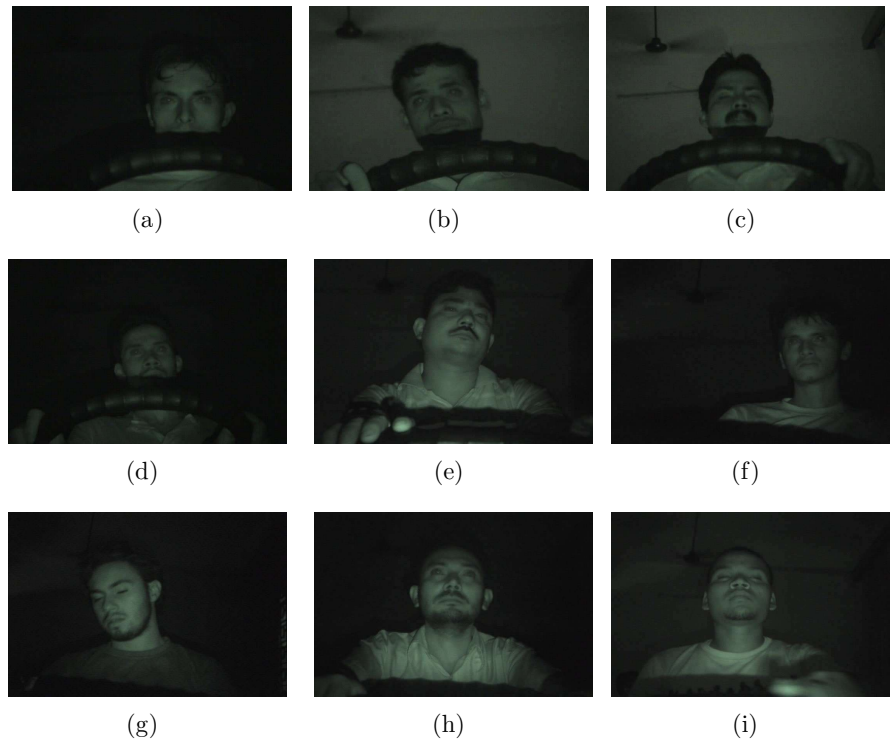

Figure 2.4: A Few Frames of Videos of the Vehicle Operators' Face for Various Alertness Level Recorded at Night

The subjective feedback can be used to correlate with the output provided by the proposed algorithm.

### 2.4.2 Collection of Data

The experimental data was collected in two stages. The time interval between two stages was 8 hr. In the first stage, the subjects were at higher level of alertness while in the second stage they carried out some activities to reduce the level of alertness. These are mentioned below.

1. Auditory and visual tasks for 15 min
2. Computerized game related to driving for about 20 min
3. Simulated driving for about 15 min

In both the stages, videos of operator's face and responses to psychometric questionnaire were collected. Procedures are detailed in the following subsections.

#### 2.4.2.1   Video Recording with High Speed Camera

The movie clip which was developed for Experiment II was also used here. It was played using a simulated car driving system. Each of the subjects was instructed to follow the road in the movie clip while they drive for about 15 min at a stretch.

The video of face of the drivers were recorded using a camera with speeds ranging from $60 - 1000$ fps and a minimum resolution of $480 \times 360$ under day driving mode. The camera was placed behind the steering in front of the driver. It was at a distance of 60 cm from the subject. The recorded videos were saved in WMV format in memory. A few frames from these videos are shown in Figure 2.5.

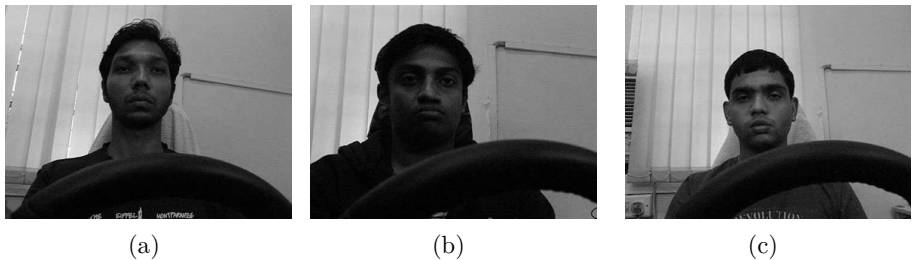

(a)  (b)  (c)

Figure 2.5: A Few Frames of Videos of the Vehicle Operator Recorded with High Speed Camera under Visible Light

#### 2.4.2.2   Collection of Subjective Assessment Data

At the end of video recording, in each stage, the drivers were asked a set of questions. The responses were recorded to analyze the subjective judgments about their state of alertness. The questionnaire is presented in Table 2.2. The analysis were carried out on a scale of $0 - 10$ with low value corresponds to lower level of alertness. The score for level of alertness is obtained as weighted average of all the responses.

Table 2.2: Questionnaire for Subjective Assessment of Alertness Level

1. Are you tired now? If yes, to what degree are you feeling tired? (Scale: 1-10)
2. To what degree may your tiredness affect your ability to work? (Scale: 1-10)
3. To what degree are you feeling alert now? (Scale: 1-10)
4. To what degree are you feeling able to concentrate? (Scale: 1-10)
5. To what degree are you feeling energetic? (Scale: 1-10)
6. To what degree are you feeling able to think clearly? (Scale: 1-10)
7. Chance of yawning:
   a. If allowed to read a newspaper inside the car. (Scale: 0-4)
   b. If allowed to listen music. (Scale: 0-4)
   c. If allowed to drive in a long and monotonous road. (Scale: 0-4)

## 2.5 Limitations of the experiments

Three types of experiments were designed for collecting data of human subjects undergoing varying level of alertness. The first two experiments generated the video data for image based measurement of PERCLOS. The third one was designed to produce the video for estimation of saccadic parameter that changes with varying level of alertness. The second and third experiments could have been combined. However, unavailability of high speed video camera during the Experiment II and the high cost insisted us to carry the experiments separately. These experiments also possess some limitations. A large number of subjects were required. This experiment is still in initial phase and has been carried out on three subjects only. The prime issues in experiments involving human vehicle drivers in laboratory and on road driving can be highlighted as below.

- Motivation of subjects for such experiments
- Ethical constraint
- Risk of accident
- Simulation of the exact road environment in laboratory

## 2.6 External Database

Apart from the databases generated in laboratory, several external benchmark databases are used for performance judgements of the proposed image based algorithms. These are specified in Table 2.3.

Table 2.3: Specification of External Databases

| Database | Original Image Type | No. of Images | Image Resolution | Other Specification |
|---|---|---|---|---|
| Berkley Segmentation Data Set (BSDS) [57] | Wide range of natural scene | 300 | $321 \times 481$, $481 \times 321$ | $5-9$ binary edge maps per image segmented by human subjects are available |
| IMM Face Database [58] | Human face | 240 | $640 \times 480$ | 7 female and 33 male subjects without eyeglasses |
| BioID Face Database [59] | Human face | 1521 | $384 \times 286$ | 23 test persons |
| CBSR Near Infrared (NIR) Face Database [60] | Human face | 3940 | $640 \times 480$ | 197 test persons with and without eyeglasses |
| UBIRIS v.1 Database of Eye [61] | Human eye | 1887 | $2560 \times 1704$ | 241 test persons |

## 2.7   Discussion

In this chapter three experiments under actual as well as simulated driving conditions have been designed. The procedures of these experiments for collection of videos of face of the drivers with varying alertness levels have been presented. In Experiments I and II the videos were recorded with a frame rate of 30 fps. These videos can be used to measure the drowsiness index such as PERCLOS. The third experiment was conducted with high speed camera ($60 - 1000$ fps) to capture saccadic eye movement under simulated driving conditions. These videos on facial images can be used for different purpose such as testing of facial as well as ocular image processing methods. Thus the 'Indian Institute of Technology, Kharagpur' (IITKGP) face and eye databases are generated from these videos for experimental validation of different algorithms developed in this thesis. Apart from these, a few other databases on eyes have also been used to evaluate the performance of the proposed algorithms. The specifications of the IITKGP database can be summarized as in Table 2.4.

Table 2.4: Recordings of Facial Images for PERCLOS and Saccadic Ratio Measurement

| Expt. No. | No. of Subjects | No. of Stages | Nature of the Experiment | Data Collected |
|---|---|---|---|---|
| 1 | 21 | 2 | Actual driving and related psychomotor vigilance tests | **Video of Facial Images using 30 fps camera**, EEG, Subjective Assessment |
| 2 | 12 | 12 | Simulated driving, physical and mental tasks, 36 hr sleep deprivation | **Video of Facial Images using 30 fps camera**, EEG, Subjective Assessment, Speech, Spirometry, Oximetry, Blood sample |
| 3 | 3 | 2 | Simulated driving, 12 hr sleep deprivation | **Video of Facial Images using high speed (60 − 1000 fps)camera** |

# Chapter 3

# Detection of Face and Eye

## 3.1 Introduction

Detection of an object from the overall image essentially falls under the field of pattern classification. Several methods have been devised to segment the image into object and non-object regions. This is based on the distribution of raw pixels in the image or that in the representative features. In all such cases the decision criterion (a simple threshold or a boundary) is derived from a number of example images with and without the object of interest. Subsequently the object is detected from the test image based on this decision criterion. The performance of this algorithm depends on the context of application and various imaging conditions. Therefore it is necessary to analyze these algorithms in different contexts and conditions. The present work analyzes a few simple and popular algorithms for detection of face in driving environment. The general framework of such algorithm can be described as shown in Figure 3.1.

The general framework for object detection considers either the raw pixel values or some feature values computed from the raw pixels. Therefore the projection on feature space can be thought of as a mapping from raw pixel to feature values. Thus, based on the nature of mapping, the object detection can be performed by segregating the object and non-object from their spatio-temporal probability distributions. In such cases the temporal variation in pixel distribution in background and foreground are modeled and a threshold is chosen to distinguish these two regions. The trained model is applied to separate the object of interest from its background. The accuracy of the object detection is based on the goodness of the background model.

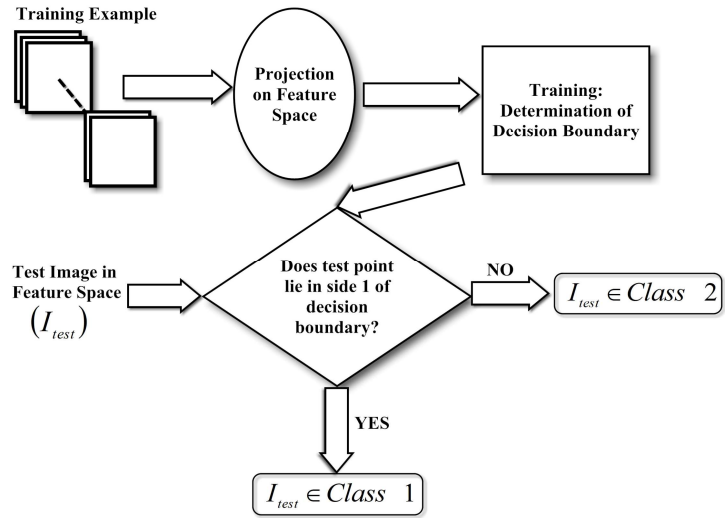

Figure 3.1: General Framework for Object Detection

On the other hand the raw pixel image transformed into feature space is used to detect object by use of thresholds on the feature values. Weakness in one feature may be compensated by combining a number of different feature based object detection. Accuracy of the object detection is based on the choice of the feature type. The speed of the operation depends on the number of features as well as the size of the image. The thresholds for each of the feature values produce a decision surface by combining them all. Determination of such decision surface produces the model of the object in the image. The test object is detected by determining the side of the surface in which it lies. The modeling can be done by several ways viz. - Ada Boost, Support Vector Machine (SVM), Neural Network based learning etc.

In this thesis recent object detection techniques such as background subtraction and Haar-like feature based Classifier (HCL) have been analyzed for face and eye detection. The background subtraction methods are chosen as probable candidates as these are based on very simple algorithms [62–66]. More recently the Haar-Like Feature (HLF) based method with Ada-boost technique for training has been extensively used in real-time object detection [67, 68]. In this chapter some of these algorithms have been compared with respect to their accuracy and computational complexity. Theoretical deductions of the time and memory complexities have been provided.

## 3.2  Background Subtraction Methods

Background subtraction techniques generally find applications in traffic surveillance systems for detecting and tracking moving objects [62, 69, 70]. There has been considerable research in the past leading to the proposition of several algorithms in the literature. However, all of them may not qualify for embedded real time applications, because of large execution time and memory requirement. This paper considers five recently proposed background subtraction algorithms [62–66] with an application to the extraction of the facial image of a human operator from the image captured inside a vehicle during driving.

A comparative assessment of the computational complexity and the robustness to varying illumination has been carried out for the following background subtraction methods.

1. Single Mode Gaussian model (SMG) [62]

2. Multi-Color Space (MC) [63]

3. Texture (TEX) [64]

4. Kernel Density Estimation (KDE) [65]

5. Code Book model (CB) [66]

The basic idea of the background subtraction technique is to extract the object of interest (foreground) by comparing the current image frame with the model of the background scene. A statistical representation or the model of scene background is obtained from number of previous empty image frames (without the object of interest). Both parametric and non-parametric approaches have been used for modeling the background. In parametric modeling, the actual intensity distribution of pixels over time is approximated by a specified statistical distribution. The parameters of this distribution are estimated from the sample background frames [65]. On the other hand, in non-parametric approach, there is no assumption on the underlying distribution. The probability density function of actual distribution of pixel intensity is estimated directly from the sample data of background scene. Once background model is compared with the current image frame, thresholding technique is used to classify each pixel in the current frame into either background or foreground. In a video sequence several initial empty frames are used for background modeling and foreground object is detected in the subsequent frames. The detection of the face of human operator during driving may be difficult for the following reasons -

- The variation of illumination level is a common phenomenon.

- Skin complexion, facial features such as beard, eye glasses may also lead to misclassification of a pixel into background or face.

Therefore the algorithm for face-extraction is required to be simple, fast, and robust against changes in illumination condition, variation in skin complexion, and blurring effect due to relative motion between camera and object (face of the driver) etc. An experimental study has been carried out on the above five background subtraction methods. Necessary modifications have been incorporated in the algorithms for improving the detection accuracy.

The present introductory section is followed by a brief description of main features of the algorithms used in this study. The subsequent sections present the experimental results along with the various performance measures such as accuracy in face detection, time and memory complexity for each algorithm.

### 3.2.1 Single Mode Gaussian Model based Method

Thongkamwitoon *et al.* [62] have used a statistical approach for background modeling. They have assumed a single mode Gaussian distribution over time for each pixel in the image. At the initial step, background image is modeled using $N_f$ frames of empty scene. The background model is obtained by computing Expected color vector ($E_{i,j}$) and Color covariance matrix ($\mathbf{C}_{i,j}$) for each pixel at $(i,j)$ over $N_f$ frames. Then any distortion of the test color vector $\mathbf{X}_{i,j}[n]$ from mean color vector ($\mathbf{E}_{i,j}$) of the background model indicates intrusion of new object in the scene (i.e. the foreground). This distortion is measured by decomposing $\mathbf{E}_{i,j}$ into mutually orthogonal distortion parameters such as 'Brightness distortion factor' ($\alpha_{i,j}[n]$) and 'Color distortion factor'

($\lambda_{i,j}[n]$) as follows-

$$\alpha_{i,j}[n] = \arg_\psi \min(\mathbf{X}_{i,j}[n] - \psi\mathbf{E}_{i,j})^2 \qquad (3.1)$$

$$\lambda_{i,j}[n] = \|\mathbf{X}_{i,j}[n] - \alpha_{i,j}[n]\mathbf{E}_{i,j}\| \qquad (3.2)$$

Each pixel is assigned to either foreground or background based on these factors [62].

### 3.2.2  Multi-color Space based Method

Hong *et al.* [63] proposed a background subtraction method that utilizes the statistical characteristics of pixel's color values in both RGB and normalized RGB ($r, g, b$) color space [63]. Each pixel at ($i, j$) in reference image is modeled by the mean color vector and standard deviation (s.d.) color vector in each of the color spaces. Two threshold vectors for both color spaces have been computed by taking fraction of the corresponding s.d. color vectors. Two discriminant functions that measure the deviation of each test pixel from its means in both the color domains are obtained [63]. Each pixel is classified into either foreground or background based on the values of these determinant functions.

### 3.2.3  Local Binary Pattern based Method

A recent background subtraction method proposed by Heikkilä *et al.* [64] uses the group of adaptive Local Binary Pattern (LBP) histogram obtained by computing over a circular region around each pixel. It is a non-parametric spatial approach unlike the two methods discussed above. LBP is obtained after taking threshold of the neighborhood of each pixel by its intensity value for each color channel (i.e. R, G, and B). A histogram of LBP has been calculated over a circular region of radius $R_{region}$ and used as the feature vector of a particular pixel. This radius $R_{region}$ is user adjustable. A few initial frames are used to obtain a group of $N_{lh}$ adaptive LBP histograms $\{\mathbf{h}_0, ......., \mathbf{h}_{N_{lh}-1}\}$ for each color channel. Here value of $N_{lh}$ is also user settable. Contribution of each of these histograms may not be equal to form the background model. Each model histogram is therefore associated with a weight ($w_n : 0 \leq n \leq N_{lh} - 1$) between 0 and 1 so that sum of all weights is equal to 1.

In a similar way, LBP histogram (**h**) of each pixel per color channel in the current frame is obtained and its proximity to the model histograms are measured using the following relations-

$$\bigcap(\mathbf{h}_n, \mathbf{h}) = \sum_{i=0}^{N_h-1} \min(h_{n,i}, h_i) \quad : 0 \leq n \leq N_{lh} - 1 \quad (3.3)$$

where $N_h$ is the number of histogram bins (it is equal to power of 2 ). If the proximity is higher than a threshold for at least one model histogram then the pixel is assigned as background; otherwise it is predicted as foreground.

### 3.2.4 Kernel Density Estimation based Method

A nonparametric approach of background subtraction has been proposed by Elgamal *et al.* [65]. The method takes the raw pixel intensity (color vector) as the input feature. It assumes no specific shape of the pixel distribution. Instead it chooses kernel density function estimation for foreground object detection. Because of its continuity, differentiability, and locality properties a Gaussian function has been taken as the kernel. In background modeling phase, intensity samples of each pixel at $(i, j)$ are stored. These are taken for kernel density estimation as:

$$P(\mathbf{x}_t) = \frac{1}{N_s} \sum_{i=1}^{N_s} \prod_{q=1}^{N_{cd}} \frac{1}{\sqrt{2\pi\sigma_q^2}} e^{-\frac{(x_{tq}-x_{iq})^2}{2\sigma_q^2}} \quad (3.4)$$

where, $N_s$ is the number of samples for each pixel. $\mathbf{x}_t$ is $N_{cd}$-dimensional color feature for the pixel at $(i, j)$. $\sigma_q$ is bandwidth of the $q^{th}$ color channel for a given pixel and $N_{cd}$ is the number of color channels (for RGB color space it is equal to 3). The bandwidth for each color channel is determined as in [65]. A global threshold $Th$ is applied on the estimated probability ($P(\mathbf{x}_t)$). If $P(\mathbf{x}_t) < Th$ for a pixel then it is classified as foreground pixel. Otherwise the pixel belongs to the background.

### 3.2.5 Code Book Model based Method

A nonparametric method that uses codeword for each pixel instead of its raw intensity value is proposed by Kim *et al.* [66]. An initial codebook is developed with $N_s$ sample values for each pixel in RGB color space. A codebook consists of one or more codeword for each pixel. The length of each codebook (say $L_{cb}$) may vary depending upon its sample variation [66]. Each codeword comprises of an RGB vector, intensity values and temporal variables as follows

$$Codeword = [V_c, X_{min}, X_{max}, f_{cb}, l_{cb}, m_{cb}, n_{cb}] \qquad (3.5)$$

Where $V_c = [R_c, G_c, B_c]^T$ is color vector in $c^{th}$ codeword. $X_{min}$ and $X_{max}$ are minimum and maximum intensity of all pixels assigned to this codeword. $f_{cb}$ is the frequency with which the codebook has occurred over the sample values. $l_{cb}$ is the longest interval during the training period that the codeword has not recurred. This is termed as Maximum Negative Run-Length (MNRL). $m_{cb}, n_{cb}$ are the first and last access times, respectively, that the codeword has occurred. A color distortion factor ($\lambda$) from the codeword color vector and a brightness factor are used to find a match between the test image and the background model. If there is no match for the test pixel it is assigned to foreground.

## 3.3 Feature based Object Detection Method

The detection of face or eye can also be accomplished by mapping the raw pixels into features space. A minute observation may reveal that different parts of a face may have different absolute values of illumination depending on the position and the intensity of the light sources. However, these regions on the face hold ordinal relationships in reflectance characteristics between each other [71, 72]. In [73] Oren *et al.* have used ratio template computation via wavelet transform to detect human pedestrians. A neural network based face detection technique has been proposed in [74]. Higher computational burden of the algorithm has motivated Viola *et al.* [67] to develop a real time algorithm with a few rectangular types Haar-Like Features (HLF) for the purpose. This technique has been found to be fast and robust for detection of frontal faces and eye. The tilted and in-plane rotated faces require an extended set of HLF developed in [68, 75]. However, the computational complexity of these algorithms with larger number of features may restrain their application in real-time. The basic concept of the technique is to classify the whole image into face and non-face categories using a supervised learning method. In this thesis the training of such classifier has been critically reviewed and analyzed for development of training set that includes larger variations in face orientation with smaller number of simple Haar-like features. The method is explained in the following subsections.

### 3.3.1 Feature Space

For face detection, primarily two rectangle, three rectangle, four rectangle features are chosen (Figure 3.2. The choice of these features depends on the type of object to be detected. It is observed that the rectangular features can characterize the ordinal relationship among the face regions effectively.

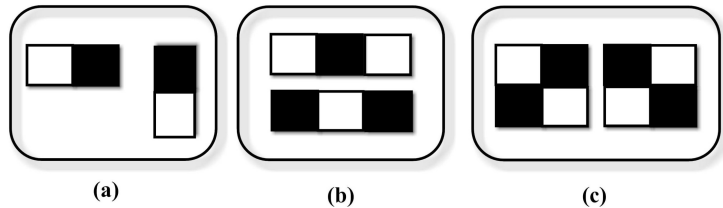

(a) (b) (c)

Figure 3.2: Different Type of Features

The feature value is computed as the difference between the area sums of the neighboring regions. The computation of these feature values has been made faster by representing the image in integral form. Thus a group of pixels in the image is transformed into a feature point in the space. The number of features determines the dimension of the feature space. Therefore any subregion of an image (group of pixel) can be represented as a position vector in $N_d$-D feature space by $N_{fe}$ such features.

### 3.3.2 Training of the Classifier

Learning of the classifier implies finding out a decision surface in the feature space as shown in Figure 3.3. In this figure three features are shown for simplicity. The decision surface can be obtained by several techniques such as Ada-boost, SVM, and neural network based learning etc. However, Ada-boost technique has been found to be computationally less expensive among these learning methods.

A separate hypothesis for each feature is formed for each point in the feature space. Thus a rule is set to classify a point as object or non-object based on single

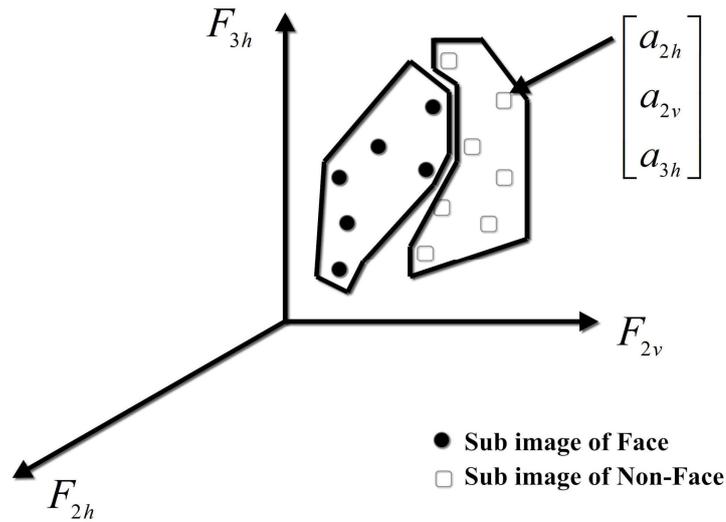

Figure 3.3: Mapping of Images into Features Space

---

$a_{2h}$, $a_{2v}$, and $a_{3h}$ are 2-rectangular horizontal and vertical, 3-rectangular horizontal feature values respectively for a particular image region feature value in 2D feature space as shown in Figure 3.4 for convenience. However, this can be extended to higher dimensional feature space. Here, the image subregions are classified into face or non-face based on the value of vertical two rectangular feature ($F_{2v}$). This leads to a weak hypothesis as the classification by single feature is imprecise and inaccurate in nature. It has been found that the boosting technique combines all the weak classifiers linearly to produce robust classification decision. The weights of such linear combination are computed based on probability distribution of image subregion as face or non-face. This linear combination of weak classifier separates the whole feature space into two sub-spaces like face and non-face.

### 3.3.3 Classification of Objects

The trained classifier can be used to detect faces in test image. The whole test image is divided into number of subregions. The number of subregions is decided by assuming an effective base resolution obtained empirically. The Haar-like features are computed for each of these subregions at multiple resolution levels. The point representing the image subregion is put into the model of the classifier to decide the position of the image relative to the decision surface. The classification algorithm is summarized in the Algorithm 3.3.1 in the form of pseudo code. The input to the algorithm is the image $(I)$ in consideration and the dimension of the feature space$(N_d)$.

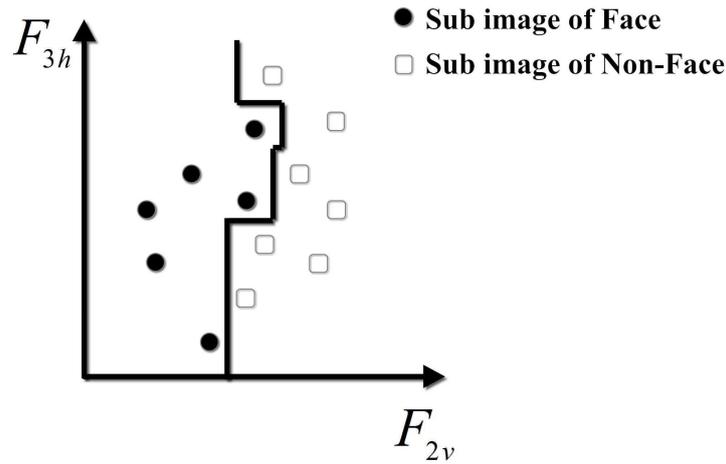

Figure 3.4: Classification of Feature Points by Weak Hypothesis

**Algorithm 3.3.1:** FACECLASSIFY$(I, N_d)$

$ImageWidth \leftarrow W_I$
$ImageHeight \leftarrow H_I$
$BaseWindowWidth \leftarrow W_B$
$BaseWindowHeight \leftarrow H_B$
$WindowNumber \leftarrow \frac{W_I H_I}{W_B H_B} = \frac{N}{n_B}$
**for** $W_c \leftarrow 1$ **to** $\frac{N}{n_B}$
$\quad$**do** $\begin{cases} \text{for } F_c \leftarrow 1 \text{ to } N_d \\ \quad \text{do } \{Compute\ HLF\ in\ constant\ time\ t_c \\ Compute\ classifier\ function\ and\ assign\ each\ window\ into\ a\ class \end{cases}$

## 3.4 Implementation

The algorithms mentioned above have been implemented in C++ on visual C 6.0 platform with some of the functions from OpenCV library. The program is executed on 3.00 GHz Pentium-D processor with Windows-Vista as the operating system. The RAM (random access memory) size is 1 GB. The details of the implementation for background subtraction as well as feature based classifier are presented in the follow-ing subsection.

## 3.4.1 Background Subtraction based Classifier

Training in background subtraction method requires frames that are devoid of object of interest. A number of such empty frames are used to model the background. It has been found that modeling with 20 frames provides reasonably accurate background characteristics. An extensive observation indicates that the effective number of training frames varies with context. It can also be observed that each of the background subtraction has its own training parameters that influence the object (foreground) detection accuracy. The numbers of tuning parameters that can be set by the user are summarized in Table 3.1. It indicates that achieving the optimum separation between background and object is easier using KDE based modeling. The tuning parameters are set by tests on the training images and the best configuration is retained.

Table 3.1: Adjustable Parameters for Background Training

| Classification Method | No. of Tuning Parameter |
|---|---|
| Single Mode Gaussian Model | 2 |
| Multi-Color Space Model | 2 |
| Local Binary Pattern Model | 3 |
| Kernel Density Estimation based Model | 1 |
| Code Book Model | 3 |

As described in Chapter 2, a number of videos recorded under varying illumination are considered from which the training images are obtained from empty frames. Two discrete levels of illumination (one with more than 200 Lux and the other with less than 50 Lux) have been considered for test frames. The data set is divided into 8 groups corresponding to 8 subjects with two levels of illumination. The background and face pixels in each of these image set have been manually identified by human scorers. The average face boundary out of these scores is considered as the reference for computation of the Receiver's Operating Curve (ROC) for all the background subtraction methods. Further, the frame size is varied to find out the time complexity of the algorithms with the number of pixels in each frame.

### 3.4.2 Haar-like Features based Classifier

The HLF based classifier models the decision hyper-surface as a linear combination of weak hypotheses. This step needs to consider variation in ambient condition such as changes in illumination, different face orientation etc. Although, training sets are available separately for frontal and profile faces, use of these in cascade increases the computational complexity. Moreover, introduction of rotated features developed in [68] to improve the detection accuracy will increase the feature space dimension. Therefore training set has been generated to map the training image in lower dimensional feature space accommodating larger variations in face orientation. The training steps are mentioned in the following subsection.

#### 3.4.2.1 Training steps

The detail procedures to develop training set for face detection using OpenCV 1.0 library can be found in [76]. The trained model is saved in EXtensible Markup Language (XML) format. The procedures are summarized as follows-

1. Training images containing faces in different orientations and those devoid of faces under varying illumination are collected. The position of faces in images is extracted. A MATLAB Graphical User Interface (GUI) is designed to automate the process.
2. The extracted information for positive (with face) and negative (without faces) images are saved in two separate text files. The file containing the positive image information uses the format given below while that for negative examples contains their name only.

    $<FileName>\_<i>\_<j>\_<width>\_<height>.jpg$

    where i, j, width, and height define the position of the faces in the positive examples.
3. These files are used to combine all information in one vector (.vec) file.
4. The classifiers are trained using the vector file with the help of training functions available in OpenCV library and saved in XML format for convenience.

#### 3.4.2.2 Training Data Set

A database is formed by extracting positive examples randomly from videos recorded in different experiments as described in Chapter- 2. Here, different orientations of full and occluded faces (like in-plane rotation, off-plane rotation in frontal and profile direction within $\pm 60°$) in gray scale as well as NIR monochrome images have been considered for training. A few example images are shown in Figure 3.5.

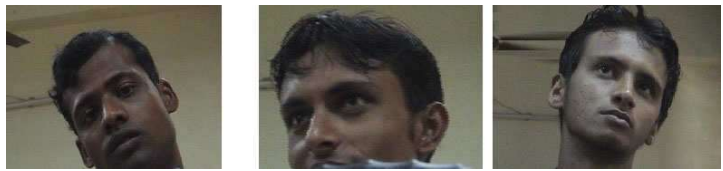

Figure 3.5: A Few Training Images

#### 3.4.2.3 Test Data Set

The test images that are used for background subtraction methods are also utilized here. In addition to that some non-face images have been added with this test data set. All images in this database are kept at same resolution. Subsequently, this data set is divided into 8 groups corresponding to 8 subjects with two levels of illumination. Each of these groups contains 115 images of both faces and non-faces. The numbers of positive and negative examples within each of these data sets are manually counted and noted for computation of the ROC curve for the classifier.

## 3.5 Experimental Results

### 3.5.1 Detection Performance

The detection performance of classifiers can be judged by computing the points on ROC under various conditions [77]. An ROC is a mapping between True Positive Rate (TPR) and False Positive Rate (FPR). The TPR and FPR are obtained from a confusion matrix as shown in Table 3.2. The points on the ROC curve computed from the test database are interpolated to construct its smooth version for each of the classifier.

The TPR (hit rate) and FPR (false alarm rate) can be computed following equations 3.6 and 3.7 respectively.

Table 3.2: Confusion Matrix

|  |  | True Class | |
|---|---|---|---|
|  |  | **p** | **n** |
| Hypothesized | **Y** | True Positive | False Positive |
| Class | **N** | False Negative | True Negative |

$$tpr = \frac{tp}{tp + fn} \quad (3.6)$$

where $tpr$ = true positive rate, $tp$ = number of true positives, $fn$ = number of false negatives.

$$fpr = \frac{fp}{fp + tn} \quad (3.7)$$

where $fpr$ = false positive rate, $fp$ = number of false positives, $tn$ = number of true negatives

Figure 3.6 and Figure 3.7 show different ROC curves for the detection of face under higher and lower levels of illumination.

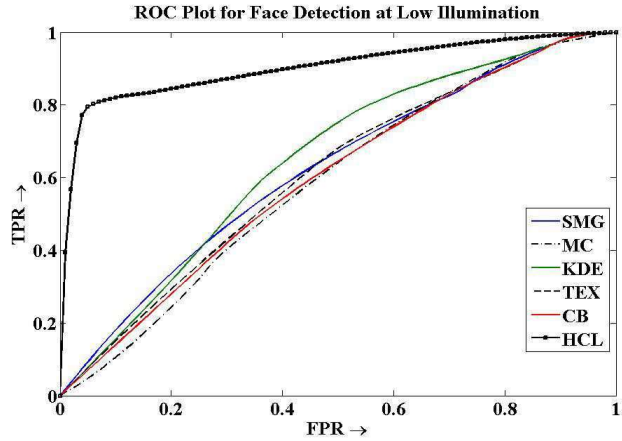

Figure 3.6: Comparative Study of ROC Curves for Face Detector with Lower Illumination Level

The overall performance of a classifier can be obtained from the Area Under the Curve (AUC) of ROC. Table 3.3 shows the AUC values for different classifiers.

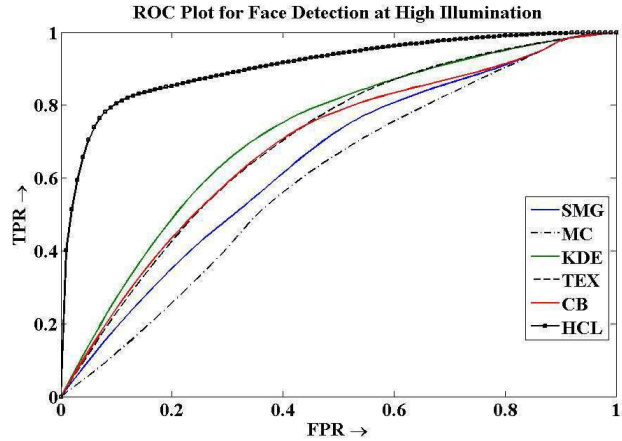

Figure 3.7: Comparative Study of ROC Curves For Face Detector with Higher Illumination Level

Table 3.3: Table Showing the AUC of ROC Curvec for the Classifiers with Different Illumination Level

| Classifier based on | AUC | |
| --- | --- | --- |
| | HIL | LIL |
| Single Mode Gaussian model | 0.6568 | 0.6214 |
| Multi color space model | 0.5988 | 0.5882 |
| LBP texture model | 0.6982 | 0.6094 |
| Kernel density estimation model | 0.7244 | 0.6475 |
| Codebook model | 0.6868 | 0.5980 |
| Haar-like feature | 0.9074 | 0.9063 |

HIL: High Illumination Level ($\geq$ 200 Lux), LIL: Low Illumination Level ($\leq$ 50 Lux)

Figure 3.8 and Figure 3.9 show some detection results using the HLF based classifier.

### 3.5.2 Algorithm Complexity

#### 3.5.2.1 Time Complexity

In order to develop a real-time system to detect driver's fatigue, the algorithms should satisfy real-time constraints. The first step in the background subtraction method is the background modeling. In this phase several frames of the background without the face of the human operator have been used. The object detection is performed using the background model. The analysis of time complexity of this phase has been carried out with respect to the number of pixels per frame.

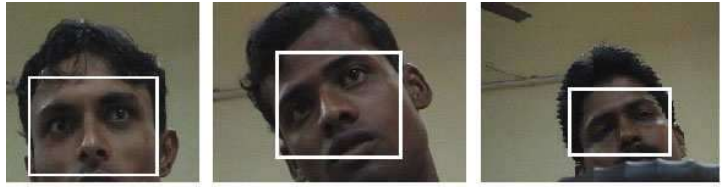

Figure 3.8: A Few Examples of True Positives

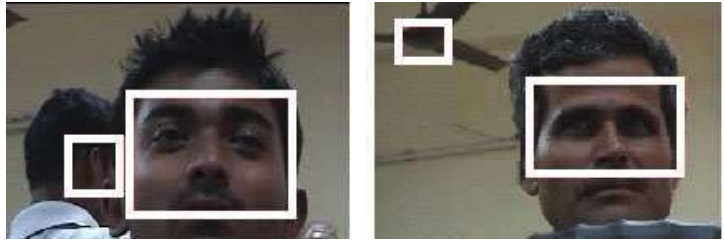

Figure 3.9: A Few Examples of False Positives along with True Positives

The time complexities of the classifiers are computed with respect to the number of pixels per frame ($N$) from the pseudo code of the algorithms [78]. From the Figure 3.10 it is evident that the method which uses the texture information for background subtraction has higher time complexity with an order of $\mathcal{O}(Nk_{mh}2^{B_{LBP}})$ , where $k_{mh}$ is the number of Local Binary Pattern (LBP) histograms and $B_{LBP}$ is the number of bits representing the LBP values. $k_{mh}$ and $B_{LBP}$ are kept at 5 and 8 respectively for the purpose. The time complexity of the method using kernel density estimation greatly depends on the number pixels per frame and also that of the initial frames ($N_{init}$) taken for background modeling. Thus it is $\mathcal{O}(NN_{init})$. The time complexity of CB is determined by both the number of pixels in each frame and the length ($L_{cb}$) of codebook. Thus it is in the order of ($\mathcal{O}(NL_{cb})$). The maximum length of codebook is the number of sample values ($N_s$) per pixel i.e. $L_{cb} \leq N_s$. The time complexity of each of the SMG and MC model based methods are found to be $\mathcal{O}(N)$. On the other hand, the upper bound of the execution time of the HLF based classifier can be computed from the pseudo code developed in algorithm 3.3.1 as $\mathcal{O}(t_c N_d \frac{N}{n_B})$. Here, $t_c$ is the constant computational time for integral image, $N_d$ is the dimension of the feature space, $N$ is the number of pixel per frame, and $n_B$ is the pixel per base window.

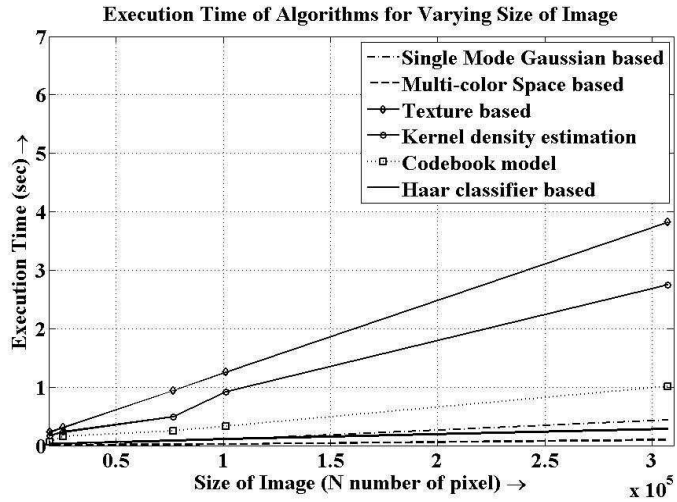

Figure 3.10: Comparison of Time Complexities of Various Face Detection Algorithms

#### 3.5.2.2 Memory Complexity

The memory complexity of each of the algorithms can also be derived from the pseudo code of the algorithms with respect to the number of pixels per frame ($N$). However, there may be algorithm specific factors that decide the memory complexity. In LBP texture model based method the number of model histograms ($k_{mh}$) and the number of histogram bins ($2^{B_{LBP}}$) also decide the complexity.

Table 3.4: Table Showing the Memory Complexity of the Classifiers

| Classification Method | Memory Complexity |
|---|---|
| Single Mode Gaussian model | $\mathcal{O}(N)$ |
| Multi color space model | $\mathcal{O}(N)$ |
| LBP texture model | $\mathcal{O}(Nk_{mh}2^{B_{LBP}})$ |
| Kernel density estimation model | $\mathcal{O}(NN_{init})$ |
| Codebook model | $\mathcal{O}(NL_{cb})$ |
| HLF based classifier | $\mathcal{O}(N_d \frac{N}{n_B})$ |

The number of initial frames ($N_{init}$) also determines the memory complexity in case of kernel density estimation based algorithm. In codebook model based method, the length ($L_{cb}$) of the codebook is another deciding parameter. The face detector based on HLF may consume memory maximally as the product of feature space dimension ($N_d$) and the number of base windows ($\frac{N}{n_B}$).

## 3.6　Eye Detection

The experimental results for face detection indicate that the HLF based classifier is more efficient than others. This is also used for eye detection. In this case, the eye images are taken as the positive example for training. The negative examples for the training are formed by the facial images without eye. There are two ways to detect eye in an image frame. It can be searched from whole image or from face region detected earlier. The later is preferable as it fast and accurate as compared to the former. The performance of the algorithm has been evaluated on the IITKGP database generated in Experiment II. The AUC of the ROC for eye detection algorithm is shown in Table 3.5. Some of the detected eyes along with face for both gray scale and NIR monochrome images are presented in Figure 3.11, 3.12, and 3.13.

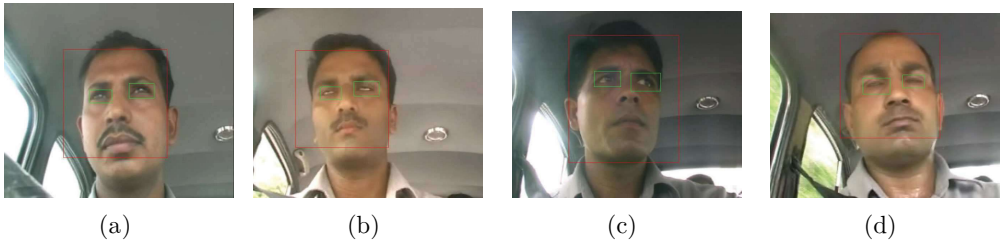

　　　　(a)　　　　　　　　(b)　　　　　　　　(c)　　　　　　　　(d)

Figure 3.11: Frame-by-frame Detection of Face and Eye in Images on Road from Experiment I of (a) Subject 2, (b)Subject 10, (c) Subject 15, and (d) Subject 21 by HLF based Classifier

Table 3.5: AUC of the Eye Detector for High and Low Illumination Level

| Classifier based on | AUC | | |
|---|---|---|---|
| | HIL | LIL | NIR |
| Haar-like feature | 0.8914 | 0.8451 | 0.9103 |

---

HIL: High Illumination Level ($\geq$ 200 Lux), LIL: Low Illumination Level ($\leq$ 50 Lux), NIR: Near Infra Red

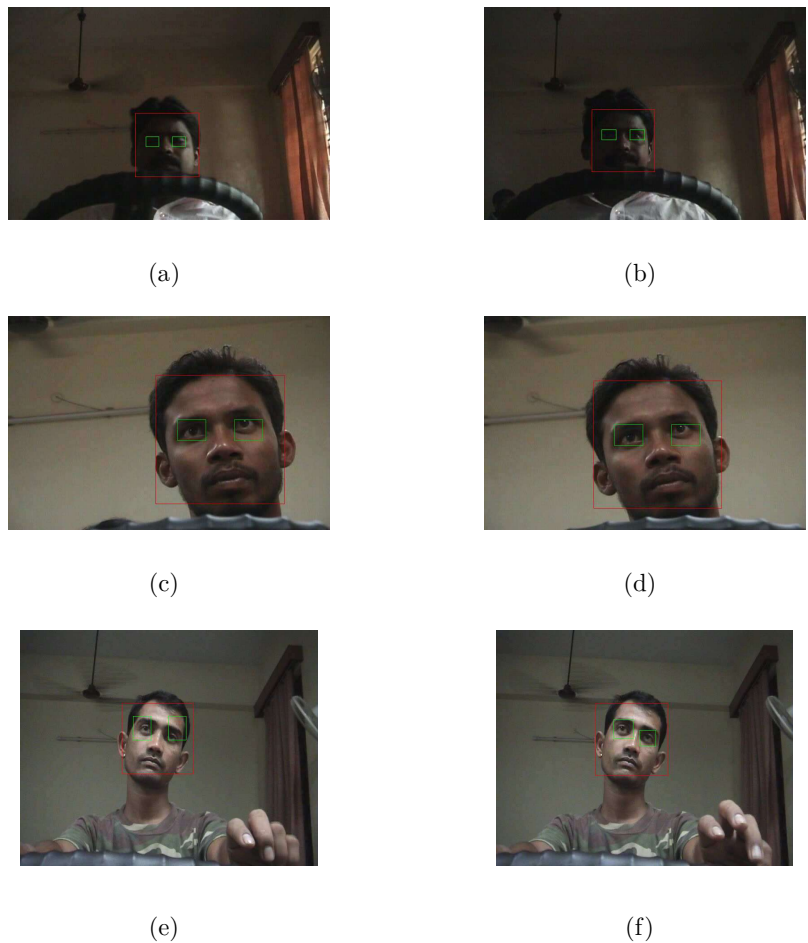

Figure 3.12: Frame-by-frame Detection of Face and Eye in Images with Varying Illumination Condition from Experiment II of (a)-(b) Subject 4 (c)-(d)Subject 5 (e)-(f) Subject 11 by HLF based Classifier

## 3.7 Conclusion

In this chapter the performance of several face detection algorithms have been evaluated. The comparative results with respect to their computational complexity and accuracy are presented. Amongst these HLF based classifier is found to be most suitable as far as the above performance criteria are concerned. Therefore, this is selected to use for eye detection from the facial images. The aim of the thesis is to measure PERCLOS and SR that are correlated to the degree of alertness. Therefore,

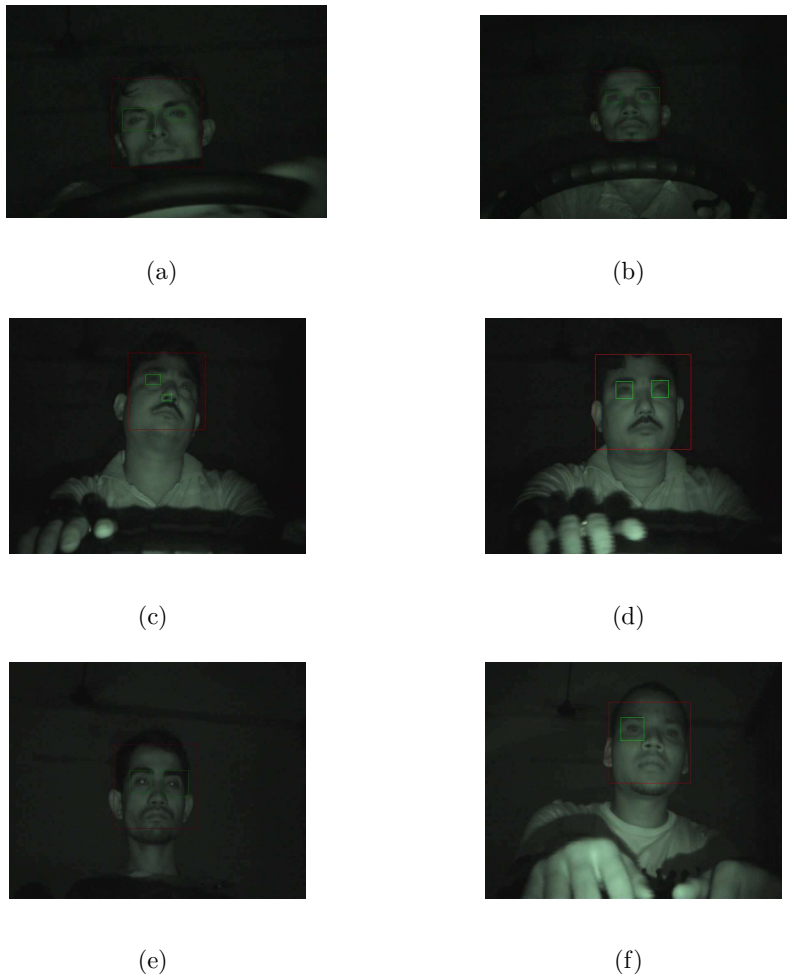

Figure 3.13: Frame-by-frame Detection of Face and Eye in NIR Monochrome Images from Experiment II of (a) Subject 1, (b)Subject 5, (c)-(d) Subject 7, (e)Subject 11, and (f) Subject 13 by HLF based Classifier

the algorithm should meet the memory as well as real-time constraints. It has been found that the HLF based algorithm may not meet the speed required to capture faster movements like eye saccade. However, it can be used to measure slow eyelid closure such as PERCLOS in real-time. The primary concern in the present context is to develop image based algorithms for accurate measurement of PERCLOS and SR and improvement of the speed of its execution can be carried out in future. Therefore, algorithms based on frame-by-frame measurement of these parameters to detect alertness in human driver have been developed in subsequent chapters.

# Chapter 4

# PERCLOS: Estimation of Eyelid Motion

## 4.1 Introduction

It is reiterated that the prime reason for most of the traffic accidents is due to reduced level of vigilance of drivers [41, 42]. The concentration of attention, perception, recognition, and vehicle-control abilities of drivers in drowsy state decrease remarkably. The judgment and skill of a driver can only be exercised so long as his/her sensory organs remain functioning, unimpaired. However, these are affected by the drowsiness of the driver. A number of investigations have been performed to find a prominent indicator of drowsiness. It is observed that slow eyelid closure is a prominent indicator of drowsiness leading to fatigue in human operators. Determination of this eyelid motion may help to measure drowsiness in the operators while they are engaged in safety critical operation. Literature suggests that Percentage Closure of Eyes (PERCLOS) over 3 minute ($P3$) may quantify this measure to a reasonable accuracy [7, 44, 79].

The technologies available in current state-of-the-art are not accurate enough to measure the level of alertness during both daylight and night conditions with equal efficiency. Some of the prior devices and systems used large hardware setups to make non-intrusive measurements while some of them involve hardware attachments to the driver leading to discomfort in driving. Some of the existing technologies involve measuring steering grip pressure to assess driver's drowsiness. This method of mea-surement is not reliable as the grip pressure changes with body movements and other behavioral patterns and hence may not be related to drowsiness. It is evident from the literature that the ocular motion based assessment of human fatigue is non-intrusive as well as reliable. During driving, the performance of this method may be affected by several factors such as variation in illumination, orientation of driver's face, relative motion between the driver and the camera etc. Hence, the estimation of the level of drowsiness may be inaccurate.

To overcome these difficulties, an algorithm for estimating the eyelid positions suitable for gray scale as well as monochrome IR images should be developed. A dual camera based measurement system has been proposed to be used one for day and the other for night driving conditions. Under daylight conditions the camera captures the gray scale image of the driver and at night (low light condition) a near-IR camera is employed with a passive near-IR illuminator for image capturing. The images are processed for detection of face and subsequently the eyes for each frame using the well-known Haar wavelets [80]. These eye images are required to be processed further for estimation of PERCLOS. In this chapter the measurement of PERCLOS has been formulated as classification of the state of eyelid position into one of the three categories: Open($0-50\%$), Closed ($90-100\%$) and Partially Closed ($60-80\%$). The class corresponding to partial closure have been used to distinguish between eye blink and the slow eye lid closure. Here, the class boundaries are obtained empirically.

The choice of classification method is determined by the accuracy and the speed of operation. One of the simplest techniques is the Eigen Space Decomposition (ESD) [81] of class instances while employing Principal Component Analysis (PCA). In this method the class variances are characterized by Eigen values in a reduced space. Thus the class mean and spread for each of the categories are obtained from some example images of eyes. A reconstruction error can be used to decide the class to which a test eye belongs. However, the classification error may be more when classes overlap in the feature space due to higher inter class correlation (as the classes are eyes with only differences in eyelid position). The choice of subspace dimension may affect the classification performance and thus the PERCLOS values.

In this context, a correlation filter based algorithm has been proposed to improve the measurement accuracy. Here, the method of synthesis of an Optimal Trade-off Maximum Average Correlation Height (OT-MACH) filter [82] in Discrete Cosine Transform(DCT) domain has been proposed to form a reference template for each of the classes. DCT domain synthesis facilitates to generate decorrelated templates better than that obtained in DFT domain [83]. A test eye is correlated with each of the templates and three parameters i.e. (i) Peak-to-Sidelobe Ratio (PSR), (ii) normalized Mutual Information (MI) between test eye and correlated output, and (iii) Fisher Ratio (FR) in transformed domain (DCT or DFT) between test image and any of the classes have been computed. The test eye is classified into one of the eye categories for which PSR and MI are maxima and FR is minimum or any two of these conditions are satisfied. The algorithm is tested on videos recorded during actual and simulated driving with varying illumination. It has also been tested under night driving conditions where a NIR camera with non-obtrusive NIR lighting scheme is used to record the image sequences of the vehicle operator.

The temporal variation of the position of eyelid over time is used to measure $P3$. The measured variation in $P3$ with different level of alertness can be judged against the subjective assessment. However, subjective feeling of the alertness level is often inaccurate and influenced by several other factors apart from drowsiness [44]. Therefore, the experimental results are compared with variations in bio-markers like blood creatinine, urea, Random Blood Sugar (RBS) etc. that are related to different state of human alertness.

## 4.2 Related Work on PERCLOS Measurement

It has been already established that PERCLOS can be effectively used as an indicator of the level of drowsiness in human beings [7,79]. In [7] PERCLOS has been measured manually by specially trained observers. However, to implement this technology in real time, automated on-board measurement is necessary, first to track the eyes and then to measure the PERCLOS.

The technology presented in [38] uses a low powered infrared (IR) based eye gaze system. The difference between pupil center and pupil corneal reflection to IR beam can be utilized to detect the pupil. However in real driving scenario due to the vibrations and random relative motion between camera and the face the IR beam cannot be focused on eyes continuously. Thus measurement of PERCLOS may be erroneous.

This technique cannot be applied always as the eyes do not exhibit good quality retinal reflections during daylight conditions.

Smith *et al.* [40] have measured the eye gaze by tracking face and eyes using color predicates. It may be used for measurement of PERCLOS only during daylight. However, this method may fail to work under low light conditions, variation of illuminations, noise, and the relative motion between driver and the imaging device. Moreover color image processing has more computational complexity than that of monochrome image processing.

Yeo [84] has developed a method for detecting the level of drowsiness in drivers by detecting an average vertical width of a driver's eye from a driver's facial image. Here, the face image is binarized for detection of eye subsequently. However, the variation in illumination may affect the binarization process. A standard vertical width and a standard drowsiness factor are computed on the basis of the average vertical width of the driver's eye wherein the average vertical width is determined by averaging the vertical width of several parts of the driver's eye over a predetermined period of time. Finally, current drowsiness factor is computed to determine the state of drowsiness at any instant by comparing it with the standards as estimated using empirical equations.

Recently literature has reported some studies on image based driver alertness monitoring algorithms and systems. A survey of these technologies has been presented in [42]. Most of these algorithms have used two cameras. However each of the algorithms as presented in [42] has to be manually initialized. Under night driving conditions the illumination inside the driver's chamber remains low. Therefore in order to make the algorithm work at night some researchers have used active IR illuminators.

The methods developed in [39] and [42] have used an active near IR illuminator with a single camera to capture driver's face under low light condition. It has utilized the 'bright and dark pupil' effect to detect pupil. The retina reflects almost all the IR light back to the camera when IR illuminator beams the light along camera's optical axis and thus the 'bright pupil' is observed in the image. The 'dark pupil' is observed when IR illuminator beams the light off the optical axis of camera. The pupil can be detected from the difference of the images thus obtained. This method may suffer from inappropriate retinal reflection during daylight and disappearance of the 'bright pupil' effect due to weather fluctuations. However it works well in controlled light condition with restricted movement of the driver's face [42].

Conventionally, minimal disturbance is allowed while carrying out measurements on the human drivers. Therefore, the researchers have so far adopted non-intrusive methods such as facial image processing for the purpose [38, 39, 40, 42]. However, they work well under certain specified conditions. Some of these techniques are suitable during daylight condition while others can only be applied during night driving situation. The drivers frequently exhibit drowsiness during night driving conditions, though the drowsiness during day driving cannot be completely ruled out [85]. Therefore, development of an image processing method which is applicable under both day as well as night driving situations seemed absolutely necessary.

## 4.3 Formulation of PERCLOS Measurement

### 4.3.1 Active Measurement

Literature survey indicates that most of the technologies for measurement of PERCLOS rely on the detection of presence of iris in the facial image by active NIR illumination technique. The iris is detected by exploiting its reflectivity to NIR beam. In these methods the face of the operator is alternately illuminated by the NIR light emitting diode of two different wavelengths. One of these NIR light beams illuminates the face along camera's optical axis while other does that off the optical axis. This active mode of NIR sources generates 'bright pupil' and 'dark pupil effect'. The pupil can be detected from the difference of the images thus obtained. This method of illumination is called active illumination and measurement of PERCLOS by this method is an active measurement. The performance of this method is degraded by several factors such as daylight, fluctuation in weather, jerks in vehicle etc. [44].

### 4.3.2 Passive Measurement

To overcome the problem associated with the active NIR illumination method, a passive illumination of face can be opted. Here, the NIR source consists of Light Emitting Diodes (LED) of single wavelength $(700-1200$ nm$)$. The LED can be kept on continuously to capture the facial image of the operator by NIR sensitive camera under dark condition. This method of lighting has been used during low light and night driving conditions.

### 4.3.3 Measurement of PERCLOS

The PERCLOS is defined as the proportion of time within a pre-defined time interval for which operator's eyes remain closed or partially closed. The pre-defined time interval may vary from 1 min to 20 min [7, 44]. Literature suggests that 20 min bout-to-bout measurement of PERCLOS value is more closely related to state of the drowsiness in human driver. However, the interval is decided empirically and can further be investigated. A PERCLOS value generated per minute by averaging it over 3 min (P3) has also been reported to show association with drowsiness. The classification output obtained per frame has been used to compute $P3$ as follows-

$$P3 = \frac{F_{pc} + F_c}{F_r \times T_F} \times 100\% \quad (4.1)$$

Where, $F_{pc}$ is number of frames in which the eyes remain 60-80% close excluding frames during blink, $F_c$ is number of frames in which the eyes remain closed fully excluding frames during blink, $F_r$ is frame grabbing speed in frames/second, $T_F$ is time factor (180 sec for $P3$). Here, the blinks are detected when open-partially closed-fully close eye sequence occurs in short time. In the following sections methods for measuring eyelid motion have been elaborated.

## 4.4 Method of Eyelid Movement Measurement

### 4.4.1 Eigen Space Decomposition based Classification

The basic concept of Eigen Space Decomposition (ESD) is to obtain separate subspace for each of the concerned class. The test object is assigned to a particular class based on a decision criterion. A number of examples of object like eye images with different eyelid positions are selected as training class. Three classes of eye images based on the degree of eyelid closure are chosen as:

1. Open$(0 - 50\%)$

2. Partially Closed (60 − 80%)
3. Closed (90 − 100%)

The training images include images with varying illumination, eye orientations etc. for each of these categories. This consideration is required to make the algorithm rotation and illumination invariant. The eigen space is reduced to a subspace keeping only the principal components for each of the classes. The test image is projected onto these subspaces and reconstructed by the corresponding eigen values. The test eye belongs to the class for which the lowest reconstruction error occurs.

#### 4.4.1.1  Computation of Eigen-eye

Eigen-eyes for each class are computed as in [81]. The steps are presented as follows:

Step-1 : For a particular class, obtain eye images $[I_1, I_2, ..., I_{N_{TR}}]$(for training), of dimension say $N \times N$.

Step-2 : Represent every image of that class as a vector $\mathbf{\Gamma}_i$ (of dimension $N^2 \times 1$).

Step-3 : Compute the average eye vector $\mathbf{\Psi}$:

$$\mathbf{\Psi} = \frac{1}{N_{TR}} \sum_{i=1}^{N_{TR}} \mathbf{\Gamma}_i \qquad (4.2)$$

Step-4 : Subtract the mean eye from each image vector $\Gamma_i$:

$$\mathbf{\Phi}_i = \mathbf{\Gamma}_i - \mathbf{\Psi} \qquad (4.3)$$

Step-5 : The covariance matrix $C$ as in (4.4) is very large.

$$\mathbf{C} = \frac{1}{N_{TR}} \sum_{i=1}^{N_{TR}} \mathbf{\Phi}_i \mathbf{\Phi}_i^T = \mathbf{A}\mathbf{A}^T, \ (N^2 \times N^2) \qquad (4.4)$$

where $\mathbf{A} = [\mathbf{\Phi}_1, \mathbf{\Phi}_2, ..., \mathbf{\Phi}_{N_{TR}}]$, ($N^2 \times N_{TR}$ matrix).
So, compute $\mathbf{A}^T\mathbf{A}$, ($N_{TR} \times N_{TR}$, $N_{TR} \ll N$) instead.

Step-6 : Compute the eigenvectors $\mathbf{v}_i$ of $\mathbf{A}^T\mathbf{A}$. Eigenvectors $\mathbf{u}_i$ of $\mathbf{A}\mathbf{A}^T$ are obtained by $\mathbf{u}_i = \mathbf{A}\mathbf{v}_i$ [81].

Step-7 : Keep only $K_E$ principal eigenvectors corresponding to the $K_E$ largest eigenvalues.

Step-8 : Normalize these $K_E$ eigenvectors

Step-9 : Repeat the above steps for other classes of eye images

#### 4.4.1.2   Testing of Eigen-eye

Step-1 : Given a test image $\mathbf{\Gamma}$, subtract the mean image of each class: $\mathbf{\Phi}_j = \mathbf{\Gamma} - \mathbf{\Psi}_j, \ j \in [1,3]$

Step-2 : Compute $\hat{\mathbf{\Phi}}_j = \sum_{i=1}^{K_E} w_i \mathbf{u}_i \ (w_i = \mathbf{u}_i^T \mathbf{\Phi}_j)$ for each class

Step-3 : Compute $e_j = \|\mathbf{\Phi} - \hat{\mathbf{\Phi}}_j\|$ for each class

Step-4 : Assign the class to the test image for which the norm $e_j$ is minimum

The performance of this method is found to be affected by several factors like noise, correlation between classes etc. It is found to be poor when closed and partially closed eyes are to be distinguished. Therefore, a robust algorithm is required to classify eye states effectively.

### 4.4.2   Correlation Filter based Classification Method

Generally in correlation based techniques a reference image of an object is correlated with the test image containing the object. If the object in the test image matches with the reference image then a correlation peak occurs at the position of the object. On the other hand if the test image does not contain the reference object no correlation peak is generated on the correlation plane. The reference image acts like a filter. The filter can be synthesized from the raw pixel values of training images. The size of the filter is same as that of training images in terms of number of pixels. Generally the images are converted from spatial domain to frequency domain. This is because of the fact that the correlation in spatial domain corresponds to multiplication of the transformed image with the complex conjugate transpose of the correlation filter in the transformed domain. The overall method is shown pictorially in Figure 4.1. The correlation outputs for a match between input and the filter, and that for an imposter are shown in Figure 4.2(a) and Figure 4.2(b) respectively.

In this work, the variation of eyelid positions into three classes (open eye, closed eye, and partially closed eye) has been characterized. Three filters (viz. correlation filter) for each of these classes have been synthesized from a number of training images. When subjected to a test eye image the output of a filter will contain a discernable peak at origin if the test eye belongs to the class specified by the filter. In this case output of each of these filters will show a peak at origin as these filters characterize three different states of the same object (eye). But the magnitudes of these peaks will be different and they can be estimated using some metrics like PSR, and Peak Correlation Energy (PCE) etc. This can be used to classify the test eye into one of these classes. Subsequently PERCLOS can be estimated by the count of the number of frames in which eyes remain partially closed within a predetermined time interval (such as 1 minute or 3 minute).

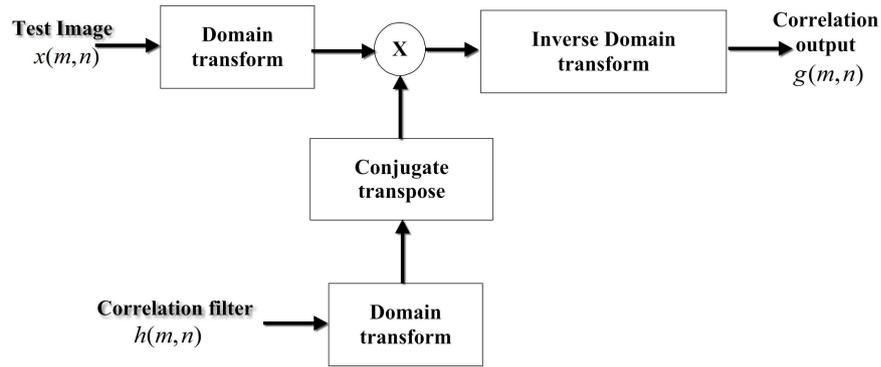

Figure 4.1: Basic Correlation Filter Technique

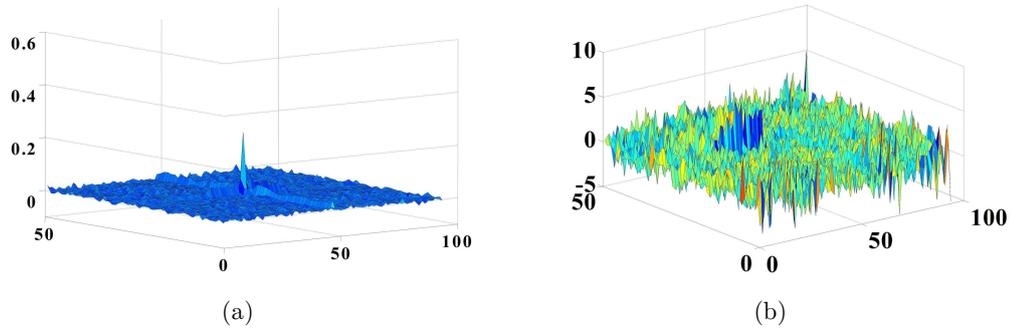

(a)  (b)

Figure 4.2: Correlation Surface for (a) True Class (b) Imposter

An exhaustive list of correlation filters can be found in [82]. After synthesizing and testing several filters the Minimum-Average-Correlation-Energy (MACE) and the OT-MACH filters turn out to be most suitable for the application. In the MACE filter, the Average Correlation Energy (ACE) is minimized with a constraint on corre-lation peak height. However, the major limitation of this kind of filter is poor built-in immunity to noise and high sensitive to intra-class variations [82].

The synthesis of OT-MACH filter is based on the optimization of three metrics. These are Average Similarity Measure (ASM), Average Correlation Height (ACH), and Output Noise Variance (ONV). The ASM is the Mean Square Error (MSE) which is a measure of distortion in the correlation surface with respect to an average shape. So minimization of ASM is required. Maximization of ACH yields high peak on the correlation surface. The effect of noise can be minimized by minimizing the ONV metric. Thus this filter provides optimum performances with respect to three major criteria simultaneously. These are the easy detection of correlation peak, good distortion tolerance, and the ability to suppress noise.

Development of this algorithm consists of two stages. The first stage being the synthesis of the filter coefficients and the second is the classification of the test images. The synthesis of these filters can be performed offline. A number of training images, recorded with all possible variations during driving, have been used to synthesize the filter. Generally filter coefficients are obtained in DFT domain as the correlation between filter and the test eye can be performed by multiplying them in this domain. However, the DFT of an image requires complex computation. To remove such complexity, it is proposed to synthesize the filter in DCT domain. The DCT operation on an image involves only real computations. Some DCT algorithms proposed recently [86, 87] can be used for the purpose. The filtering over the test eye has been performed by multiplying it with the filter in DCT domain. However, the multiplication operation in DCT domain does not correspond to exact correlation in spatial domain. The detail of this operation is presented in section 4.4.2.2 of this section. After the synthesis with extensive training classes the filters can be tested for their performance using the metrics i.e. PSR. The maximum PSR indicates the class to which the test eye belongs. An accuracy of 90% with filter synthesized by the DFT is achieved while that for the filter synthesized by DCT is found to be 94%. To improve the accuracy further, two additional features are proposed to be used as follows:

- Mutual Information (MI) between the test eye and the output of the filter.
- Fisher Ratio (FR) between the test eye and mean of each of the classes in transformed (DCT or DFT) domain.

Mutual information provides the knowledge about the modulation of input image surface by the filters. The filters modulate the input surface (eye image) by producing a peak at origin of the output plane and suppressing the other parts. Such modulation is maximum when the filter specifies the same class to which the input image belongs. On the other hand FR is a distance measurement technique for pattern classification. The training images corresponding to a particular class are clustered around the class mean. FR indicates the distance between test image and the class mean of each of this classes. This ratio is minimum for the class to which the test image belongs. It is proposed to estimate this ratio in DCT domain as it has been found that the pixels on the images are efficiently decorrelated by the DCT operation [83].

### 4.4.2.1 Synthesis of OT-MACH Filters

Design of OT-MACH filter is performed by maximizing the performance measures such as ACH, while minimizing ASM. An OT-MACH filter can be developed to meet these conflicting goals simultaneously [88]. Other performance measures, e.g., ACE and ONV are also required to be minimized for better classification in presence of noise in the background scene. The following energy function is minimized to achieve the desired filter:

$$\begin{aligned} E(h_f) &= a(ONV) + b(ACE) + c(ASM) - d(ACH) \\ &= a\mathbf{h}_f^+ \mathbf{C}_n \mathbf{h}_f + b\mathbf{h}_f^+ \mathbf{D}_x \mathbf{h}_f + c\mathbf{h}_f^+ \mathbf{S}_x \mathbf{h}_f - d|\mathbf{h}^\mathbf{T} \mathbf{m}_\mathbf{x}| \end{aligned} \quad (4.5)$$

Where $\mathbf{h}_f$ is the filter and $\mathbf{h}_f^+$ is its conjugate transpose. The OT-MACH filter (in DFT or DCT domain) is given as [82]

$$\mathbf{h}_f = (a\mathbf{C}_n + b\mathbf{D}_x + c\mathbf{S}_x)^{-1} \mathbf{m}_x \quad (4.6)$$

where $a$, $b$, $c$ and $d$ are nonnegative optimal trade-off parameters associated with $ONV$, $ACE$, $ASM$, and $ACH$. $\mathbf{m}_x$ is the average of the training image vectors $\mathbf{\Gamma}_1, \mathbf{\Gamma}_2, ...., \mathbf{\Gamma}_{N_{TR}}$ (in DFT or DCT domain). These image vectors are obtained by lexicographic ordering of 2-D images of size $M \times N$. $N_{TR}$ is the number of training images. $\mathbf{C}_n$ is the diagonal power spectral density matrix of additive input noise. Since exact knowledge about $\mathbf{C}_n$ may not be always available, white noise covariance matrix, i.e., $\mathbf{C}_n = \sigma^2 \mathbf{I}$, can be assumed in practice [88]. $\mathbf{I}$ is an identity matrix. $\mathbf{D}_x$

is an $MN \times MN$ diagonal average power spectral density of the training images:

$$\mathbf{D}_x = \frac{1}{N_{TR}} \sum_{i=1}^{N_{TR}} \mathbf{X}_{Di}^* \mathbf{X}_{Di} \qquad (4.7)$$

where $\mathbf{X}_{Di}$ is an $MN \times MN$ diagonal matrix of training image. $\mathbf{S}_x$ denotes the similarity matrix of the training images of size $MN \times MN$:

$$\mathbf{S}_x = \frac{1}{N_{TR}} \sum_{i=1}^{N_{TR}} (\mathbf{X}_{Di} - \mathbf{M}_x)^* (\mathbf{X}_{Di} - \mathbf{M}_x) \qquad (4.8)$$

Where $\mathbf{M}_x$ is the average of $\mathbf{X}_{Di}$. The OT-MACH filter in (4.5) can behave as minimum variance synthetic discriminant function (MVSDF) filter when $b = c = 0$ [82]. The same filter equation can be used as MACE filter with $a = c = 0$. The MVSDF filter provides relatively good tolerance against noise but with broad peaks. On the other hand MACE filter, which generally produces sharp peaks and good noise suppression, is very sensitive to distortion [88]. For low value of $a$ and $b$ the filter is a MACH filter which is designed with high tolerance for distortion. So it is evident that the choice of $a$, $b$ and $c$ determines the nature of the filter. The optimal values of the optimal trade-off parameters can be obtained by Monte Carlo simulation. In this application, $a = 0.1$, $b = 0.2$, and $c = 0.7$ have been found to generate best results. As $c$ is larger than $a$ and $b$ the resulting filter is a MACH filter.

#### 4.4.2.2    Testing of OT-MACH Filters

The filter coefficients obtained by (4.6) can be used for real-time classification of test images. Testing is performed by 2-D correlation between test image and the filter. Initially the test image is transformed into DFT domain. Then transformed image is multiplied with the filter coefficients. Finally the result is transformed back into spatial domain by Inverse DFT (IDFT). The output is processed further to estimate different features for classification.

This operation is equivalent to the circular autocorrelation operation which is a time aliased version of linear correlation operation. This aliasing can be ignored when the filter corresponds to the class to which the test image belongs [82]. Moreover, the advantage of such aliasing can be used when the filter differs from the class to which the test image belongs, as this is a cross correlation operation [82]. The output correlation plane obtained using DFT and IDFT can be represented as:

$$G(i,j) = I(i,j) \odot H_f(i,j) = IDFT(I^{dft} * H_f^{dft}) \qquad (4.9)$$

where, the symbol $\odot$ denotes circular correlation and the symbol $*$ denotes multiplication. $G(i,j)$ is the correlation surface. $I(i,j)$ is the test image. $H_f(i,j)$ is the filter coefficients in 2-D spatial domain. $I^{dft}$ and $H_f^{dft}$ are the DFT of test image and the filter respectively.

It is proposed to use 2-D DCT and 2-D Inverse DCT (IDCT) operations for synthesis of these filters and classification. Initially the motivation to use DCT was to eliminate the complex computations required for DFT and IDFT operations. Moreover, the advantage of fast DCT computation without the use of DFT can also be utilized [89]. The synthesis of OT-MACH filter is performed as stated earlier with all image data transformed into DCT domain. The test image is classified by post processing the output surface as:

$$G'(i,j) = IDCT(I^{dct} * H_f^{dct}) \qquad (4.10)$$

where $G'(i,j)$ is the output surface. $I^{dct}$ and $H_f^{dct}$ are the DCT of test image and the filter respectively.

The operation in (4.10) is not the exact circular correlation operation as the case in (4.9). The operation in spatial domain corresponding to that in (4.10) can be presented as [90, 91]:

$$G''(i,j) = \hat{I}(i,j) \otimes \hat{H}_f(i,j) \otimes Z(i,j) \qquad (4.11)$$

where the symbol $\otimes$ denotes circular convolution. $\hat{I}(i,j)$ is the symmetrically extended (about $(-\frac{1}{2}, -\frac{1}{2})$) version of test image i.e.

$$\begin{aligned}\hat{I}(i,j) &= x_{m,n}, \qquad \forall i \in [0, M-1], j \in [0, N-1] \\ &= I(1-i, 1-j), \quad \forall i \in [-M, -1], j \in [-N, -1]\end{aligned} \qquad (4.12)$$

where $M$ is the number of rows and $N$ is the number of columns of the image $I(i,j)$.

$\hat{H}_f(i,j)$ is the symmetrically extended (about $(-\frac{1}{2}, -\frac{1}{2})$) version of the filter and can be defined similarly as in (4.12). $Z(i,j)$ is a modulating surface symmetrical about $(\frac{1}{2}, \frac{1}{2})$ and can be defined as-

$$Z(i,j) = 16 \left\{ \left[\frac{1}{2\sqrt{2}} - 1\right] + cos\left[\frac{\pi}{4M}(2i-1)(M-1)\right] \frac{sin\left[\frac{\pi}{4}(2i-1)\right]}{sin\left[\frac{\pi}{4M}(2i-1)\right]} \right\} \\ \times \left\{ \left[\frac{1}{2\sqrt{2}} - 1\right] + cos\left[\frac{\pi}{4N}(2j-1)(N-1)\right] \frac{sin\left[\frac{\pi}{4}(2j-1)\right]}{sin\left[\frac{\pi}{4N}(2j-1)\right]} \right\} \quad (4.13)$$

$Z(i,j)$ shifts the point of symmetry of the surface obtained from the operation $\hat{I}(i,j) \otimes \hat{H}_f(i,j)$ by $(\frac{1}{2}, \frac{1}{2})$.

The relation between $G'(i,j)$ in (4.10) and $G''(i,j)$ in (4.11) is given as-

$$G'(i,j) = G''(i,j) \quad \forall i \in [0, M-1], j \in [0, N-1] \quad (4.14)$$

It has been found that peaks occur at four points $((-1,-1), (-1,0), (0,0), (0,-1))$ of $G''(i,j)$ surface. This is because the test image and the filter are symmetrically extended prior to circular convolution operation. The circular convolution of symmetrically extended signal is flipped version of circular correlation operation. So, in effect the correlation plane is obtained by the operation in (4.10). The relation between the $G'(i,j)$ and $G''(i,j)$ surfaces reveals that the peak occurs at the origin of $G'(i,j)$. There will be higher PSR values when the filter corresponds to the same class to which the test image belongs.

Testing of the algorithm has been performed on the recorded video of driving operation. The face and eyes of the driver have been detected from the video frames using Haar classifiers [67, 80, 92]. Subsequently the eyes are classified into one of these three classes. The DFT and DCT based filters are tested separately. Initially, the PSR (or PCE) value of the correlation plane (as defined in (4.9) and (4.10) for each of the filters) is considered as the classification parameter. The eye will be considered to belong to a particular class for which the PSR (or PCE) magnitude is the highest.

Thus the state of eyelid of the driver can be detected effectively.

#### 4.4.2.3 Classification Parameter

**Peak-to-Sidelobe Ratio (PSR):** The PSR can be defined in different ways [82]. First, the correlation peak is searched in the correlation plane. A region of size $20 \times 20$ excluding a small region of size $5 \times 5$ around the peak (shown in Figure 4.3) has been considered for the computation of mean ($\mu$) and Standard Deviation ($\sigma$). These values are used to define PSR as shown in the following equation [82].

$$PSR = \frac{(peak - \mu)}{\sigma} \qquad (4.15)$$

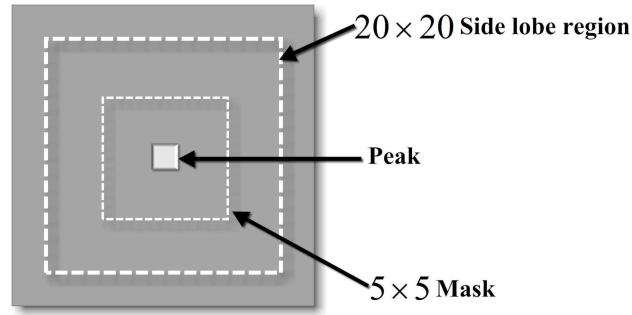

Figure 4.3: Region for PSR Computation

It is observed that the test accuracy of DFT-based OT-MACH filter technique is 90%, while that for DCT based technique turns out to be 94% when only PSR is considered for classification. The correlation plane can be further processed to improve the accuracy using multiple criteria for classification. Two parameters have been chosen in addition to PSR.

**Mutual Information (MI):** The normalized mutual information between the test image and the correlation plane has been used as one of the classification parameters. The value of MI is maximum when the test surface is modulated maximally by the filter. This is the case when the filter corresponds to the same class to which the test image belongs. Therefore the test image can be classified to a particular class for which the MI is highest.

The mutual information between two images $I_1(i,j)$ and $I_2(i,j)$ are computed from the joint intensity distribution, $(P(I_1(i,j), I_2(i,j)))$ which is closely related to joint entropy, $(H_{JE}(I_1(i,j), I_2(i,j)))$ [93]. The normalized mutual information is defined as:

$$MI(I_1(i,j), I_2(i,j)) = \frac{H_{ME}(I_1(i,j)) + H_{ME}(I_2(i,j))}{H_{JE}(I_1(i,j), I_2(i,j))} \quad (4.16)$$

Where the joint entropy is defined as

$$H_{JE}(I_1(i,j), I_2(i,j)) = - \sum_{I_1(i,j), I_2(i,j)} P(I_1(i,j), I_2(i,j)) \ln(P(I_1(i,j), I_2(i,j))) \quad (4.17)$$

and the marginal entropies are

$$H_{ME}(I_c(i,j)) = - \sum_{I_c(i,j)} P(I_i(i,j)) \ln(P(I_c(i,j))) \text{ for } c = 1,2 \quad (4.18)$$

and (**ln**) is the natural logarithm.

**Fisher Ratio (FR):** To improve the accuracy further a widely accepted statistical pattern recognition metric Fisher ratio (FR) has been employed. The computation of such feature is very simple and thus computational complexity does not increase much. The Fisher ratio can be defined as the squared difference between means of two classes divided by the average of the variances of the classes [82].

$$FR = \frac{(\mu_{c1} - \mu_{c2})^2}{\sigma_1^2 + \sigma_2^2} \quad (4.19)$$

where $\mu_{c1}$ and $\mu_{c2}$ are the class means of two classes and $\sigma_1^2$ and $\sigma_2^2$ are the corresponding class variances.

During synthesis of the filters mean and variance of each class has been computed in transformed domain (i.e. in DFT or DCT domain). Fisher ratio between test image and each of the classes can be computed as per (4.19). Here, $\mu_{c2}$ is replaced by the transformed test image and variance ($\sigma_2^2$) is assumed to be zero.

#### 4.4.2.4 Decorrelation Property of Transformations

The decorrelation property of DCT may be better than that of the DFT for Markov-1 sequence. The off diagonal elements of covariance matrix of the Markov-1 sequence in transformed domain may be considered for the purpose. The proof is provided for 1-D Markov-1 sequence. Since the DFT and DCT are separable operations, so the proof can be extended for 2-D images (such as eye image in current context).

**Proof**: Let the covariance matrix in sequence domain is $\mathbf{C}_u$. For Markov-1 sequence, $\mathbf{C}_u$ can be expressed as-

$$\mathbf{C}_u = \begin{bmatrix} 1 & \rho & \rho^2 & \cdots & \cdots & \rho^{N_{ld}-1} \\ \rho & \ddots & \ddots & \ddots & \ddots & \vdots \\ \vdots & \ddots & \ddots & \ddots & \ddots & \vdots \\ \vdots & \ddots & \ddots & \ddots & \ddots & \rho^2 \\ \vdots & \ddots & \ddots & \ddots & \ddots & \rho \\ \rho^{N_{ld}-1} & \cdots & \cdots & \cdots & \rho & 1 \end{bmatrix} \quad (4.20)$$

Where the adjacent correlation coefficient between consecutive data is $0 \leq \rho \leq 1$ and $N_{ld}$ is the length of data sequence [83]. For highly correlated data points $\rho$ is close to 1.

The DFT and DCT operation can be expressed in matrix form as below -

$$\mathbf{A}_{dft} = \frac{1}{\sqrt{N_{ld}}} \begin{bmatrix} 1 & 1 & 1 & \cdots & \cdots & 1 \\ 1 & W & W^2 & \cdots & \cdots & W^{N_{ld}-1} \\ 1 & W^2 & W^4 & \cdots & \cdots & W^{2(N_{ld}-1)} \\ \vdots & \vdots & \vdots & \ddots & \ddots & \vdots \\ \vdots & \vdots & \vdots & \ddots & \ddots & \vdots \\ 1 & W^{N_{ld}-1} & \cdots & \cdots & \cdots & W^{(N_{ld}-1)(N_{ld}-1)} \end{bmatrix} \quad (4.21)$$

Where $W = e^{-j\frac{\pi}{N_{ld}}}$ and the scalar $\frac{1}{\sqrt{N_{ld}}}$ is multiplied to make it as unitary transform i.e. $\mathbf{A}_{dft}\mathbf{A}_{dft}^{*T} = \mathbf{I}$; $\mathbf{I}$ is an identity matrix of size $N_{ld} \times N_{ld}$ and $*T$ denotes the complex conjugate transpose operation.

$$\mathbf{A}_{dct} = \sqrt{\frac{2}{N_{ld}}} \begin{bmatrix} \frac{1}{\sqrt{2}} & \frac{1}{\sqrt{2}} & \cdots & \frac{1}{\sqrt{2}} \\ cos(\theta) & cos(3\theta) & \cdots & cos((2N_{ld}-1)\theta) \\ cos(2\theta) & cos(6\theta) & \cdots & cos((4N_{ld}-2)\theta) \\ \vdots & \vdots & \ddots & \vdots \\ cos((N_{ld}-1)\theta) & cos((3N_{ld}-3)\theta) & \cdots & cos((2N_{ld}-1)(N_{ld}-1)\theta) \end{bmatrix}$$
(4.22)

Where $\theta = \frac{\pi}{2N_{ld}}$. The covariance matrix in transformed domain can be obtained by (4.23) and (4.24) for DFT and DCT transform respectively [94].

$$\mathbf{C}_{vdft} = \mathbf{A}_{dft}\mathbf{C}_u\mathbf{A}_{dft}^{*T} \tag{4.23}$$

$$\mathbf{C}_{vdct} = \mathbf{A}_{dct}\mathbf{C}_u\mathbf{A}_{dct}^{*T} \tag{4.24}$$

The diagonal element of the covariance matrix denotes the variance ($\sigma_{i,i}^2$) of the corresponding transform coefficient. The off diagonal element ($(i,j); \forall i \neq j$) of the covariance matrix in transform domain corresponds to the expectation of cross correlation between $i^{\text{th}}$ and $j^{\text{th}}$ coefficients. From (4.23) and (4.24), the absolute value of adjacent correlation coefficients $|\rho_\nu(i,j)|$ can be found. The general expression of $|\rho_\nu(i,j)|$ is given in (4.25) [94]. Thus, $|\rho_\nu(i,j)|$ in both DFT and DCT domain can be computed for a given $\rho$ in sequence domain.

$$|\rho_\nu(i,j)| \triangleq \left|\frac{\mathbf{C}_\nu(i,j)}{\sigma_{i,i}\sigma_{j,j}}\right| = \left|\frac{\mathbf{C}_\nu(i,j)}{\sqrt{\mathbf{C}_\nu(i,i)\mathbf{C}_\nu(j,j)}}\right| \tag{4.25}$$

The surface plot of $|\rho_\nu(i,j)|$ in DFT and DCT domain are shown in Figure 4.4 and Figure 4.5 respectively. Here the adjacent correlation coefficient ($\rho$) in sequence domain is 0.99 and the sequence length ($N_{ld}$) is assumed to be 64 which is equal to the width of the eye image used in our application.

The Figure 4.4 and Figure 4.5 show that the off diagonal elements of covariance matrix are very less in DCT domain compared to that in DFT domain. Thus the decorrelation power of DCT is better than that of DFT for Markov-1 sequence.

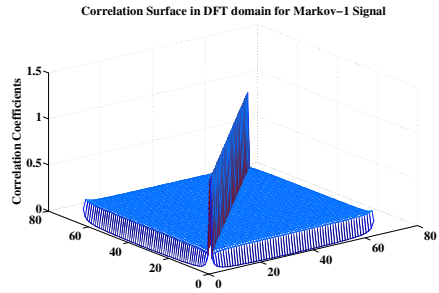

Figure 4.4: Correlation Coefficient Plot in DFT Domain

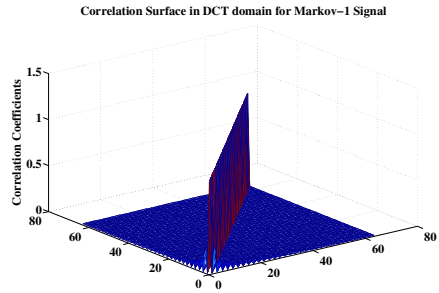

Figure 4.5: Correlation Coefficient Plot in DCT Domain

## 4.5 Experiment and Performance Measurement

The aim of this work has been to capture the temporal variation of eyelid position for measuring PERCLOS. In this context, a method of synthesis of an Optimal Trade-off Maximum Average Correlation Height (OT-MACH) filter [82] in DCT domain has been proposed to form a reference template for each of the classes. DCT domain synthesis facilitates to generate decorrelated templates better than that obtained in DFT domain. A test eye is correlated with each of the templates and three parameters (i) PSR, (ii) normalized MI between test eye and correlated output, and (iii) FR in transformed domain (DCT or DFT) between test image and any of the classes have been computed. The test eye is classified into one of the eye categories for which PSR and MI are maximum and FR is minimum or any two of these conditions are satisfied. The Receiver's Operating Characteristics (ROC) curve (plot of True Positive Rate (TPR) vs. False Positive Rate (FPR)) of the classifier synthesized in DCT domain with various training data set and deciding criteria have been shown in Fig. 4.6. The results are also compared with those obtained from the filter synthesized in Fourier domain in Table 4.1-4.2. The algorithm can be applied on both gray scale as well as NIR monochrome eye images. The experimental results indicate that OT-MACH filter synthesized in DCT domain performs better than that in DFT domain.

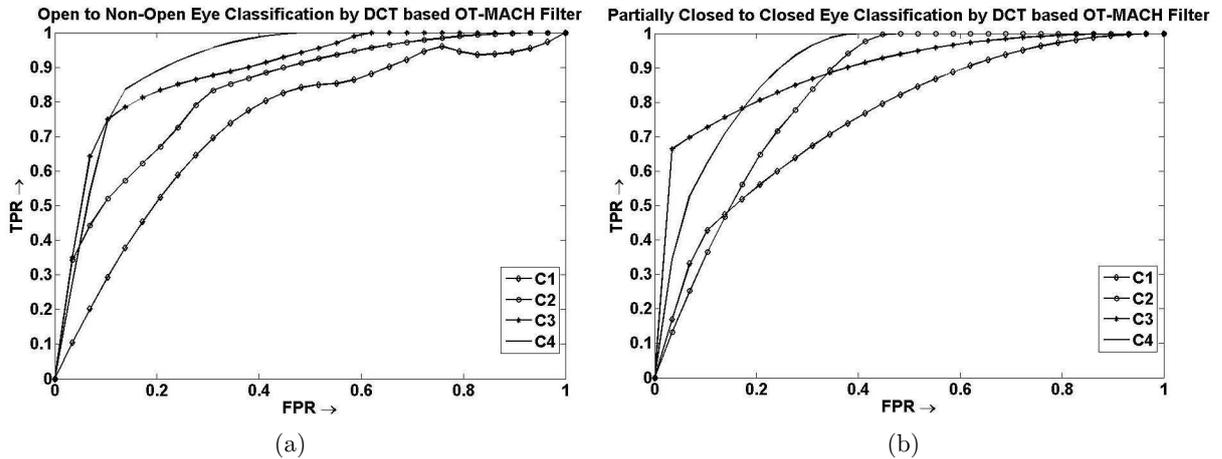

Figure 4.6: ROC Curve for (a) Open Eye to Non-open Eye, and (b) Partially Closed to Closed Eye Classification with Correlation Filter Synthesized in DCT Domain

Part of Training data from whole database C1: 0% with all criteria, C2: 20% with only PSR criterion, C3: 20% with all criteria, C4: 30% with all criteria

Table 4.1: Open to Non-open Eye Classification Performance based on Area Under (AUC) the ROC Curves

| Domain of Synthesis | Proportion of Training Data from Whole Database | | | |
|---|---|---|---|---|
| | 0% (C1) | 20% with decision criteria | | 30% (C4) |
| | | PSR only (C2) | All (C3) | |
| DCT | 0.732 | 0.831 | 0.889 | 0.912 |
| FFT | 0.690 | 0.732 | 0.857 | 0.832 |

A compact, non-intrusive and low cost real-time on-board driver drowsiness monitoring system has been proposed based on the above mentioned technique. The system detects eyes from facial image captured using a) gray scale Charge Coupled Device (CCD) camera during daylight and b) NIR (700-1200 nm) sensitive monochrome CCD camera combined with compact passive NIR illuminator during night. It also

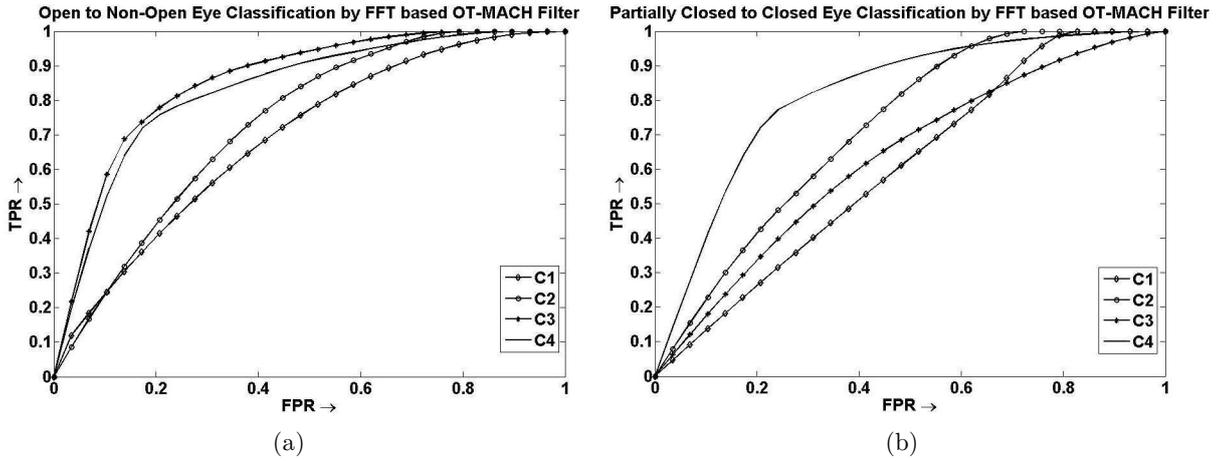

Figure 4.7: ROC Curve for (a) Open Eye to Non-open Eye, and (b) Partially Closed to Closed Eye Classification with Correlation Filter Synthesized in DFT Domain

Part of Training data from whole database C1: 0% with all criteria, C2: 20% with only PSR criterion, C3: 20% with all criteria, C4: 30% with all criteria includes a light intensity measuring unit to activate the required camera based on the illumination level. The embedded software can process both the gray scale and NIR monochrome eye images to assess the level of drowsiness of human vehicle driver through day and night. The PERCLOS has been computed from the variations in eyelid position per minute as a 3 minute moving average using the OT-MACH correlation filter synthesized in DCT domain.

Table 4.2: Partially Closed to Closed Eye Classification Performance based on AUC of ROC Curves

| Domain of Synthesis | Proportion of Training Data from Whole Database | | | |
|---|---|---|---|---|
| | 0% (C1) | 20% with decision criteria | | 30% (C4) |
| | | PSR only (C2) | All (C3) | |
| DCT | 0.757 | 0.830 | 0.893 | 0.902 |
| FFT | 0.606 | 0.620 | 0.633 | 0.816 |

The experimental results on Eigen space decomposition based eye classification technique for estimation of PERCLOS reveals that it may fail if the inter variance between classes are less. Moreover, the space is actually the variance space. Thus AUC of the ROC for different conditions is poor particularly for partially closed and closed eye. Points to be mentioned here, three classes of eye with different eyelid positions are considered to eliminate the blinking instances from the slow and longer eyelid closure.

## 4.6  Correlation Analysis with Drowsiness Bio-Markers

A few temporal variations of $P3$ have been shown in Figure 4.8 for Experiment-I and in Figure 4.9 for Experiment-II. The output of the algorithm is started from the 3rd minute and continued up to the duration of each stage in respective experiment per subject.

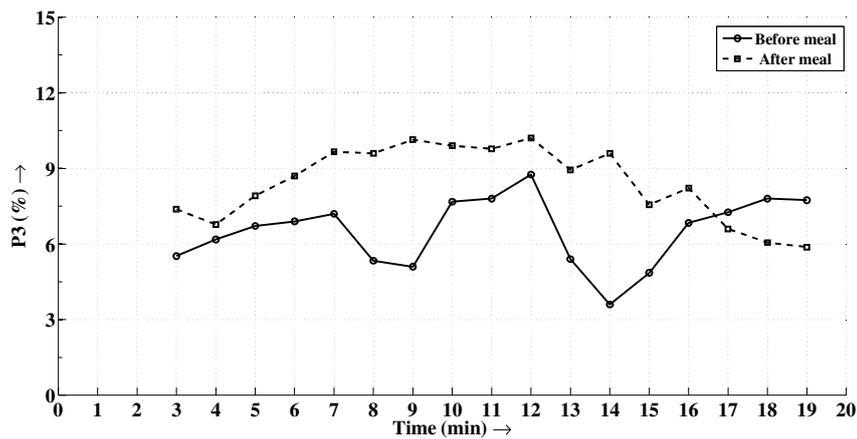

Figure 4.8: Result Showing Variation of $P3$ of Subject 21 of Experiment-I

Figure 4.10 shows the PERCLOS values of 12 subjects for five stages with a separation of 8 hours between stages from Experiment II. Overlap of PERCLOS values can be observed at lower states of drowsiness due to several other factors apart from drowsiness.

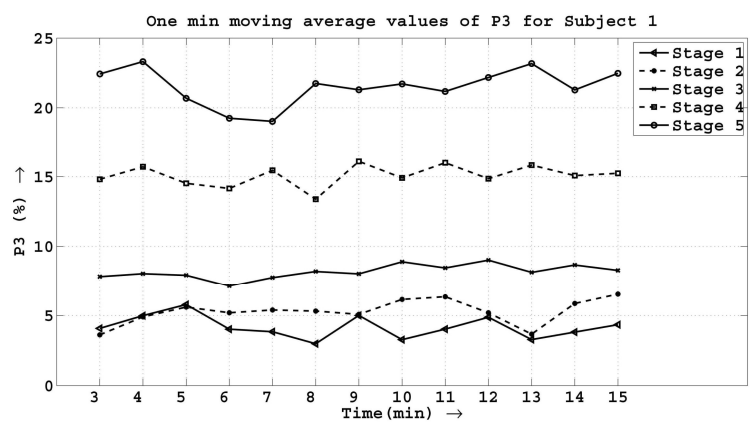

(a)

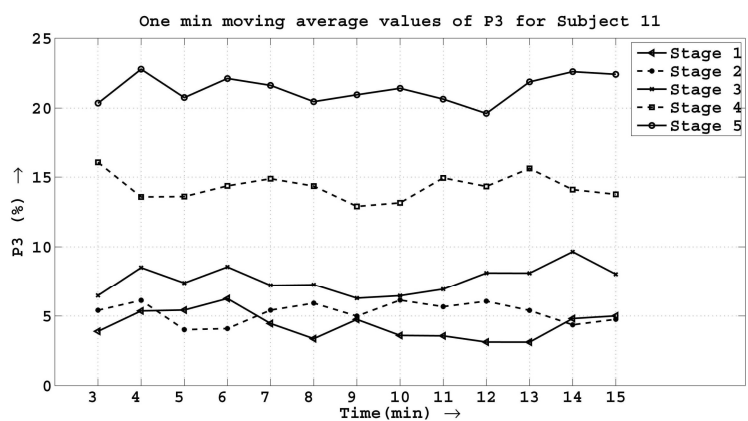

(b)

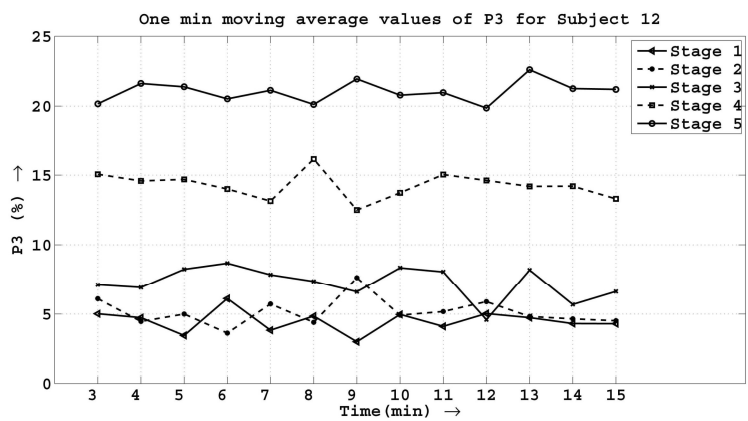

(c)

Figure 4.9: Results Showing Variation of $P3$ of (a) Subject 1, (b) Subject 11, and (c) Subject 12 of Experiment-II

The average of $P3$ over 15 min are computed and presented in Figure 4.10. The variations in these values with different levels of alertness and among subjects are clearly visible from the figure.

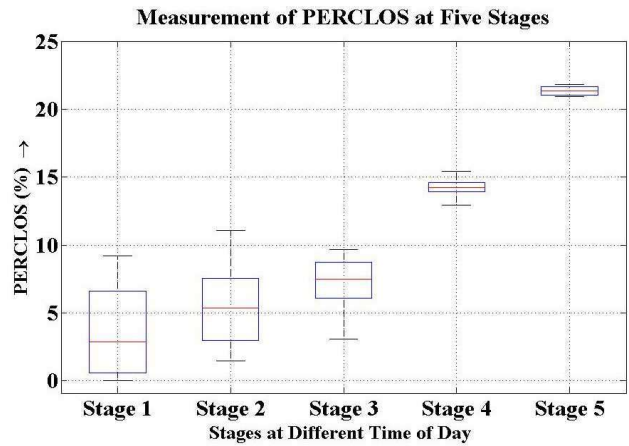

Figure 4.10: Results Showing Inter Subject Variation of PERCLOS at Five Stages at Different Time of Day

The image-based method of estimation of PERCLOS is required to be validated by the measurement of more direct index of alertness level. Literature suggests that amount of various neurotransmitters and their metabolic products reflect the level of central as well as peripheral fatigue [95]. In the present context the serum creatinine and urea, and RBS have been analyzed. A correlation analysis has been carried out for the average $P3$ values over 15 min (duration of video recording in each stage) per subject for five stages in Experiment-II.

The level of RBS indicates the rate of oxidative metabolism of live cells. It is evident that increasing demand of cell energy with decrease in alertness requires more glucose. Thus glucose production from non-carbohydrate substrates (gluconeogenesis) is increased in the body leading to hyperglycemia. Creatinine is a chemical waste molecule that is generated from muscle metabolism. It has been found to be a fairly reliable indicator of kidney function as well as muscle degradation due to overuse. Normal levels of creatinine in the blood are approximately $0.6 - 1.2$ mg/dL in adult males and $0.5 - 1.1$ mg/dL in adult females. Any condition that impairs the function of the kidneys or increase the physiological load to kidney will probably raise the blood creatinine level [96]. It is also found that the Blood Urea Nitrogen (BUN) test is a measure of the amount of nitrogen in the blood in the form of urea, and a measurement of renal function. Urea is secreted by the liver, and removed from the blood by the kidneys. The liver produces urea in the urea cycle as a waste product of the digestion of protein. Normal adult blood should contain between $7-21$ mg of urea nitrogen per 100 ml of blood. Bio-chemical analysis of these blood parameters (mean and standard error of all subjects) at five stages have been shown in Figure 4.11. The blood glucose level is seen to have increased gradually at successive stages in most of the subjects. The result also showed an increasing trend of serum creatinine levels with decrease in the level of alertness.

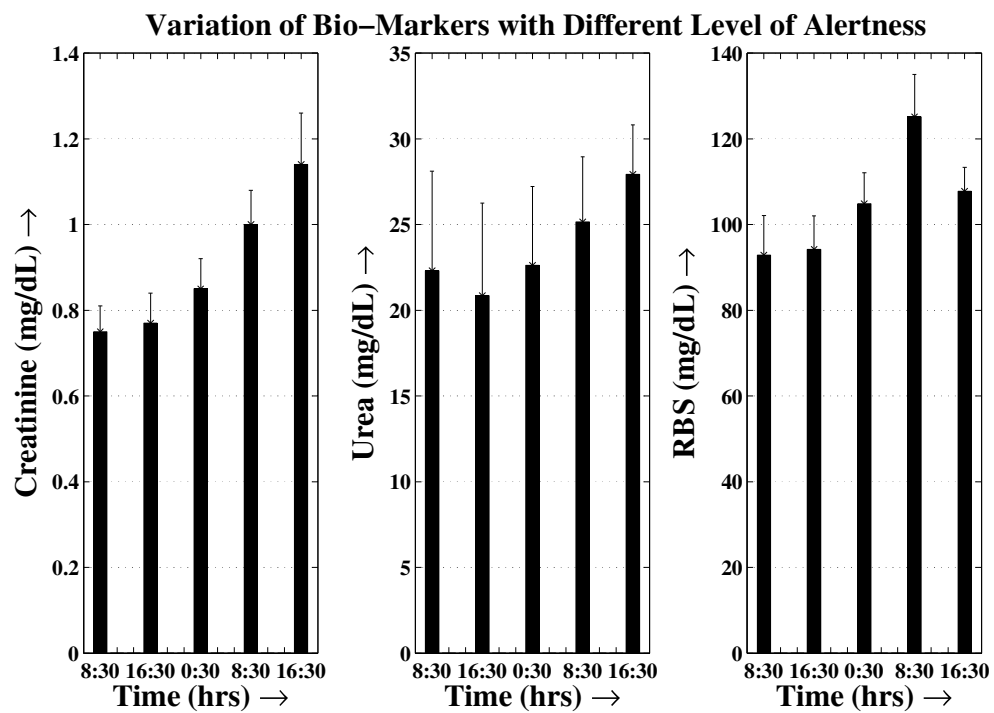

Figure 4.11: Variations of Serum Creatinine, Serum Urea, and RBS for 12 Subjects at 5 Different Stages of Experiment-II

There is a regular increase of serum urea levels as shown in Figure 4.11 and BUN for all 12 subjects from second stage in the course of experiment. An increased protein catabolism is observed to supply demand of energy with increasing level of sleep deprivation accompanied by exercise. This fact is also reported in [97]. Thus, gradual increment in sleep deprivation along with monotonous activities that affect the body metabolism can be considered as a reliable indicator of initial as well as advanced state of alertness. Correlation of the inter-subject variations in creatinine, urea and RBS with the trends in average of $P3$ over 15 min are computed separately for 5 stages of the experiment. The results are presented in Table 4.3. It shows high correlation between $P3$ and creatinine, $P3$ and urea and a moderate correlation between $P3$ and RBS. The bio-markers like serum creatinine and urea are found to be highly correlated with drowsiness.

Table 4.3: Correlation Values with $P3$. The Values Indicate the Correlation Coefficient and Level of Significance ($p$-Value Inside the Parenthesis)

| Parameters | Serum Creatinine | Serum Urea | Blood RBS |
|---|---|---|---|
| **P3** | 0.9974(0.00016) | 0.9670(0.00716) | 0.6491(0.23591) |

## 4.7 Real Time Implementation of PERCLOS Measurement

The method has been extended to design a system for online measurement of PERCLOS for human driver. The system comprises an innovative combination of gray scale CCD camera for imaging during daylight and an NIR ($700 - 1200$ nm) sensitive monochrome CCD camera combined with compact passive NIR illuminator for imaging during night. A light intensity measuring unit is connected to activate any of said CCD cameras based on level of illumination. The NIR illuminator (shown in Figure 4.12) has been developed for illuminating the face during dark condition and an NIR sensitive camera was used to capture facial images.

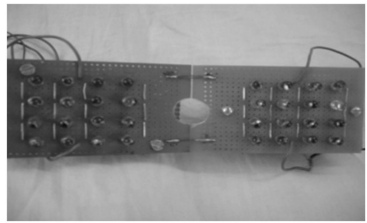

Figure 4.12: System for Passive NIR Illuminator

The drowsiness assessment unit consists of a processing unit with windows operating system to classify ocular images using OT-MACH correlation filter synthesized in DCT domain. The state of eyelid position produced by the unit is used to compute $P3$ value as explained in section 4.3.3 and thresholds are applied to determine the current level of drowsiness.

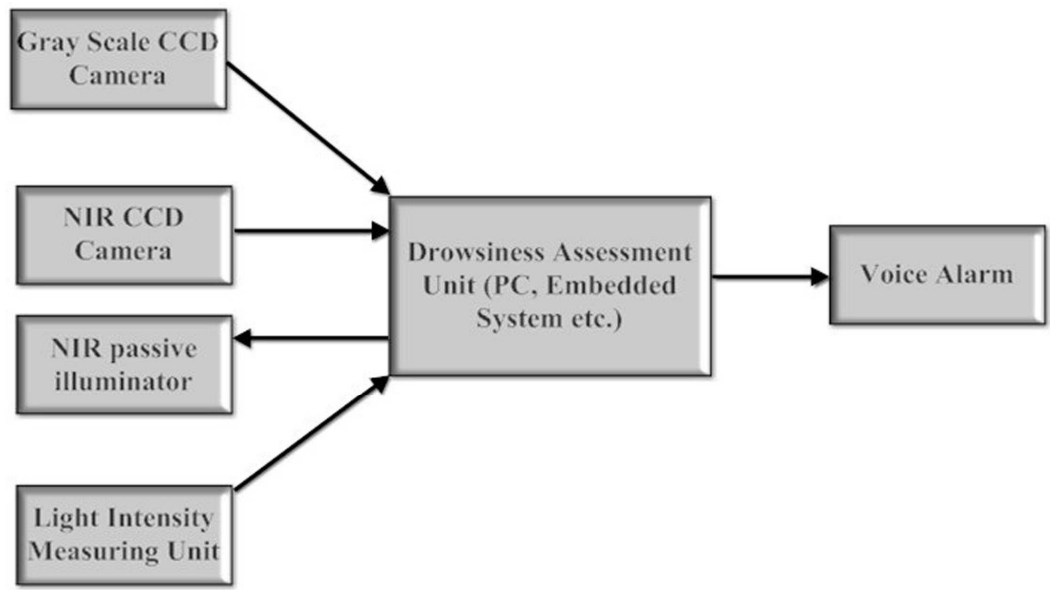

Figure 4.13: Block Diagram of Drowsiness Detection System

The classification algorithm is coded in C on Visual C 6.0 platform with some of the functions from OpenCV library. The program is executed on 3.00 GHz Pentium-D processor platform with Windows-Vista as the operating system. The Random Access Memory (RAM) size is 1GB. The average speed of the algorithm has been found to be 16fps. In the context of detection of driver fatigue by measurement of PERCLOS over time this speed of operation is satisfactory. The speed of operation is directly related to the size of the image and the filters. From the recorded video of on road driving it has been found that the size of the extracted eyes varies around pixels which is also the size of the training images considered. The smaller size of images leads to faster execution of the algorithm. The proposed algorithm is explained through flow chart as shown in Figure 4.14

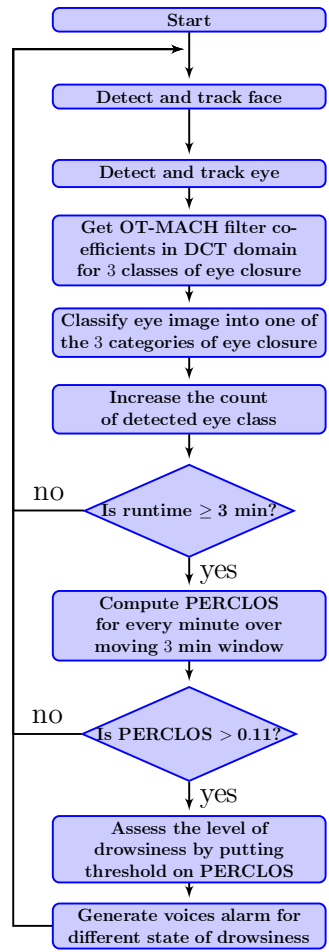

Figure 4.14: Algorithm for Drowsiness Detection

## 4.8　Conclusion

In this chapter, the measurement of PERCLOS is formulated as classification of eye images into three classes of different eyelid positions. Since the inter variance of the classes are close enough the classification performance by Eigen space decomposition based method is found to be poor. An OT-MACH correlation filter based algorithm for measurement of PERCLOS has been developed. Synthesis and testing of the filter is proposed to be carried out in DCT domain to get the benefit of decorrelation property. The experimental outcome indicates that DCT based operation for this filter outperforms that obtained with FFT based method. The same algorithm can be applied during night on the NIR monochrome eye images. During night driving the face of the operator is illuminated passively to obtain the images. Thus the problem associated with active NIR illumination based method can be avoided. The color image processing is not also feasible during night driving condition. The proposed system based on OT-MACH correlation filter based method synthesized in DCT domain has been used to measure PERCLOS. It can be observed from the Figure 4.10 that wide inter personal variations in the PERCLOS values at lower states of drowsiness exist. It indicates to use other indices of alertness such as SR at these levels.

## Chapter 5

# Form Factor: A New Image Feature for Pupil Motion Detection

## 5.1 Introduction

Literature suggests that Saccadic Ratio (SR) provides information related to human alertness level [54]. Eye saccade is the simultaneous and quick horizontal movement of iris of both eyes in same direction. Estimation of this ratio requires to find the position of the iris center in the eye images. Studies show that partial occlusion of iris by eyelid, variation in illumination, low resolution, noise etc. may pose difficulties in finding the iris center as well as other parameters. A new image feature called Form Factor (FF) has been introduced to estimate ocular features and to measure SR effectively. FF is defined as the ratio of Root Mean Square (R.M.S.) to average of a set of pixel values. In this chapter, the methods for edge detection and determination of eye center and pupil diameter by this feature have been developed and presented as two case studies. The edge detection method can be applied to eye images for corner detection. Blind estimation of Signal-to-Noise Ratio (SNR) or Signal-to-Total-Variance Ratio (STVR) is investigated for recovery of FF under zero-mean independently and identically distributed (i.i.d.) noise.

## 5.2 Form Factor in Image Analysis

The form factor is defined as the ratio between R.M.S. to average value of a pixel sequence in an image as shown in (5.1).

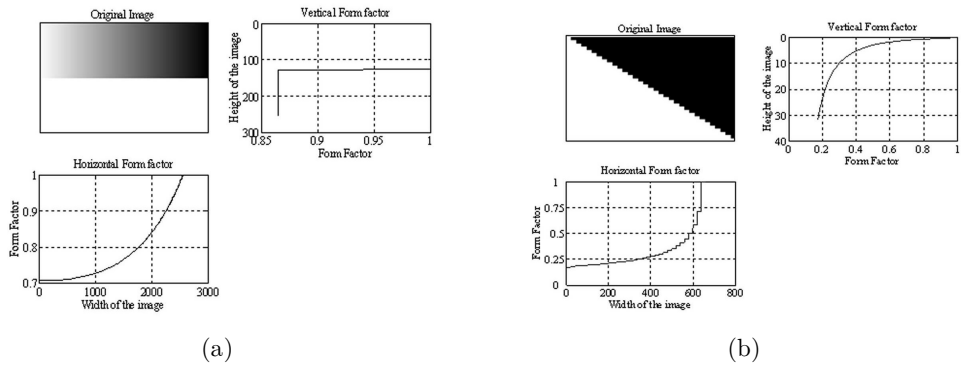

(a)  (b)

Figure 5.1: Horizontal and Vertical Form Factor for (a) Varying Contrast Image, (b) Varying Low Intensity Pixel Number

$$F = \frac{R.M.S.}{Average} = \sqrt{\left(1 + \frac{\sigma^2}{\mu^2}\right)} \qquad (5.1)$$

where, $\mu$ and $\sigma$ are the mean and standard deviation of the corresponding row or column.

This ratio provides two measures of an image. One is the contrast and the other is the volume of low intensity pixels. The FF increases with the increase in contrast level and the lower intensity pixel volume. This fact can be seen easily from Figure 5.1(a) and 5.1(b). Figure 5.1(a) shows the variations of the horizontal and the vertical FF with variation of contrast while keeping volume of the lower intensity pixel constant. In Figure 5.1(b) the contrast level is kept fixed and the volume of the lower intensity pixel is varied to find the change in both the form factors.

The form factor varies largely with wide variation in mean of the data. However, the mean of natural images like face, eye generally varies between pixel values of $50 - 240$ in both gray scale and NIR monochrome images. Within this operating range of image pixel values the form factor remains less sensitive to mean of the image pixels. This can be observed in sensitivity curve of form factor to the mean as shown in 5.2.

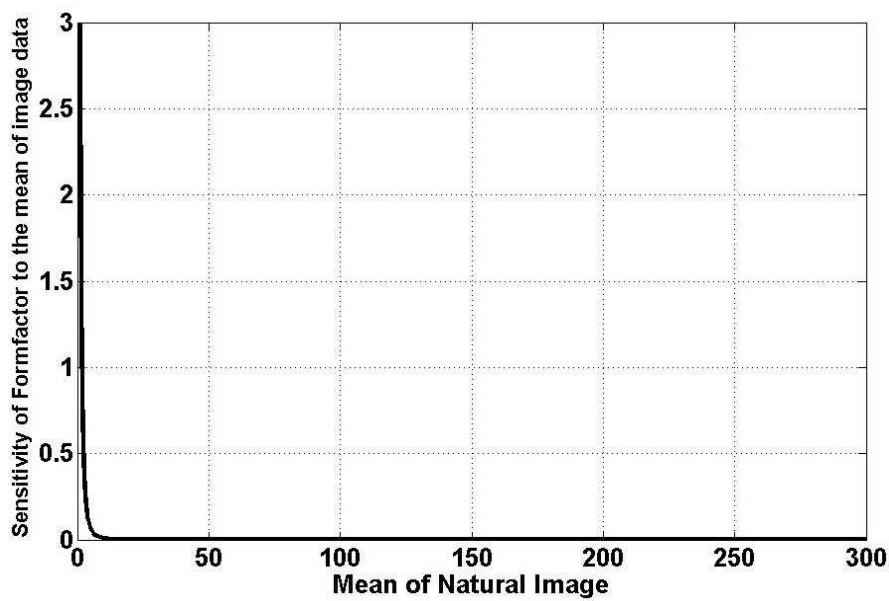

Figure 5.2: Sensitivity curve of Form Factor to Mean of image data

## 5.2.1 Directional and Local Form Factor

Form Factor can broadly be categorized in two groups as follows-

**Local Form Factor:** A small region within an image can be considered to form the pixel sequence to compute FF as in (5.1). This local FF can segregate the image into uniform and non-uniform zones. Thus edges between two regions can be highlighted by the local form factor. The edge detection method using this feature has been elaborated in Case Study I.

**Directional Form Factor:** This is determined using (5.1) in horizontal, vertical, and radial directions in an image. The pixel sequences for horizontal and vertical FF are single column and row respectively. Transformation of an image in polar form and subsequent computation of FF along angular direction provides the radial form factor. The former FF can be used for determining the pupil center in an eye image while the radial FF gives the estimation of pupil diameter. The details of the use of directional FF for finding the pupil center and its diameter are discussed in Case Study II.

### 5.2.2 Range of Form Factor

The FF defined in (5.1) can be rearranged to a form proportional to the ratio of norm-2 to norm-1 of the pixel sequence within the region of an image. The proportionality constant is the square-root of number pixels ($N$) in the region. Thus the form factor can be expressed as:

$$F = \frac{\left(N \sum_{i=1}^{N} x_i^2\right)^{\frac{1}{2}}}{\sum_{i=1}^{N} x_i} \tag{5.2}$$

Where, $x_i$ can assume integer values within its range (such as $0 - 255$ for digital image). This ratio provides the information of contrast and the relative number of low-intensity pixels within an image region. It is also a measure of the local energy. The FF is a dimensionless and derivative free quantity which is also insensitive to variations in illumination.

In any image it can be observed that the regions fall into two broad categories: homogeneous, and non-homogeneous that contains edges, lines, corners etc. In homogeneous regions the variance is zero and hence the FF is unity (from (5.1)). The non-uniform regions will result in FF values different from unity. The minimum limit is reached when the ratio of variance to mean becomes smallest. This is only possible when variance is the lowest and mean is the highest. A vector (element values within $0 - 255$) of all but one non-zero element satisfies the condition for the ratio to be extremum. Thus the range of FF can be found out analytically as derived below.

Differentiating FF in (5.2) partially by each of the $x_i$ and equating it to zero (i.e. $\frac{\partial F}{\partial x_i} = 0$) we get-

$$2F \frac{\partial F}{\partial x_i} = 2N \left(\sum_{j=1}^{N} x_j\right) \frac{x_i \sum_{j=1}^{N} x_j - \sum_{j=1}^{N} x_j^2}{\left(\sum_{j=1}^{N} x_j^2\right)^2} = 0 \tag{5.3}$$

The above equation leads to $\sum_{j=1}^{N} x_j = 0$ or $\sum_{j=1}^{N} x_j^2 - x_i \sum_{j=1}^{N} x_j = 0$. $F$ is set to be unity $\forall x_i = 0$ to avoid its indefiniteness. So the second option is the only possibility for FF to be extremum. The later condition results in either (a) all but one $x_i$ are zero or (b) all $x_i$ are equal. Thus from (5.1) we get-

$$1 \leq F \leq \sqrt{N} \tag{5.4}$$

### 5.2.3 Form Factor under Noise

In practice, the FF computed in different modes as described in section 5.2.1, are prone to be corrupted by noise. In this work, the zero-mean i.i.d. noise that occurs frequently in images has been considered. The FF can be recovered from its measurement under this type of noise and an estimation of SNR. The relationship can easily be established from (5.1) as-

$$F_s^2 = \frac{F_g^2 + SNR^{-1}}{1 + SNR^{-1}} \tag{5.5}$$

Where, $F_s$ and $F_g$ are FF without and with noise respectively. Here, SNR is required to be estimated from single instance of image region. An investigation on the blind estimation of SNR indicates that the estimation may be highly erroneous at lower level of noise. This leads to introduce the STVR for its recovery under such circumstances. Further analysis of (5.1) leads the following relationship.

$$F_s^2 = STVR(F_g^2 - 1) + 1 \tag{5.6}$$

Where, $\frac{\sigma_s^2}{\sigma_g^2}$ is STVR. The analysis on the blind estimation of SNR and STVR can be used for blind recovery of FF under noisy condition. It is presented in following section. Derivations of (5.5) and (5.6) are given in the Appendix-A

## 5.3 Blind Recovery of Form Factor under Noise

The FF can be recovered under zero-mean additive i.i.d. noise using a prior estimation of SNR or STVR. The estimate of SNR or STVR should be computed horizontally, vertically and radially for respective directional FF to obtain its noise-free version. On the other hand the recovery local FF requires the estimation in the local region.

However, in practice the reference noise-free images are often not available a priori. This constraint has led to opt the method for estimation of SNR or STVR blindly from single observation.

A theoretical analysis presented in this work shows that blind estimation of SNR from single image is highly inaccurate at lower noise level. This fact contradicts the experimental results reported in different literatures [98–101]. In this circumstance our proposed measure, STVR, is found to be a robust for its no reference estimation from single frame corrupted by zero-mean i.i.d. noise. Closed form relations have been developed to recover FF under zero-mean i.i.d. noise from the blind estimation of SNR and STVR as expressed in (5.5) and (5.6) respectively. Both theoretical and numerical assessments of the some promising methods [99–101] demonstrate the robustness of blind estimation of STVR over SNR from single instance.

SNR has been widely used as one of the Image Quality Metric (IQM) under noisy conditions [99–102]. A number of IQM estimation methods have been developed. Most of the methods are based on its full reference estimation [102, 103]. Here, the IQM can only be measured when clean image is available along with the noisy image. A recent work by [104] exploits the statistics of DCT of image for blind estimation of image quality distortions. However, the method requires a training phase for feature based classification of images into distorted and non-distorted observations. Some researchers have attempted to measure IQM blindly from single instance of image using the index SNR (or peak signal-to-noise ratio/PSNR) [98–101]. Albeit, no theoretical justification on the accuracy of its estimation is found.

In this section, three simple estimators have been analyzed for blind estimation of SNR as well as STVR from single instances. These are as follows -

1. Methods based on Local Statistics (LS)

    (a) Estimation by minimum of local noise variance (M1)

    (b) Estimation by average of local noise variance (M2)

2. Methods based on Subspace Decomposition (SSD) (M3)

3. Methods based on Autocorrelation Sequence (ACS)

(a) Estimation along horizontal direction (M4)

(b) Estimation along vertical direction (M5)

The analysis shows that the presence of estimation error in the image variance (or the noise variance) yields highly inaccurate SNR-estimates at lower level of noise even in the usable range of SNR ($0-20$ dB). Thus the estimated SNR may poorly recover the true FF in such circumstances. Analysis of other related index like Total-Variance-to-Noise Ratio (TVNR) can be carried out in the same direction as that of SNR. It follows that these indices are also erroneous at reduced level of noise.

The UBIRIS.v2 image data set [105] is used to test measurement efficiency of these blind estimation techniques under different levels of noise. The database is contaminated by zero mean i.i.d. noise of varying true SNR levels. The experimental results show that STVR provides higher accuracy through different levels of noise in contrast to that in case of SNR. The findings are found to comply with the theoretical justification of the proposition. The results also aid to select the best estimator of the index.

The remaining portions are organized as follows. The estimation techniques of SNR and a theoretical justification of its limitation have been presented in section-5.3.1. In the next section, the new index STVR is proposed and analyzed to demonstrate its effectiveness over SNR. In section 5.3.4 numerical results and discussion on the same are provided. Finally, section 5.6 concludes the present work.

### 5.3.1  Methods of Blind Estimation of Variance

#### 5.3.1.1  Local Statistics based Method

The estimation of noise level based on local variance relies on the existence of homogeneous region in the noise-free image [106, 107]. The variance of the additive noise can be determined by searching for minimum of the local variances over the entire image [98, 99] as shown in (5.7). However, this estimation may be lower than the true noise level of the image [99].

$$\hat{\sigma}^2_{min} = \min_{l \in [1, N_R]} \{\sigma_l^2\} \tag{5.7}$$

To overcome the inaccuracy in this estimation, the average of the local variances may be considered as shown (5.8). However, this estimate is found to be higher than the true variance.

$$\hat{\sigma}^2_{avg} = \frac{1}{N_R} \sum_{l=1}^{N_R} \sigma_l^2 \tag{5.8}$$

where $N_R$ is the number of regions and $\sigma_l^2$ is the variance of $l^{th}$ region.

Therefore, to minimize the estimation error some authors have proposed to take weighted average of these two estimates as shown in (5.9) [99].

$$\hat{\sigma}^2_n = c_\alpha \hat{\sigma}^2_{min} + (1 - c_\alpha)\hat{\sigma}^2_{avg} \tag{5.9}$$

where $c_\alpha \in [0, 1]$ and the estimate of noise variance lies in $[\hat{\sigma}^2_{min}, \hat{\sigma}^2_{avg}] \ \forall c_\alpha$. The image variance can be obtained by subtracting the estimated noise variance from the Total Variance (TV) of the noisy observation. Therefore the SNR can be computed as shown in (5.10).

$$\widehat{SNR} = \frac{\hat{\sigma}^2_s}{\sigma_g^2 - \hat{\sigma}^2_s} \tag{5.10}$$

where the $\sigma_g^2$ is the total variance of the noisy image.

### 5.3.1.2 Subspace Decomposition based Method

The eigen space of the observed noisy image comprises of image and noise subspace for a zero-mean additive i.i.d. noise. Thus -

$$R_g = R_s + R_n \tag{5.11}$$

where $R_g$ is the sample autocorrelation matrix of noisy image and $R_s$ and $R_n$ are that of image and noise respectively. Here, the image can be expressed as lexicographically ordered vector to determine the matrices.

Therefore, decomposition of the $R_g$ is equivalent to determine the span of image and noise space [100] as shown in (5.12).

$$\sigma_{gi} = \begin{cases} \sigma^2_{si} + \sigma^2_n & i = 1, 2, ...., N_{sd} \\ \sigma^2_n & i = N_{sd} + 1, N_{sd} + 2, ...., N_{tsd} \end{cases} \tag{5.12}$$

where $N_{tsd}$ and $N_{sd}$ are the total and image subspace dimensions respectively. $\sigma_{gi}$

and $\sigma_{si}^2$ are the total and image variances along $i^{th}$ eigen direction respectively. $\sigma_n^2$ is the variance of the i.i.d. noise. A Minimum Description Length (MDL) criterion defined similarly as in [100] can be employed to determine $N_{sd}$. The details of the method are available in [100].

The noise variance can be estimated as the average of eigen values of $R_g$ beyond $N_{sd}$ as in (5.13). Subsequently, the image variance is computed as shown in (5.14). Thus the SNR can be estimated following (5.10).

$$\hat{\sigma}_n^2 = \frac{1}{N_{tsd} - N_{sd}} \sum_{i=N_{sd}+1}^{N_{tsd}} |\sigma_{gi}| \tag{5.13}$$

$$\hat{\sigma}_s^2 = \sum_{i=1}^{N_{sd}} |(\sigma_{gi} - \hat{\sigma}_n^2)| \tag{5.14}$$

#### 5.3.1.3 Auto Correlation Sequence based Method

A blind estimation of SNR based on 2-D ACS of single noisy observation of magnetic resonance image has been developed recently [101]. The 2-D ACS can be computed as shown in (5.15).

$$r_g(i,j) = \frac{1}{(2W+1)(2H+1)} \sum_{m=-W}^{W} \sum_{n=-H}^{H} I_g(i,j) I_g(i+m, j+n) \tag{5.15}$$

where $2W+1$ and $2H+1$ are the width and height of the observed noisy image $I_g$.

The peak of 2-D ACS of noisy image (as shown in (5.15)) at the origin provides the sum of total variance and the square of the average of the observation. In case of zero mean i.i.d. noise the peak of $r_g$ at origin boils down to (5.16) and that of the clean image ACS can be expressed by (5.17).

$$r_g(0,0) = \sigma_s^2 + \mu_s^2 + \sigma_n^2 \tag{5.16}$$

$$r_s(0,0) = \sigma_s^2 + \mu_s^2 \tag{5.17}$$

where, $\sigma_s^2$ and $\sigma_n^2$ are the variances of image and noise respectively. $\mu_s$ is the average of clean image pixels and is equal to that of noisy observation for zero mean i.i.d. noise.

The estimation of $r_s(0,0)$ can be carried in several ways. Among these, the first order extrapolation of two adjacent neighborhood points of peak on $r_g$ in either horizontal or vertical direction may provide the best blind estimate $r_s(0,0)$ for i.i.d. noise [101]. Therefore, the noise variances can be computed by subtracting (5.17) from (5.16). The image variance can be determined from (5.17) for zero-mean additive i.i.d. noise model. Subsequently the SNR of the observed image can be estimated following (5.10).

### 5.3.2 Analysis of Estimation Error in SNR

In all the above methods for SNR estimation the total variance of the noisy image is known from the observation as well as its eigen space. Thus (5.10) can be considered as blind estimation of SNR in general. This expression can be used to determine accuracy in the estimate over different levels of noise. The estimated SNR obtained from (5.10) can be expressed as shown in (5.18).

$$\begin{aligned}\widehat{SNR} &= SNR_{true} + \Delta SNR \\ &= \frac{\sigma_s^2 + \Delta\sigma_s^2}{\sigma_g^2 - \sigma_s^2 - \Delta\sigma_s^2}\end{aligned} \quad (5.18)$$

where $\Delta\sigma_s^2$ is relative error in image variance estimation. Further, percentage error in SNR ($SNR_{pc}$) can be expressed as -

$$SNR_{pc} = \frac{100\sigma_g^2 \Delta\sigma_s^2}{\sigma_s^2(\sigma_n^2 - \Delta\sigma_s^2)} \quad (5.19)$$

It can be observed from (5.19) that the estimation error vanishes as the noise dominates over the image (i.e. $\sigma_n^2 \to \infty$) while a steady error of $-100\%$ can be observed when the noise variance is reduced to zero. Therefore the error at the extreme levels of noise can be determined as shown in (5.20) and (5.21). However, the $SNR_{pc}$ can go toward infinity when $\Delta\sigma_s^2$ possess positive magnitude. A negative value of $\Delta\sigma_s^2$ (i.e. if the estimate is lower than true image variance) keeps the error magnitude within 100%.

$$SNR_{pc}^0 = \lim_{\sigma_n^2 \to 0} SNR_{pc} = -100\% \quad (5.20)$$

$$SNR_{pc}^{\infty} = \lim_{\sigma_n^2 \to \infty} SNR_{pc} = \frac{100\Delta\sigma_s^2}{\sigma_s^2}\% \qquad (5.21)$$

It can be computed from (5.19) that for 20 dB true noise level, the error in the estimation of SNR in linear scale is equal to $-91.82\%$ for $\frac{\Delta\sigma_s^2}{\sigma_s^2} = 0.1$.

### 5.3.3 Analysis of Estimation Error in STVR

In practice, it is unlikely that the blind estimation of image variance from a single observation is completely error free. The analysis in Section 5.3.2 indicates that a robust index for recovery of FF is required other than SNR. STVR (shown in (5.22)) is proposed to be used for this purpose. The index determines the fraction of noise-free image embedded in the noisy observation. Estimation error analysis of this index in the following subsection reveals the advantage.

$$STVR = \frac{\sigma_s^2}{\sigma_g^2} = \frac{\sigma_s^2}{\sigma_s^2 + \sigma_n^2} \qquad (5.22)$$

The proposed ratio in (5.22) can be computed using the same estimators mentioned in section 5.3.1. The relative error in STVR can be obtained using (5.22) as shown in (5.23).

$$\begin{aligned}\Delta STVR &= \frac{\hat{\sigma}_s^2}{\sigma_g^2} - \frac{\sigma_s^2}{\sigma_g^2} \\ &= \frac{\Delta\sigma_s^2}{\sigma_g^2}\end{aligned} \qquad (5.23)$$

Thus the percentage error in STVR estimation is as follows.

$$STVR_{pc} = \frac{100\Delta\sigma_s^2}{\sigma_s^2} \qquad (5.24)$$

Equation (5.24) indicates that $STVR_{pc}$ is equal to the percentage error in image variance for all levels of noise. A small error in image variance can keep the error in STVR to a low value. At this point STVR can be proved to be effective over any other indices like PSNR, TVNR, and noise-to-total variance ratio (NTVR) etc. following similar procedure carried out in this chapter.

### 5.3.4 Experimental Results & Discussions

#### 5.3.4.1 Data set

The numerical analysis of the proposed index and SNR have been performed on randomly selected 120 eye images images from UBIRIS.v2 database [105]. These images also provide large variations in local statistics. The selected images are corrupted by zero mean i.i.d. noise of different levels within $[0.001\sigma_s^2, \sigma_s^2]$ or $(0-30)$ dB to generate the test image set.

#### 5.3.4.2 Experiment

Five estimation techniques (M1, M2, M3, M4, M5) as mentioned in the section 5.3, have been used to estimate image variance and subsequently the SNR and STVR on the test data set. Here, M1 and M2 are used to have lower and upper bound of the estimates given by (5.9). The numerical results of the percentage estimation error in image variance, SNR, and STVR of a sample image with $\sigma_s^2 = 45.332$ are presented in Figure 5.3, Figure 5.4, Figure 5.5, and Figure 5.6 respectively.

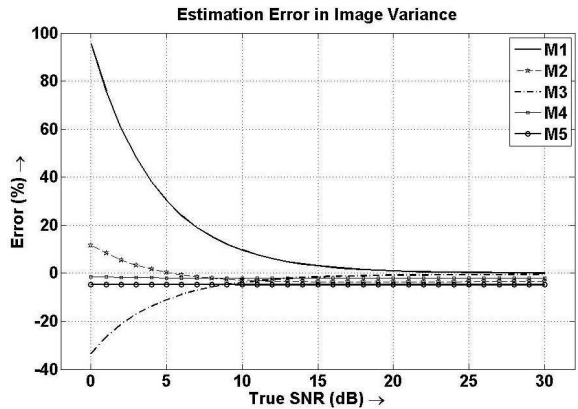

Figure 5.3: Percentage Error in Variance Estimation of an Image

It can be observed from the Figure 5.3 that the relative errors in image variance ($\sigma_s^2$) estimation assume positive values over the given noise level for method based on M1. This yields percentage error in SNR estimation beyond 100% as can be observed in the Figure 5.4. It can also be noticed from Figure 5.5 that the SNR estimates are well within ±100% and the error magnitude increases as the noise level is re-

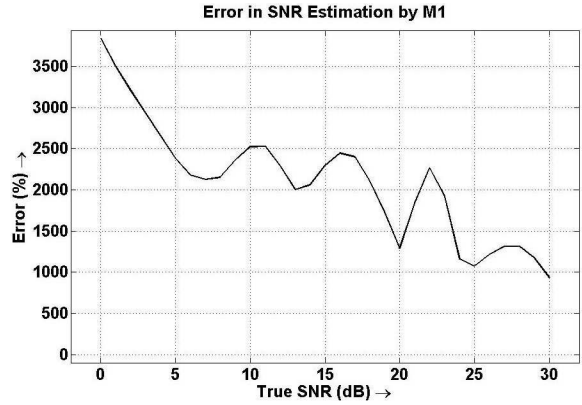

Figure 5.4: Percentage Error in Estimation of SNR by M1

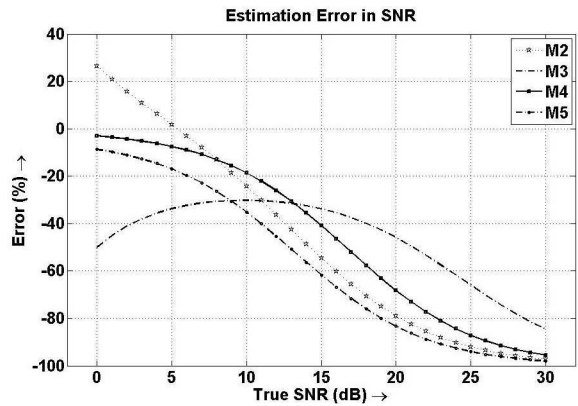

Figure 5.5: Percentage Error in Estimated SNR

---

Method based on M1: Min. of Local Variance, M2: Mean of Local Variance, M3: Subspace Decomposition, M4: Horizontal extrapolation of ACS, M5: Vertical extrapolation of ACS duced. On the other hand, Figure 5.6 shows that the percentage errors in STVR are equal to those in $\sigma_s^2$ by all the estimators as it is expected from (5.24). It can also be observed that the $\sigma_s^2$ and the STVR estimated by the method based on M4 are low over the given range of noise. The upper bound of the estimates of $\sigma_s^2$, SNR and STVR values over the test data set for all the estimators are summarized in Table 5.1.

The Table 5.1 shows that the upper bounds of the Absolute Percentage Error (APE) vary largely in the estimated $\sigma_s^2$. These variations may be due to the differences in the global and local statistics of the test images. The method based on horizontal extrapolation of ACS is found to yield lowest upper limit of the estimation error.

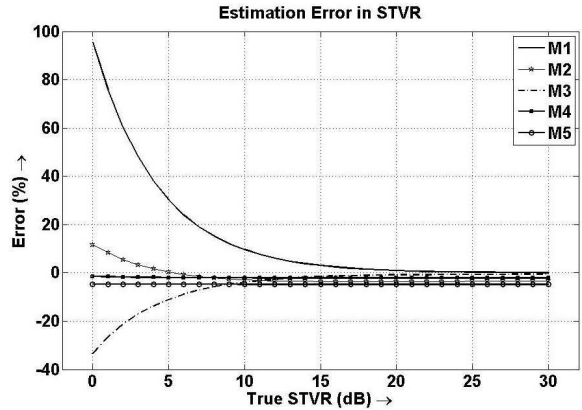

Figure 5.6: Percentage Error in Estimated STVR

Table 5.1: Upper Limit of Estimation Error in %

| Index | Methods | | | | |
|---|---|---|---|---|---|
| | M1 | M2 | M3 | M4 | M5 |
| $\sigma_s^2$ | 96.48± 0.90 | 12.46± 1.46 | 33.39± 0.03 | 1.90 ± 1.13 | 3.87 ± 1.12 |
| $SNR$ | $(5.1 \pm 1.2) \times 10^3$ | 96.28± 1.69 | 83.28± 0.70 | 91.07± 9.55 | 97.25± 1.14 |
| $STVR$ | 96.48± 0.90 | 12.46± 1.46 | 33.39± 0.03 | 1.90 ± 1.13 | 3.87 ± 1.12 |

Method based on M1: Min. of Local Variance, M2: Mean of Local Variance, M3: Subspace Decomposition, M4: Horizontal extrapolation of ACS, M5: Vertical extrapolation of ACS

However, the maximum bounds of the absolute percentage error in SNR are well beyond or close to 100% for all these estimation techniques while the upper bounds of APE of STVR follows those of the image variance. Here, also the method M4 produces best upper bound of APE of STVR.

## 5.4 Case Study I: Edge Detection in Images by Local Form Factor

### 5.4.1 Related Works in Edge Detection

Extraction of image features like edges, lines etc. finds its application in image analysis as well as computer vision for several decades. It requires a number of local image pre-processing techniques [108–111]. The objective of these methods is to detect and locate the pixel where abrupt change in intensity is observed. An edge which is essentially associated with individual pixel [108], can be described as a boundary dividing two regions of abruptly different average intensity levels. Literature suggests that the intensity level in an image can be described by local energy measure with a magnitude and a direction. A significant variation in local energy which can be perceived by human eye indicates the existence of edges in that region [112, 113].

In the past few decades considerable research has been performed for development of optimal edge detectors using various models. These models can be broadly divided into two categories. One that detects the edges based on computing gradient of intensity [108, 114–117] where the edges are located at the pixel when the gradient is maximum. The other model is based on the human visual perception mechanism [109, 112, 113, 118–120].

The gradient based detectors, like Sobel, Prewitt, Kirsch, and Robinson etc. have used masks to approximate the first derivative of intensity at an image pixel to find the edges [108, 109]. However, these methods are insensitive to the orientation of region boundaries. Performance of these convolution based methods depends on the size of the region bounded by the edges and achieves accuracy only for specific images [108]. On the other hand, Laplacian operator finds the second derivative to determine the edge magnitude at each pixel in an image using mask of different sizes. It provides no information of orientation of the edge and may produce double response to a few types of edges [94, 108, 109]. The later development in this line such as [111, 114–116] utilizes the fact that second derivative crosses zero value at the edge pixel in an image. The advantage of these methods over the earlier ones is that zero crossing point is easier to search compared to finding the point of the largest first derivative. All of these zero crossing based methods use a smoothing filter to remove the noise present in the image before the computation of second derivative at the cost of pushing the edges away from their true positions. In general the inaccuracy in localization depends on the scale or the standard deviation of the Gaussian smoothing function used to filter the image. However, in practice precise localization of edges is required for image analysis as well as machine vision (such as robotic vision). Literature reports that the Gaussian smoothing may lead to suboptimal detection of edges [108]. Thus choice of the smoothing scale is crucial. However no systematic method exists to find a suitable scale for an optimum performance. Moreover, all these methods are highly sensitive to noise and illumination variations. In [114] Canny has developed an optimal detector for step edges. However, it cannot accurately detect other types of edge profiles such as roofs, lines and gratings etc. Amongst all variants and derivatives, the Canny's method is found to be the most popular choice as reported in the literature [111, 114].

The other approach induced by the human visual perception mechanism uses local energy measure. Literature has suggested that the congruency of phase achieves maximum at the edge pixel [112, 113]. Thus the Phase Congruency (PC) has been reported as an index of local energy [118–120]. This method essentially operates on image in the frequency domain. The phase congruency is defined as the ratio of the magnitude of vector sum of a quadrature pair of filter outputs to the scalar sum of the amplitude of the filter pair outputs [118]. Here the choice of filter kernel may affect the output of the detector. The performance is also dependent on the amount of frequency spread and scales considered for the measurement [118–120]. The local energy is weighted based on the spread of frequency in the computation of PC to avoid the anomalies due to low and limited frequency spread. These weights are computed using arbitrarily fixed weighting fraction and a gain. It has been reported that the PC is highly sensitive to noise being a normalized dimensionless index [118–120]. A threshold, which is used to remove the effect of noise, is also dependent on a set of constants which are chosen by trial and error. This threshold is an approximate measure of the upper bound of the noise with the consideration that the noise spectrum is uniform in all orientations. The other problem associated with the PC based feature detection is its computational complexity [118]. Although, the algorithm is developed using wavelet based filtering technique the method is implemented via fast Fourier transform. The high computational burden is due to large number of convolution operations with filters for many wavelet scales and orientations.

In [110] another edge detection algorithm based on local energy measure is presented. However, the method is not simple and the computational complexity is high due to the same reason as mentioned in the phase congruency measure. It is also reported that the detector may fail under practical circumstances as the integration of the edges at different scales may not be accurate [110].

Apart from the methods mentioned above there are a few other concepts proposed in the literature for detecting the edges. In Smallest Univalue Segment Assimilating Nucleus (SUSAN) technology the feature strength is computed using the difference between the relative intensities of surrounding pixels and the center pixel of a small circular region which is called nucleus [121]. An exponential function of this difference is used as the feature value at the center pixel of the nucleus. The detector requires two user settable thresholds, the choice of which affects the detection performance. An improved SUSAN technique has been reported in [122]. However, the complexity of this algorithm increases two fold as compared to the original SUSAN proposed in [121]. Moreover, it assumes that the edges are generated at the intersections of region where there are little or no variations in the intensity level. This may limit the application of this algorithm when this condition is not met [120].

The edge detection using the multi-scale image processing approach reported in the literature may not exhibit good performance as the feature count and the location vary at different scales. Moreover, the computational burden of these algorithms is often high for practical uses [123].

An 'on and off' edge filter based on learning probability distribution has been reported in [124, 125]. Here, the learning is based on gradient edge map. The performance has been tested using misclassification error of edge pixel. However, the misclassification error has been reported to be a poor performance index for edge detection algorithm [126]. Moreover, choice of filter scale during both learning and test may influence the outcome.

The constraints of the edge detection algorithms are summarized in Table 5.2. The limitations of the algorithms indicate that a universal solution of the problem is required.

The aim of this case study is to develop an approach of edge detection based on variation of local energy in an image. It can be observed that the local energy can be measured by locally computed form factor. Inverse of the square of local FF around a center pixel is defined as an Edge Strength Index (ESI) at that pixel. It also provides the measure of entropy within a region of an image. The advantage of such measure is the ability to recover it under zero-mean i.i.d. additive noise.

Table 5.2: Constraints of a Few Edge Detection Algorithms

| Algorithms | Limitation/Constraints |
|---|---|
| Gradient based edge detector | Sensitive to noise, Pre-smoothing introduces error in edge localization |
| Local energy based edge detector | Computationally expensive, Empirical determination of scales needed, number of orientations, adjustments of tuning parameters are by trial and error |
| SUSAN technology based edge detector | Difficulty may arise in optimizing the two user settable parameters, Noise sensitivity due to differential term in feature strength function may be present, assumption made on formation of edges may not be true in practice |
| Multi-scale edge detectors | High computational complexity. Integration of edge at different scale may be erroneous. |

It only needs a local estimate of either SNR or STVR to compensate the effect of noise as explained in Section 5.3. The compensated ESI which is computed within a region of size of $3 \times 3$ is passed through Non-Minimum Suppression (NMS) within $5 \times 5$ window and a universal threshold has been applied to produce the final edge map.

The results from both the proposed algorithm and the Canny edge detector are compared. The performance of the proposed algorithm has been evaluated quantitatively using Baddeley Error Metric (BEM) [126] and compared with those resulted from the Canny edge detector at different scales. The experimental results indicate that the form factor based edge detector can be used as a potential candidate for general feature extraction.

### 5.4.2 The Proposed Method

#### 5.4.2.1 Edge Strength from Local FF

The FF defined in (5.2) is not a normalized quantity. In this section, inverse of square of FF is defined as a normalized index to quantify the image feature such as an edge.

The normalization facilitates to adopt universal thresholds for finding the edges. Thus the edge strength index can be expressed as:

$$\alpha = \frac{1}{F^2} = \frac{\left(\sum_{i=1}^{N} x_i\right)^2}{\left(N \sum_{i=1}^{N} x_i^2\right)} \quad (5.25)$$

The $\alpha$ is known as the ESI which indicates the degree of uniformity of pixel values within the region of interest. In any image, regions belonging to two broad categories may be observed such as (a) homogeneous, and (b) non-homogeneous that contains edges, lines, corners etc. In homogeneous regions the variance is zero and hence the ESI is unity (from (5.1) and (5.25)). The non-uniform regions will result in ESI values different from unity. Since it is a normalized index the maximum value is 1. The minimum limit is reached when the ratio of variance to mean becomes largest. Thus from (5.4) and (5.25) we get-

$$\frac{1}{N} \leq \alpha \leq 1 \quad (5.26)$$

Figure 5.8 confirms the range pictorially for the case of a nine element vector. The result will be same for any other combinations.

All $\alpha$ values within its range do not represent an edge pixel. A coarse idea on the values of ESI for edge can be obtained from the relationship between $\alpha$ and entropy of a region. It is observed that the ESI varies inversely with entropy. It can be expressed as in (5.27) for a Gaussian distribution.

$$\alpha = \frac{\mu^2}{\mu^2 + \frac{1}{2\pi} e^{(2H-1)}} \quad (5.27)$$

Where, $H$ is the entropy of the region.

The relation in (5.27) can easily be established for the entropy of a Gaussian distribution. It is logarithmically proportional to variance of signal concerned [127]. It follows that the extreme values of entropy produce extreme values of ESI. It is evident that a region with low entropy is completely uniform while the high entropy is associated with higher degree of irregularities in the region. Thus prominent edges cannot exist in the region with such extreme values of entropy or ESI. Therefore the ESI values of an edge pixel will tend to lie mid-way within its range. Statistical test results indicate the boundary of ESI for an edge as:

$$0.5 \leq \alpha^{edge} \leq \alpha_{high} \qquad (5.28)$$

where, $\alpha_{high}$ is close to unity and limits the contrast level up to which edges (weak edges) are to be detected. The range of ESI in (5.28) implies that an edge exists in a region where RMS value of pixels is less than $\sqrt{2}$ times of their average value.

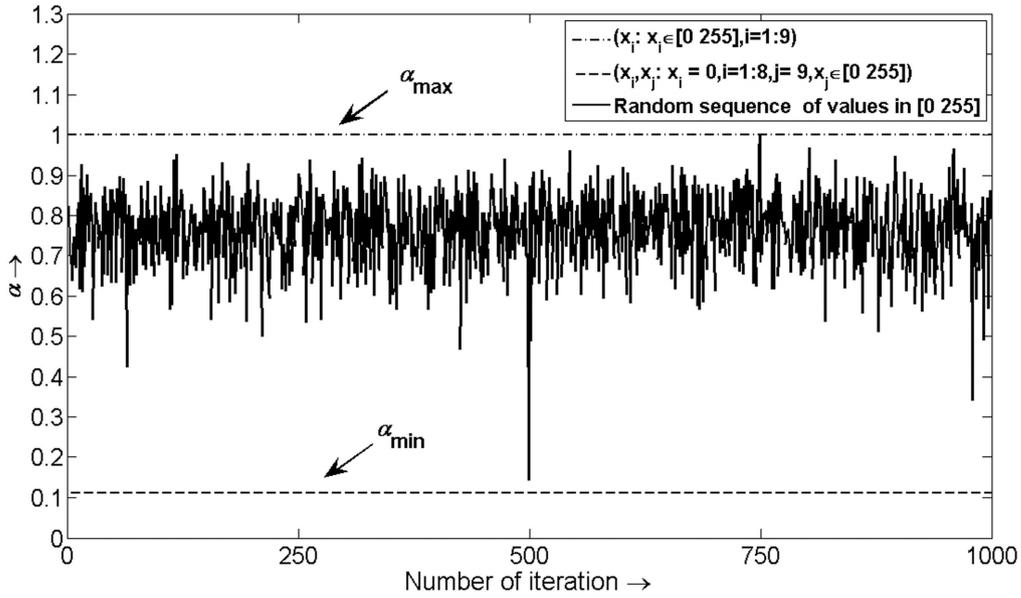

Figure 5.7: Range of $\alpha$ for a Sample Sequence of Length Nine

Figure 5.8 depicts the edge strength values for a step edge in 1-D (similar to different level of contrast in image). Here, $\alpha$ is computed within a moving window of constant size. It is apparent from the figure that the ESI falls within the range as in (5.28) in presence of an edge.

### 5.4.2.2 Edge Strength under Noise

The true edge strength index can be recovered from its measurement under zero-mean i.i.d. noise and an estimation of SNR. The relationship can easily be established from (5.25) and (5.5) as:

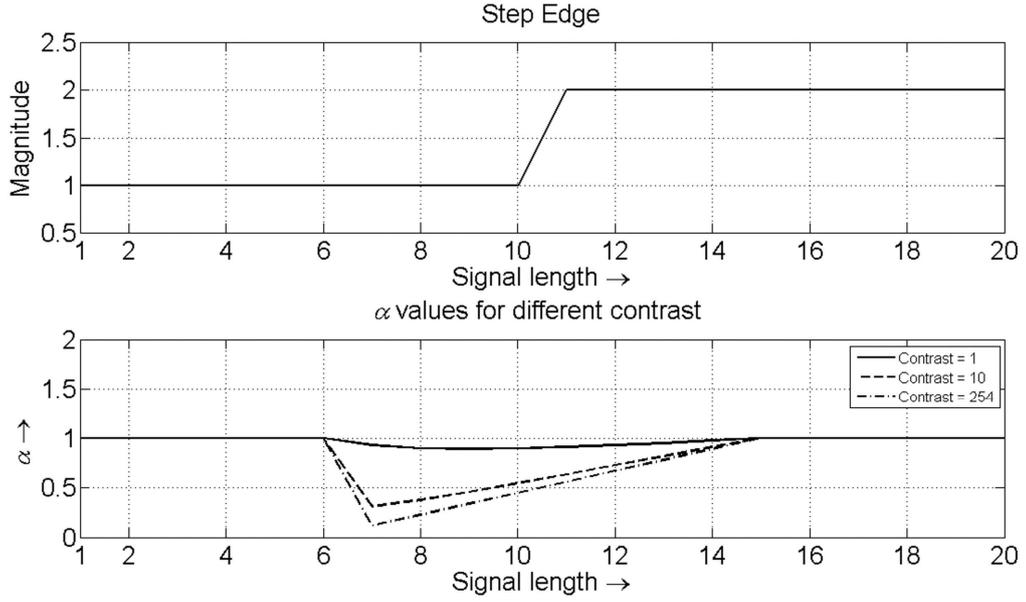

Figure 5.8: Range of $\alpha$ of Edge Points for Step Edge

$$\alpha_s^2 = \frac{1 + SNR^{-1}}{\alpha_g^{-2} + SNR^{-1}} \qquad (5.29)$$

Where, $\alpha_s$ and $\alpha_g$ are ESI without and with noise respectively. Here, SNR is required to be estimated from single instance of image region. The inaccuracy in its estimation limits the use of (5.29) for recovery of true ESI. On the other hand STVR can be used for its recovery with following relationship.

$$\alpha_s^{-2} = STVR(\alpha_g^{-2} - 1) + 1 \qquad (5.30)$$

Where, $\sigma_s^2/\sigma_g^2$ is STVR.

### 5.4.2.3  Algorithm for Proposed Methodology

The steps of the algorithm are presented in the flow chart as in Figure 5.9. The $\alpha_{high}$ is decided by the saliency of the weak edges to be detected. The edge index for noise free image is computed following (5.30).

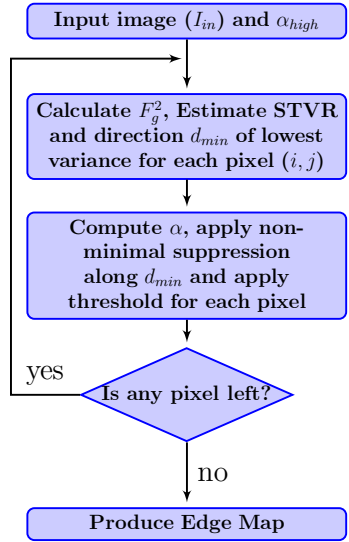

Figure 5.9: Proposed Edge Detection Algorithm

The proposed algorithm generates a noise compensated ESI-map of the image. Along the true edges of the original image the ESI values in the map are supposed to attain minimum in the direction of lowest variance. However, there may be spreading around the true edges due to local moving windowed operations. These ESI values is passed through an NMS algorithm along the direction of lowest variance within the window of size $5 \times 5$ around each pixel. It retains the lowest of the ESI values generated from the multiple responses around the true edges. Finally, a threshold can be applied following (5.28) to generate the edge map.

### 5.4.3 Experimental Results

In this section, both the qualitative and the quantitative performance results of the proposed algorithm for edge detection have been presented. A number of standard images referred in different literature are considered for qualitative assessment [118–123]. The images contaminated with different SNRs are subjected to the proposed algorithm. The performance of our algorithm has been evaluated with respect to the Canny edge detector as it has been accepted as some short of standard by several authors.

In the second phase of experiment, quantitative performance of our algorithm has been judged on randomly selected sixty images from Berkley Segmentation Data Set [128]. This data-base also provides edge maps generated by human subjects for widely varying natural images. These are used as the ground truth for performance tests. The BEM has been chosen as the quantitative performance measure. Here also the relative performance has been evaluated with respect to Canny's method.

### 5.4.3.1 Edge Map of Images without Noise

The images contain different type of edges and regions of varying contrast. In this subsection the results of our proposed method and those by Canny's edge detector have been presented for the noise free images.

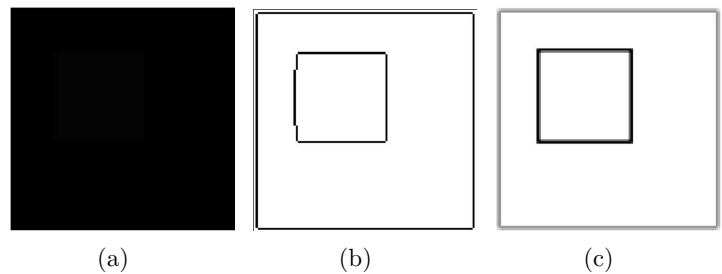

Figure 5.10: (a) Original Image with Low Contrast and Lower Pixel Value, (b) Edge Image by Canny Edge Detector, (c) Edge Image by Proposed Method

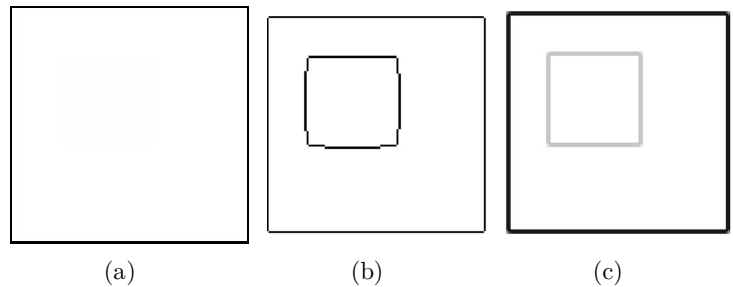

Figure 5.11: (a) Original Image with Low Contrast and Higher Pixel Value, (b) Edge Image by Canny Edge Detector, (c) Edge Image by Proposed Method

In Figure 5.10 and Figure 5.11 two simple images which contain two regions with intensity difference of 1 pixel have been considered. A small square is embedded in both the images. In Figure 5.10 the intensity level of these two regions are 0 and 1, while those in the Figure 5.11 is 254 and 255. These images are considered to demonstrate the effectiveness of the proposed algorithm with contrast level as low as 1 for gray scale images which are invisible to human eye too. The output from Canny's method with default scale 1 is also presented for comparison.

It can be observed that the above images contain only the step edges. Literature suggests many other types of edges such as roof, lines and gratings etc. [120]. The performance of the proposed algorithm has been demonstrated for various types of edges in Figure 5.12(a) and Figure 5.13(a). Figure 5.12(a) shows an image with gratings. This may exist in practice due to various orientations of sources of light and those of imaging devices. The decay exponent is considered to be equal to 1. Details about such types of edges are available in [120].

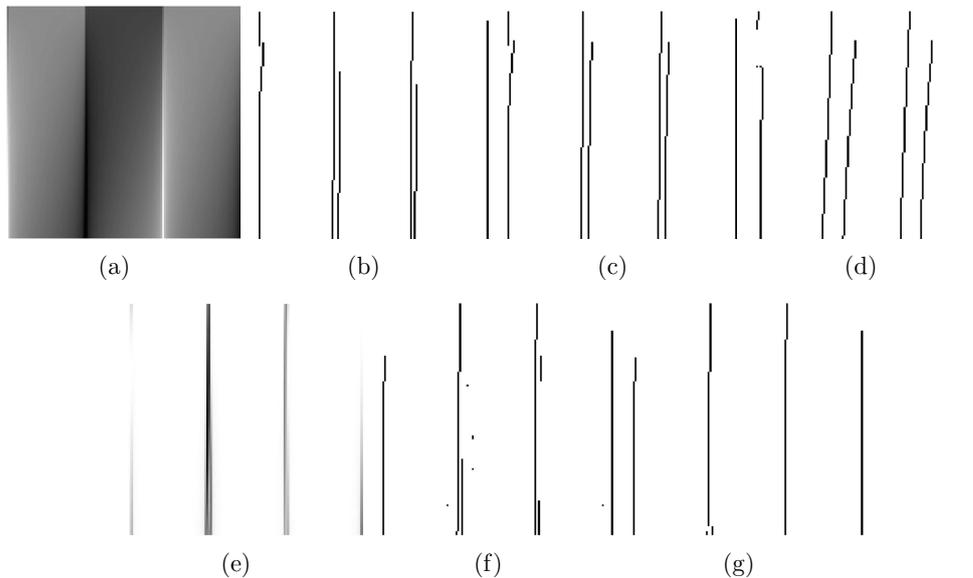

Figure 5.12: (a) Original Image with Gratings, (b) Canny Edge Detection with $\sigma = 0.75$, (c) Canny Edge Detection with $\sigma = 1.0$, (d) Canny Edge Detection with $\sigma = 3.0$, (e) Proposed Edge Detection without Non Minimum Suppression (NMS), (f) Proposed Edge Detection with $3 \times 3$ NMS, (g) Proposed Edge Detection with $5 \times 5$ NMS

The outputs of our algorithm applied on an image with curved edges are shown in Figure 5.13(f), 5.13(g), and 5.13(h). In the original image (shown in Figure 5.13(a)) a star like curved edge is present in a varying contrast region. Canny's method is also applied on the same image and presented for comparison.

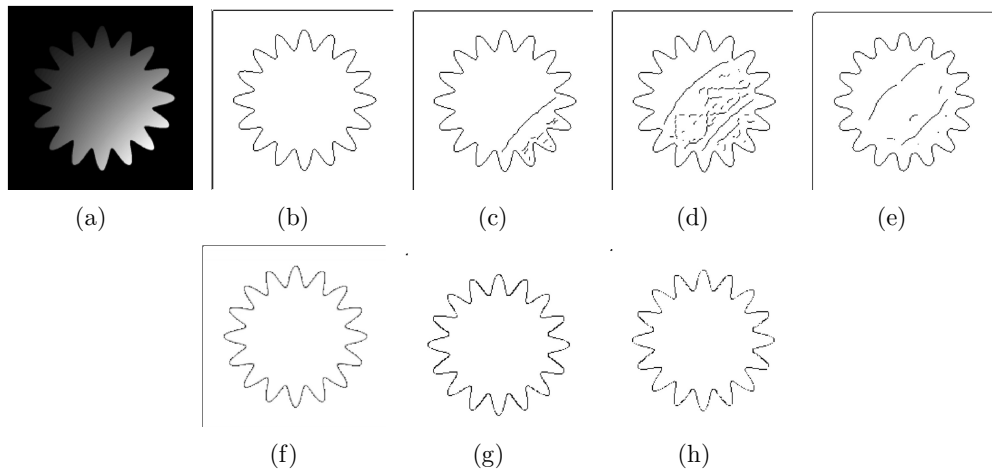

Figure 5.13: (a) Original Star Image without Noise, (b) Canny Edge Detection with $\sigma = 0.75$, (c) Canny Edge Detection with $\sigma = 1.0$, (d) Canny Edge Detection with $\sigma = 1.5$, (e) Canny Edge Detection with $\sigma = 3.0$, (f) Proposed Edge Detection without Non Minimum Suppression (NMS), (g) Proposed Edge Detection with $3 \times 3$ NMS, (h) Proposed Edge Detection with $5 \times 5$ NMS

#### 5.4.3.2  Edge Map of Images with Noise

Under practical situations any image is corrupted with various kinds of noises. Therefore, the performance of any edge detection algorithm is needed to be judged in presence of noise. The zero mean i.i.d. noise of different SNR has been added to the image as shown Figure 5.13(a). Such a noisy image with 15 dB SNR is shown in Figure 5.14(a). The edge maps produced by the Canny's edge detection algorithm as well as the proposed algorithm are shown in Figure 5.14(b) to Figure 5.14(h).

### 5.4.4  Quantitative Evaluation

**Data Set:**

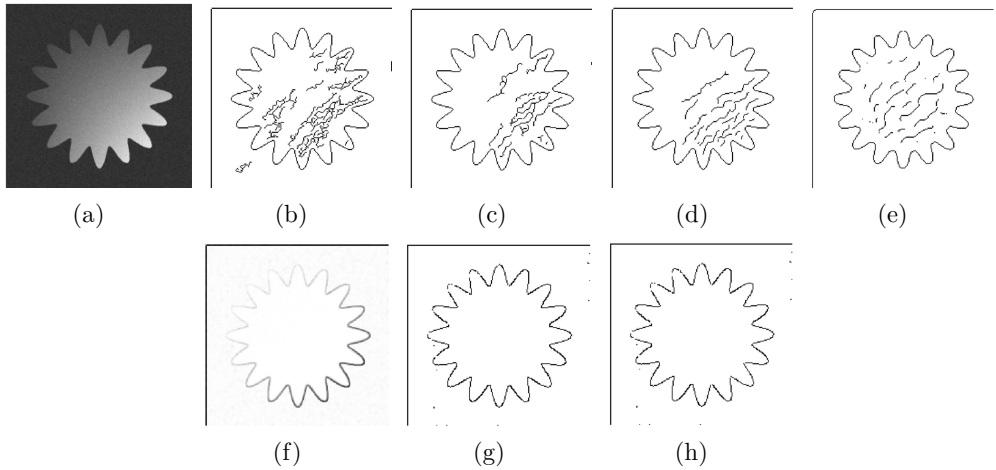

Figure 5.14: (a) Original Star Image with Noise of 15 dB, (b) Canny Edge Detection with $\sigma = 0.75$, (c) Canny Edge Detection with $\sigma = 1.0$, (d) Canny Edge Detection with $\sigma = 1.5$, (e) Canny Edge Detection with $\sigma = 3.0$, (f) Proposed Edge Detection without Non Minimum Suppression (NMS), (g) Proposed Edge Detection with $3 \times 3$ NMS, (h) Proposed Edge Detection with $5 \times 5$ NMS

The performance of the proposed algorithm has been evaluated on randomly selected sixty images from the Berkley Segmentation Dataset [128]. It consists of a wide range of natural images along with $5 - 9$ binary edge maps for each of these images obtained by human subjects. These binary edge images can be considered as ground truth for comparison of the output of any edge detection algorithm [129, 130]. The images in this data set are of sizes $321 \times 481$ and $481 \times 321$.

**The Performance Index:**

Literature suggests three criteria [108, 114, 118–120, 126, 131, 132] to evaluate the performance of edge detectors. These are as follows-

1. **Good detection:** Almost all edges must be discovered with minimal false-positives i.e. assigning a non-edge pixel to an edge.
2. **Good localization:** The pixel detected as an edge point should be very close to the center of true edge.
3. **Single response:** Single edge must be detected corresponding to an edge with

single pixel width.

Canny had shown that the first two criteria contradict each other [114, 126]. The third criterion is generally alleviated by employing non maximum suppression followed by a hysteresis thresholding. In the context of present algorithm, non minimum suppression has been used since the ESI is inversely related to local energy. Therefore a performance index for edge detector is primarily concerned with the first two criteria.

Studies suggest few indices generally employed to assess the performance of an edge detector with the available ground truth [123, 126, 130, 133, 134]. These are as follows-

- Misclassification Error
- Pratt's Figure of Merit (FOM)
- Baddeley Error Metric

The misclassification error is related to finding the probability of incorrect classification for every pixel in an image. It is evident that this index may be incapable of capturing inaccuracies due to shape distortion of object boundaries. This leads to the edge localization error. A detail review on this index can be found in [134]. The Pratt's FOM suffers from the above anomaly too. In addition to this the index is found to be insensitive to false negative error. The proper theoretical justification is also unavailable [126]. The BEM proposed in [126] has been considered in several research works for measuring errors in detection and localization. This index is devoid of the shortcomings prevailed in the earlier two metrics [130, 135].

The BEM is computed using Hausdorff distance for binary images [126]. The metric is expressed as follows:

$$\Delta_w^p(I, I_{ref}) = \left[\frac{1}{N}\sum_{x \in X}|w(d(x,I)) - w(d(x,I_{ref}))|^p\right]^{\frac{1}{p}} \quad (5.31)$$

Where $N$ is the number pixels on edge, $d(x, I)$ is the shortest distance from $x \in X$ to $I \subseteq X$, $w(t)$ is a continuous function on $[0, \infty]$, concave and strictly increasing and $1 \leq p \leq \infty$. The transformation $w(t)$ is defined in [126] as $w(t) = \mathbf{min}(t, c)$ for a fixed $c > 0$. Thus BEM indicates that replacing the reference edge image ($I_{ref}$) by the edge map ($I$) will change the image by $\Delta_w^p$. The parameter $p = 2$ denotes the Euclidean distance transform. The error in localization is weighted by the distance between $I_{ref}$ and $I$ of size $K \times L$, when the scale factor $c = \sqrt{K^2 + L^2}$ [126, 129, 130, 135]. The lower value of $\Delta_w^p$ indicates that $I$ is close to $I_{ref}$. Detailed analysis of the BEM can be found in [126]. In the present work, $\Delta_w^2$ is chosen as the performance index for the proposed edge detector.

**The Experiment:**

The experimental procedure for quantitative evaluation of the edge detector on a given image taken from the BSDS can be performed in two ways. In one method a common ground truth is obtained by determining a consensus image from the human-made edge maps of an image in the data set [129, 130]. The other way is to compute BEM for each of the segmented image associated to the test image and average all BEMs per image.

In this context first, a consensus image is developed for each of the gray scale image from the data set using the mini mean and mini max methods as explained in [136]. A sample image and corresponding human-made segmentations and the consensus ground truth obtained by both the methods are shown in Figure 5.15. The ground truth which is obtained through mini-mean method is found to be poorer than any of the available segmented images. On the other hand the consensus image generated from mini-max method includes the edge pixels detected and localized by one or very few human subjects. Use of either of the consensus image may produce inaccurate BEM values. The second option is chosen i.e. averaging all the BEMs computed for the edge maps per ground truth images. The procedure can be summarized below-

**Step-1** Compute binary edge map for a given image with and without noise following the algorithm depicted in Figure 5.9. Here, the weak edge detection limit ($\alpha_{high}$) is chosen to be 0.90.

**Step-2** Compute $\Delta_w^2$ by considering each of the human-made segmented image as reference and the generated edge map as test image using (5.31). Here the scale factor is $c = \sqrt{321^2 + 481^2} \approx 578$. In this case the BEM in (5.31) lies within $[0, 1]$.

**Step-3** Average all the BEM to produce a single measure of performance of an edge detector for an image.

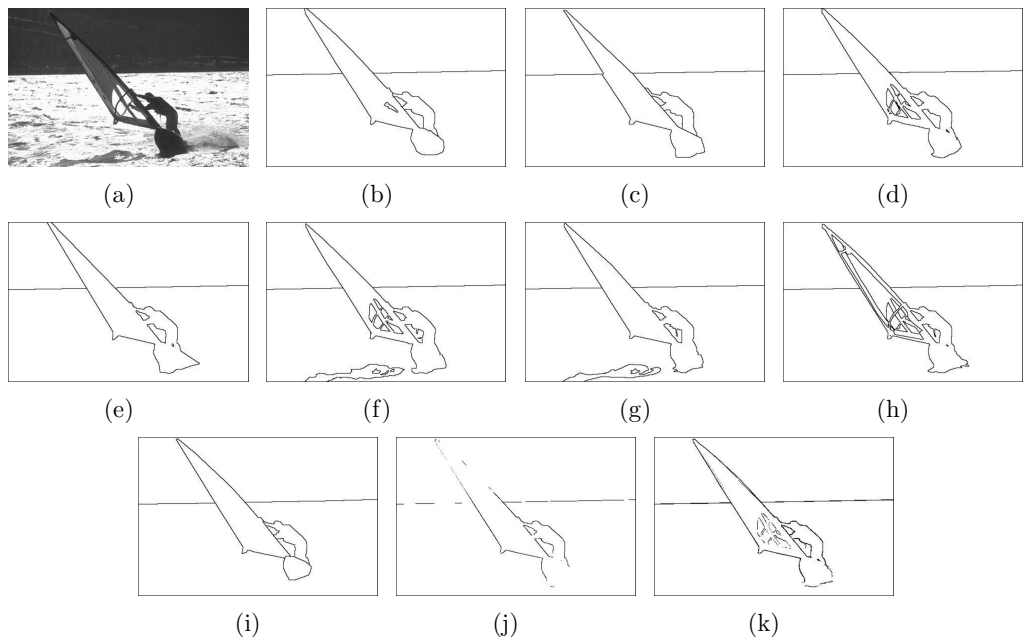

Figure 5.15: (a) Original Image Sample from BSDS Database, (b)-(i) Ground Truth Images by Different Human Subjects, (j) Consensus Ground Truth by Mini Mean Method, (k) Consensus Ground Truth by Mini Max Method

Same steps have been followed to compute the average BEM per image of the same data set for the Canny's edge detector. The results for the images without noise are shown in Figure 5.16. The same is obtained for noisy images and the results for SNR = 4 dB and 20 dB are shown in Figure 5.17 and Figure 5.18. The overall average BEM with standard deviation have also been computed for the whole data set. This is shown in Figure 5.19. Here, the variations of the index for image without and with noise of SNR = 0 dB, 4 dB, 16 dB and 20 dB have been presented.

### 5.4.5  Discussions

The effectiveness of the proposed algorithm has been shown qualitatively as well as quantitatively. It can be observed in Figure 5.10 and Figure 5.11 that the algorithm can detect the edges formed at the junction of two image regions with pixel difference of 1 which can be considered as weak edge. The outputs for the Canny edge detector have also been presented for the same images for comparison. A little distortion can be seen in the edge map by the Canny edge detector (Figure 5.10 and Figure 5.11).

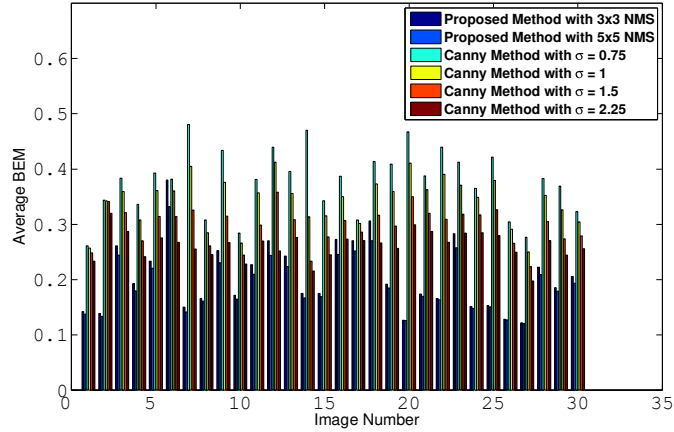

(a)

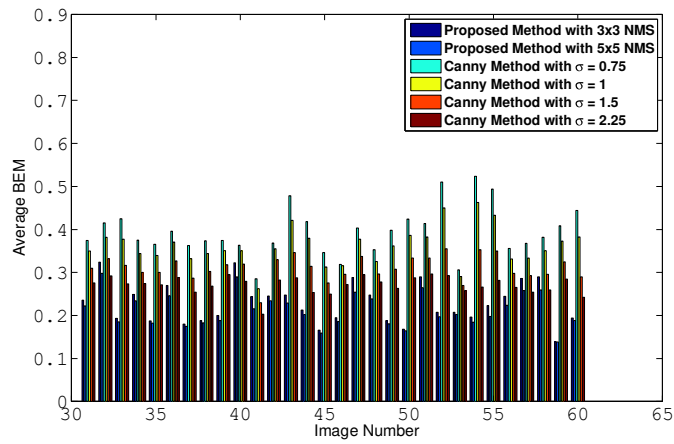

(b)

Figure 5.16: BEM on BSDS Image Data Set without Noise for Proposed Method with $3 \times 3$ NMS, $5 \times 5$ NMS, Canny Edge Detector with Different Scale ($\sigma$)

The detection of weak edges is limited by the choice of the upper bound of ESI. The proposed method computes the edge strength from the local FF which is a measure of local energy. Thus it can be associated with local entropy which is inversely proportional to ESI. It is apparent that the entropy of an edge region will assume the mid values in the range. Similarly the ESI for edges also falls in the mid-way of its range. It is observed that the ESI lies within $[\frac{1}{N}\ 1]$. The image features such as edges assume a value within $[0.5, \alpha_{high}]$. The upper limit of the feature value

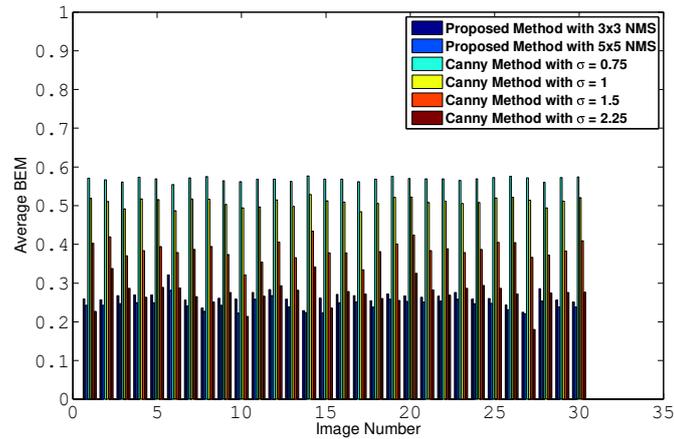

(a)

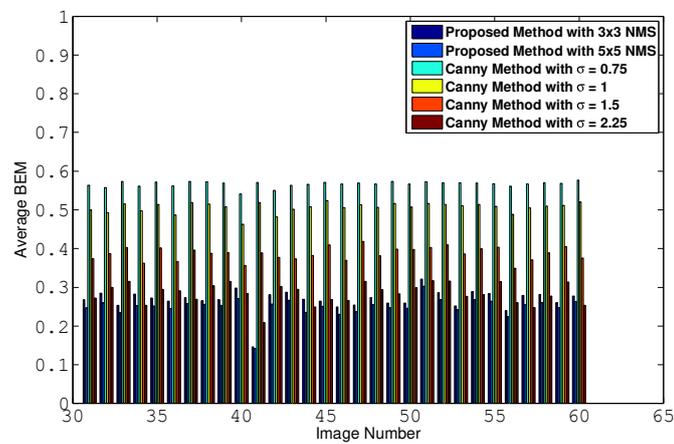

(b)

Figure 5.17: BEM on BSDS Image Data Set with Noise of 4 dB SNR for Proposed Method with $3 \times 3$ NMS, $5 \times 5$ NMS, Canny Edge Detector with Different Scale ($\sigma$)

determines the contrast level across weak edge that can be detected by the method. Empirically, it can be found out that $\alpha_{high}$ may be varied in the range of $[0.85, 0.9996]$.

The proposed algorithm has been tested on the images as shown in the Figure 5.12(a)-5.13(a). These figures consist of different types of edges (such as grating, curve etc.) embedded in varying contrast regions. The results obtained from Canny's method have been shown in Figure 5.12(b)-5.12(d), and Figure 5.13(b)-5.13(b) for

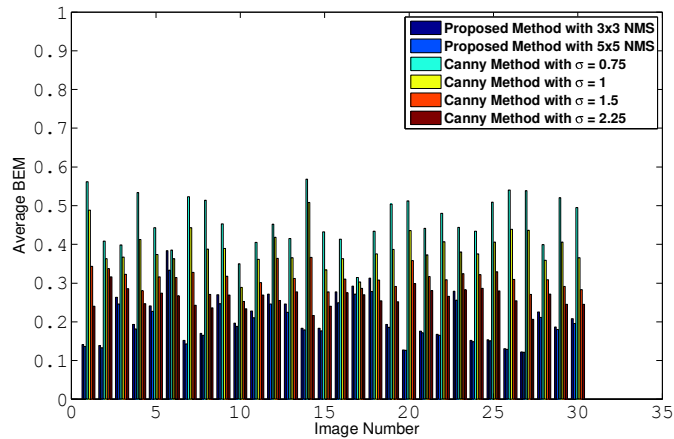

(a)

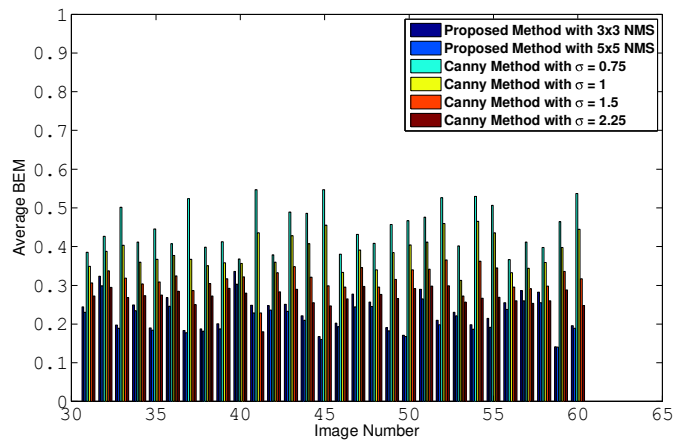

(b)

Figure 5.18: BEM on BSDS Image Data Set with Noise of 20 dB SNR for Proposed Method with $3 \times 3$ NMS, $5 \times 5$ NMS, Canny Edge Detector with Different Scale ($\sigma$)

different scales of the smoothing filter.

In Figure 5.12(e) and Figure 5.13(f), the edge maps by the proposed method without the non minimum suppression are presented. These figures show that our algorithm is capable of capturing all the required edges. However, widths of these edges are not of single pixel. Therefore, the NMS method has been applied within $3 \times 3$ as well as $5 \times 5$ moving region to obtain edge-maps as presented in Figure 5.12(f)-

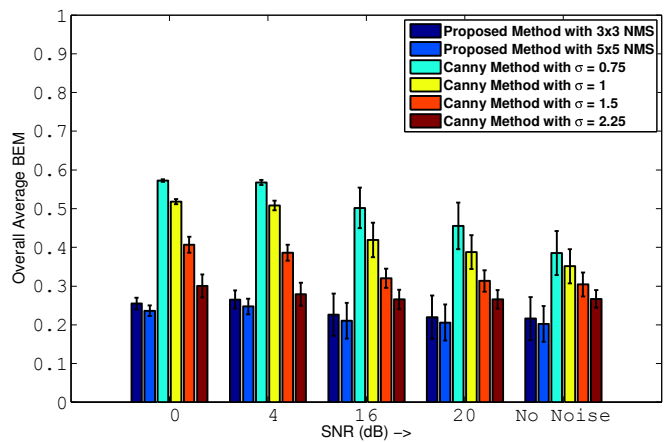

Figure 5.19: Overall Average BEM on BSDS Image Data Set Different Noise Condition for Proposed Method with $3 \times 3$ NMS, $5 \times 5$ NMS, Canny Edge Detector with Different Scale ($\sigma$)

5.12(g), and Figure 5.13(g)-5.13(h).

In Figure 5.12 it can be observed that Canny's method produces multiple edges around the true edge even after non-maximum suppression. This is because the edges in the original image change their edginess from step to line for which the Canny detector may not be effective [118–120] even after extensive experiments to find possible improvements with change of scales. It is found that our algorithm finds a single edge for the same image as it only measures the local energy variation instead of the gradient measure. It should also be noted that the edges in such cases as well as for the other images may not be localized accurately as the smoothing scale is increased. This fact can be observed with a scale of 3 in the images as shown in Figure 5.13(e). On the other hand spurious edges are observed when the scale is reduced. From Figure 5.13 it is observed that the best results are obtained by Canny's method with the scale of 1 and 0.75 respectively. Thus scale of the pre-filter affects the output of the detector. It is noticed that the proposed method with $5 \times 5$ NMS produces better results than that of $3 \times 3$ NMS. These also demonstrate that the proposed method find better edge map compared to those obtained by Canny's edge detector.

The robustness of the proposed algorithm is examined in presence of zero-mean

i.i.d. additive noise. Figure 5.14 shows the results of both these methods on the curved edge image with 15 dB SNR. In this case it can be noticed that Canny's method produces higher false negative for all scales as well as higher localization errors for the higher scale. The better performance of our algorithm can be attributed to the integral nature of the proposed edge index. However, the index is not fully insensitive to such noises. In presence of noise, it is observed that the upper limit is not affected while the lower limit of the ESI is further reduced. It can easily be shown to be determined by the factor $SNR^{-1}$. So a noise compensation of the index is required. The results are shown in Figure 5.14(g)-5.14(h) employing the compensation method proposed in the algorithm. The edges may be localized more accurately by the proposed method since it does not require any pre-smoother.

The quantitative assessment is performed for the proposed algorithm on randomly chosen 60 images of BSDS [128]. The ground truth images available for each of the image data compared with the edge map obtained by the proposed as well as Canny algorithm. The Baddeley Error Metric for binary images has been chosen for the purpose. The BEM values are measured for each of the available reference images. The average BEM for Canny's method with different scales as well as the proposed algorithm with $3 \times 3$ and $5 \times 5$ NMS are compared. The results are shown in Figure 5.16-5.18. Figure 5.16 is yielded for image without noise while the other two figures show the average BEM for the images with noise of 4 dB and 20 dB SNR. The average BEM over the whole data set with different noise conditions is presented in Figure 5.19 to show the effectiveness of the proposed algorithm. Consistent low BEM is observed for the proposed algorithm as compared to the Canny edge detector. The upper bound of ESI is chosen to be equal to 0.9. However, it can be noted that changes in the upper bound of the index may influence the results as it will either introduce or omit a few weak edges. The decision on the upper bound of the feature index can be made based on the context i.e. how far it is required to detect and localized the weak edges.

It can be observed from the figures that the average BEM of Canny detector reduces as the scale increases. The results obtained for the Canny edge detector may be attributed by the uncertainty between the detection and localization of edges [114, 126]. On the other hand, the proposed algorithm does not use any pre filter, instead noise compensation is made in the feature image during post processing.

Thus it avoids the localization error and provides better BEM as compared to the Canny's detector.

## 5.5 Case Study-II: Pupil Center Detection and Diameter Measurement using Directional Form Factor

### 5.5.1 Pupil Center Detection

Literature suggests that saccadic ratio can be associated with vigilance level of human operators [137, 138]. SR is defined as the ratio between peak saccadic velocity and saccadic duration. Thus, measurement of this parameter requires determination of position of pupil (or iris) center. Subsequently, a model based estimation of saccadic velocity can be applied on the pupil center positional data. However, the variation of illumination, orientation of eye, partial occlusion of the iris by eyelid, noise from the capturing device (viz. webcam) etc. may pose difficulty in its determination.

In the past few years, many camera based eye detection techniques have been developed [139–142]. These techniques usually detect the face region from an image and then proceed with eye detection. Although a few schemes have been proven to be effective for images captured under active near IR illumination [143], the detection rate may be degraded due to poor retinal reflection to NIR during day time. Robust eye detection schemes for color images have also been reported lately [144]. However, color image processing is computationally expensive and is not feasible in real-time applications. Hence, the discussion is restricted to gray scale and NIR illuminated images. Measures for eye pupil position detection have been provided in [141]. However the scheme relies on edge detection technique, which may not work well in the presence of clutter in the image or illumination variance. In [145, 146], the authors try to detect the pupil position based on Circular Hough Transform (CHT) or similar circle detection measures, which might not be always feasible especially when the eyes are partially occluded. Moreover, these methods may fail for images of poor quality caused by vehicle jerks and variation in illumination. Therefore, a robust image processing algorithm is required to find pupil center.

In this case study form factor has been applied in eye image along horizontal as well as vertical directions to find the pupil center. The values around maxima of the horizontal and vertical FF are used to determine the coordinates of center of pupil or iris. The eye images are initially gamma-corrected to remove spurious pixels due to improper illumination. The effectiveness of our approach over CHT based methods is demonstrated on standard image databases as well as our own eye image database which includes both gray scale and NIR monochrome images.

#### 5.5.1.1 Form Factor and the Pupil center Detection

It can be observed that the image of an eye consists of iris with lower intensity zone around which there is white sclera region. This makes the eye image a high contrast zone. Since the iris consists of lower intensity pixels, higher form factor will be observed in this zone. The FF in horizontal direction is computed on each column of the image. The peak value of the horizontal form factor is generated at the center of the iris. Similarly, the vertical FF is computed on each row of the same image and thus the vertical position of the center of iris is obtained.

In practice, the eye images are prone to reflect ambient object during daylight and the NIR led image during night driving condition. This may shift the position of the peak of the FF and thus it introduces error in the estimation of pupil center. To overcome this problem, a zone is selected around the position of the peak spanned over 20% of the eye image width on both sides. Subsequently, the FF values that are 1% less than that of the peak FF are chosen. Figure 5.20 shows the selected FF in horizontal and vertical direction.

The center of pupil is obtained by determining the position of the Center Of Mass (COM) of the selected FF values as shown in (5.32). The parameters that decide the selection of FF values are based on the relative size of the parts of the eye and empirical understanding.

$$[x_{pc}, y_{pc}] = \left[ \frac{\sum_{i=1}^{N_x} F_{hz}(i) x_I(i)}{\sum_{i=1}^{N_x} F_{hz}(i)}, \frac{\sum_{i=1}^{N_y} F_{vt}(i) y_I(i)}{\sum_{i=1}^{N_y} F_{vt}(i)} \right] \quad (5.32)$$

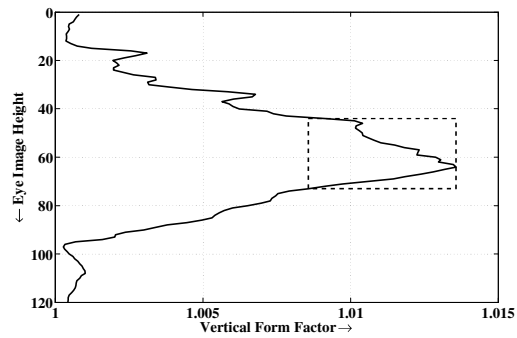

(a)

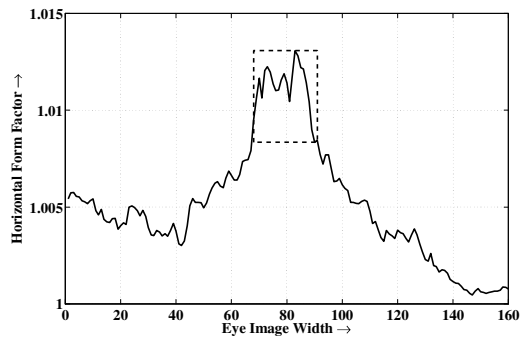

(b)

Figure 5.20: a) Horizontal and b) Vertical Form Factor Variation of an Eye Image with Selected Zone around Peak for COM Computation

Where, $F_{hz}$ and $F_{vt}$ are the form factor values in horizontal and vertical directions respectively. $N_x$ is the number of FF values selected from $F_{hz}$ while that in vertical form factor is $N_y$.

Apart from the above challenge, there may be spurious low intensity pixels due to poor illumination which may result incorrect estimation of pupil center. Therefore the eye images are gamma corrected before the form factors are calculated. Gamma value in the range of 2 to 3 is chosen for gray scale data sets while a gamma value of 1.5 is used for NIR databases in the experiments carried out. In general, a value of gamma greater than unity is needed to lighten the image. On the other hand the noise contaminated FF values can be recovered under zero-mean i.i.d. noise ( from (5.5) and (5.6)). This requires a blind estimation of STVR (or SNR). The detail of such estimation has been presented in section 5.3.

#### 5.5.1.2 Experimental Results

The proposed method is tested against standard data sets available on the internet as well as test images captured in driving conditions. The databases include both the gray scale and NIR monochrome images. Gray scale eye images are obtained from the IMM-face database [58] and BioID face database [59]. The IMM-database has a total of 240 images of 40 different people, while BioID database contains 1521 images of 23 test persons under different illumination conditions. All the 240 face images in the IMM-face database and 920 face images (40 images per person) from the BioID database have been used for testing the proposed algorithm. The CBSR NIR image database [60] is used for testing the algorithm on NIR eye images. The method is tested on 800 images from the NIR face database. The other images are from our own databases recorded under simulated driving environment in the laboratory. The video covers two drivers in actual illumination settings. The video was recorded under varying illumination condition through both day and night time. Out of this database 800 frames are used to obtain eye images. A few sample eye images from each database are provided in Figure 5.21.

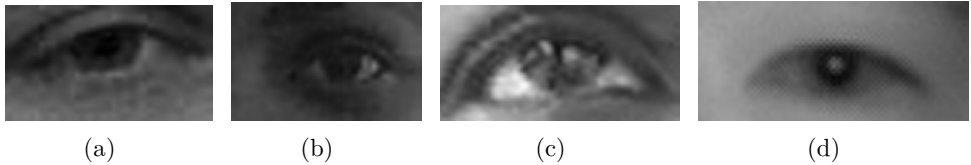

(a)    (b)    (c)    (d)

Figure 5.21: Sample Eye Images Extracted from Different Data Sets used for Testing. a) IMM-Face Database, b) BioID Face Database, c) IITKGP Database and d) CBSR NIR Database

The pupil center coordinates are available in all the three annotated face databases (IMM, BioID, and CBSR) and are considered as the benchmark. The pupil coordinates were manually marked several times by three persons for each image in our own database (IITKGP database) and their average is considered as the true center of the pupil. The eye images were resized to $25 \times 50$ in the gray scale image databases and $90 \times 40$ in the NIR face database in order to enable a uniform comparison throughout the respective databases.

The performance is evaluated by marking the pupil center coordinates in each image using the proposed algorithm and testing if the marked coordinate is close to the true pupil center. An image is tagged as correctly marked, if the coordinates found by the method lie within an error circle of radius $r_e$ pixels centered at the true pupil center. The value of $r_e$ is set to 4 pixels for gray scale and 7 pixels for IR images (within 10% of the diagonal length of the image). The reflectance characteristics of iris under visual and NIR light are different. This difference may affect the contrast within eye region and thus lead to set the different error radii in these types of images. The percentages of images tagged correctly by the proposed method by searching the position of peak as well as by computation the center of mass of FF values around the peak and that by CHT are presented in Table 5.3.

Table 5.3: Results Showing Percentage(%) of Correctly Tagged Images for Various Eye Image Databases

| Algorithms | | % of correctly tagged images from databases | | | |
|---|---|---|---|---|---|
| | | IMM | BioId | IITKGP | CBSR NIR |
| Circular Hough Transform | | 64.37 | 77.96 | 55.00 | 1.875 |
| Proposed | Peak FF based | 88.12 | 91.44 | 92.75 | 93.75 |
| | COM near peak FF based | 96.22 | 98.14 | 98.85 | 99.17 |

### 5.5.1.3 Discussions

Table 5.3 shows the percentage of correctly tagged images from various databases to demonstrate the efficiency of the proposed algorithm. A comparative study with traditional Hough transform based circle fitting method is also presented. It can be observed that the COM based proposed algorithm posses the higher accuracy in locating the pupil center in the databases. The Hough transform method poorly detects pupil center in the NIR database. It indicates that the CHT based method may not work with smeared edges around the pupil (or iris). A few sample results are shown in Figure 5.22. The figure also shows a plot of the vertical and horizontal FF values for the different cases. Sample results of the proposed algorithm on partially occluded images are also shown in Figure 5.23. In Figure 5.24 our proposed method are compared with that using the CHT. It can be observed that the later method fails to determine pupil center successfully.

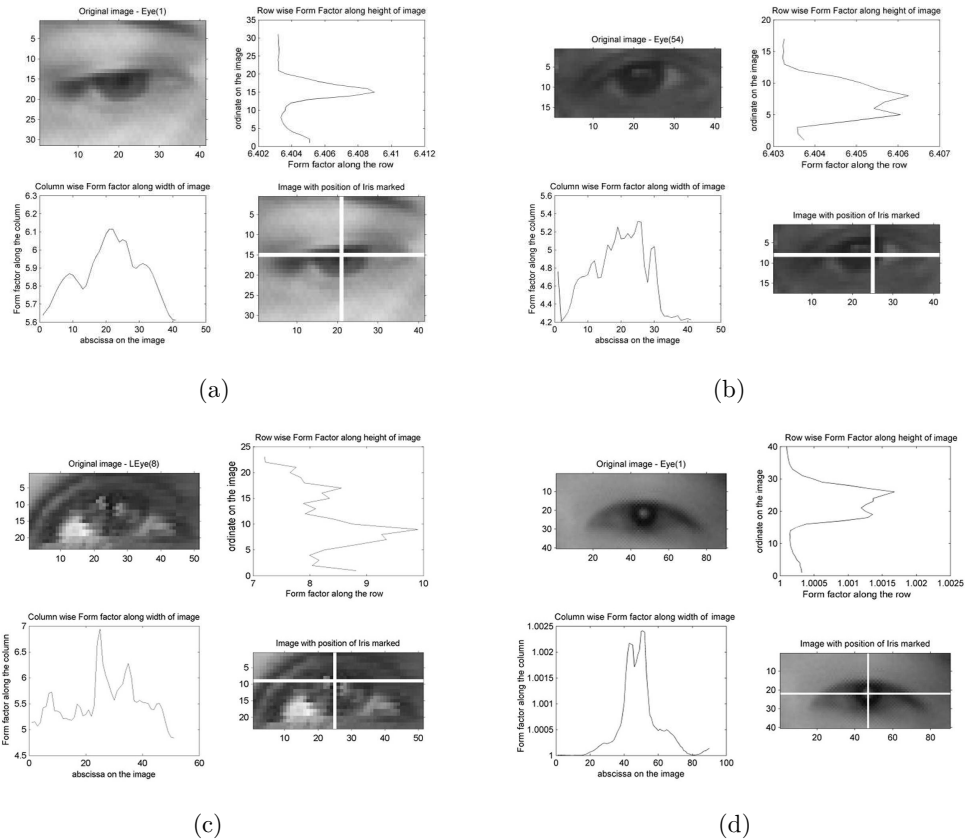

Figure 5.22: Sample Results for Different Data Sets used for Testing. The Plot of FF along the Row and Column along with the Iris Center Marked by Our Method. a) IMM-Face Database, b) BioID Face Database, c) IITKGP Database and d) CBSR NIR Database

### 5.5.2 Pupil Diameter Measurement

Literature suggests that dilation of pupil may reflect the cortical processing associated with onset of saccade [147]. Therefore, measurement of pupil diameter can be used to determine the inception of saccadic eye movement in advance. The radial FF which is computed from the polar image of eyes has been proposed to measure the pupil diameter per frame. This measurement may aid to decide the state of alertness in human drivers. The pupil diameter is determined manually by 3 persons with repeated trial for the eyes taken from IMM, BioID, UBIRIS.v2, CBSR NIR and IITKGP image database. These are averaged for each of the eyes to produce ground truth for pupil diameter measurement. Again the reflection from the pupil region prevents to consider the boundary of pupil as the position of peak radial FF. Therefore, a region around peak of radial FF are selected to compute their COM which indicates the radius (so the diameter) of the pupil. The polar image of eye and the radial FF with selected radius zone are shown in Figure 5.25. The constraint in radial distance and that in FF values for the region of concern are decided by relative size of the parts of eye. The percentage errors of the proposed algorithm are shown in Table 5.4.

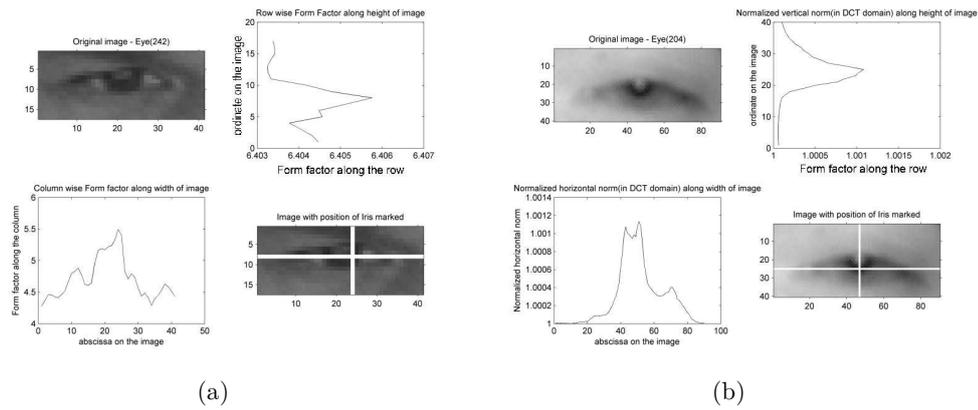

Figure 5.23: Sample Results of the Proposed Method for Partially Occluded Eye Images. a) IMM-Face Database, b) CBSR NIR Database

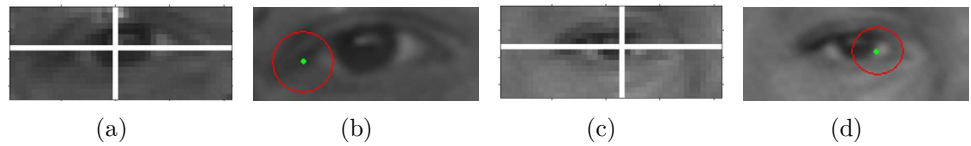

Figure 5.24: Sample Results where the Proposed Method Outperforms the Hough circle Detection Based Approach. (a) and (c) The Proposed Method (b)and(d)The CHT Based Method

Table 5.4: Percentage Error in Measuring Pupil Diameter for Image Databases

| Database | IMM | BioId | UBIRIS v.1 | IITKGP | CBSR NIR |
|---|---|---|---|---|---|
| Percentage Error(%) | $5.12 \pm 0.83$ | $5.67 \pm 1.02$ | $6.67 \pm 1.21$ | $8.34 \pm 0.25$ | $7.56 \pm 2.13$ |

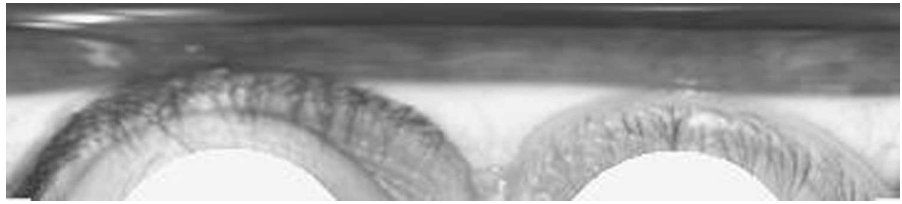

(a)

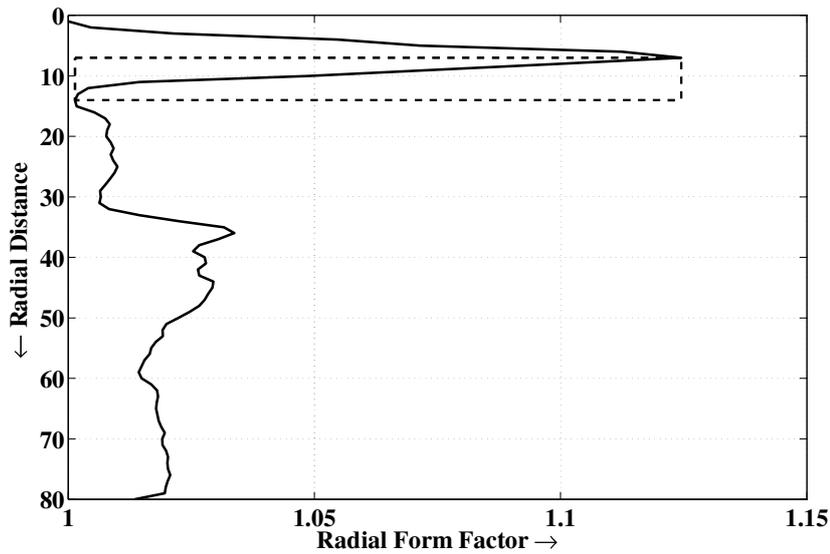

(b)

Figure 5.25: a) Polar Eye Image and b) Radial FF Variation with Selected Zone Near Peak for COM Computation

The experimental results of the proposed pupil diameter method on various eye image database demonstrate its effectiveness under different challenges that may exist on road driving situation through day and night.

## 5.6 Conclusion

A new image feature called Form Factor has been introduced. The feature can be recovered blindly under additive zero-mean i.i.d. noise with an estimation of SNR or STVR. Blind estimations of SNR and STVR by various methods have been analyzed. Both theoretical analysis and experimental results endorse the merit of STVR over SNR and other index such as TVNR. It is observed that the method based on horizontal extrapolation of ACS yields best estimation of STVR and thus recovers FF with higher accuracy. Although the recovery scheme of FF under the zero-mean i.i.d. noise has been presented, there are other variations like multiplicative noise in practice. Exploration of these features under such types of noise can be carried out in future. The application of FF is explored for two cases.

In Case Study I, local FF has been applied to measure local energy in an image region. A novel edge detection algorithm based on this measure has been proposed. The edge strength computed as inverse of the square of FF. The novelty of this index can be summarized as below-

- It is a dimensionless and normalized index.
- Integral nature (or derivative free nature) of the index makes it less sensitive to noise.
- The effect of noise can easily be alleviated using an estimation of STVR for zero-mean i.i.d. noise.
- It does not require any pre-smoothing. Thus the uncertainty between edge detection and localization is avoided and better accuracy can be achieved.

An NMS algorithm along the direction of lowest local variance has been developed to eliminate the multiple responses to an edge. A consistent low BEM over the BSDS images demonstrates the efficacy of the proposed algorithm over the popular Canny edge detector. Finally, it is observed from these measurements that our algorithm with $5 \times 5$ NMS produces best results on an average throughout the image data set. The experimental validation proves its effectiveness for detection of edges in natural images under different noise conditions.

Pupil (or iris) center and diameter has been detected using directional form factor which is elaborated in Case Study II. The algorithm has been tested on four databases of eye images. These images are gamma corrected and subsequently the horizontal and vertical Form Factors are computed to find the center of pupil. Its coordinate is determined as the COM of form factor values around the peak FF. It has been found that the proposed methods out performs the CHT method in various conditions like varying illumination, partial occlusion of pupil (or iris), noise, reflection etc. The CHT based method depends on edge detection algorithms which may lead to error when spurious edges occur due to noise. Moreover, pre-filtering of these noises may shift the iris edge and thus produces error in detection of its center. The experimental results show that the proposed method can be useful for processing of eye images in both day light and night driving conditions.

Computation of SR requires the positional information of pupil center over time. The algorithms developed in this chapter can be used to generate the required information. The method has been elaborated in the next chapter.

## Chapter 6

# Detection and Interpretation of Saccadic Eye Movement from Image Sequence Analysis

## 6.1 Introduction

The pattern of eye movement might be informative about the cognitive state of mind and the level of alertness in human subjects [46, 48–50]. Human alertness monitoring has been studied using both invasive biological signal indicators which require physical contact with the user and non-invasive camera based eye monitoring systems. Invasive systems have shown to provide higher accuracy between the two but continuous contact with the body makes them impractical for application in daily working contexts such as the need for vehicle driving. On the contrary, camera based techniques which capture human eye motion are indirect, but convenient. In everyday context, eye movement patterns consist of fixation, smooth pursuit and saccade. Literature suggests that the saccadic movement of eye [137, 138] pattern of saccade fixation and smooth pursuit [148] might be indicative of visual attention. It has also been seen that the velocity profile of these movements possess variability in peak velocity, saccadic duration, time to reach peak velocity etc. [149].Thus estimation of velocity profile of pupil may facilitate detection of the pattern of eye movement over time.

It has been found by the researchers that the occurrence of natural saccade falls within $15° - 20°$ [51]. It has also been reported that the bandwidth of the naturally-occurring saccades remains well within $25 - 30$ Hz [51–53]. The saccadic movement has been observed to become sluggish with increase in eye muscle fatigue. Hence the bandlimit of this motion during fatigue can be considered as 30 Hz. Thus the video frame rate of 60 fps or more may be useful for capturing eye saccade. In this phase of work, the high speed camera with speed within the range of $60 - 500$ fps is used to capture the eye saccade of human subjects.

The aim in this chapter is to develop a video based measurement technique of Saccadic Ratio (SR) and to apply it for determination of alertness level. SR is the ratio between Peak Saccadic Velocity (PSV) and the Saccadic Duration (SCD). Thus the estimation of saccadic ratio requires the determination of PSV and SCD. Both of these quantities can be obtained from the saccadic velocity profile. However, the velocity profile can be determined from the temporal information of positions of pupil (or iris) center. Therefore a robust algorithm for detection of pupil center that can be used under varying illumination, free movement of operator etc. is necessary.

A method based on Directional Form Factor (DFF) has been developed for finding out the center of pupil. The method is explained and shown to be effective for both daylight and night driving conditions. The algorithm can be applied on the eye images detected per frame. Saccadic movement is analyzed for the fully and partially open eye images as simultaneous occurrence of eye closure and saccade is rare. The eye image in each frame is detected as open or partially open (or partially closed) by DCT based correlation filter as explained in Chapter 4. It is observed from the IITKGP database and survey with other vehicle operators that head posture during driving may vary along three degrees of freedom. It can be noted that the horizontal saccade made by the subject may not be detected as horizontal by the camera for in-plane (in the image plane) and off-plane (inclined on steering) rotated faces. Thus, horizontal movement of pupil with respect to image plane axis with origin at middle may not give the correct measure of horizontal saccade. However, it can easily be noticed that the horizontal saccade occurs along the line joining the two eye corners. So, the problem can be resolved if the eye corner points are considered as reference and the position of the pupil center is expressed with respect to the mid-point between the corners. The corners of eye is detected from the eye edge map obtained using the Local Form Factor (LFF) algorithm.

The position information of eye center with respect to the eye corners can be used by a real-time estimator like Kalman Filter (KF) to track the horizontal saccade. The saccadic velocity can directly be obtained from such estimator. The onset of saccade in any direction (left or right) can be found if slope of velocity profile changes from zero to non-zero values. Subsequently, the peak of the velocity is obtained at the point when slope changes its sign after the inception of saccade. The end of saccade is indicated by the change of slope of the profile from non-zero to zero. In practice, errors may be introduced in such measurement by noise retained in the eye images. The independent and identically distributed (i.i.d.) noise in images can be obviated during the computation of Form Factor (FF). The recovery of this feature under i.i.d. noise has been explained in Chapter 5. The measurement uncertainty is reduced by using the state space filters like KF or Extended Kalman filter (EKF).

The performance of the algorithm has been analyzed in two phases for both simulated eye movement and actual eye movement data from the database generated from the experiments. The estimated position and velocity by the proposed algorithm have been compared with those obtained from trained manual scorer. The pupil center and the corners detected by a few scorers are used to yield the reference pupil center relative to eye corners. The inter subjective estimation error is reduced by taking the average of the position of pupil center for similar states of eye. The reference velocity profile is obtained by commonly used 5-tap or 7-tap Finite Impulse Response (FIR) differentiator [148]. However, it can easily be noticed that the KF or EKF estimators are robust under noise and faster than FIR differentiator as in later case larger past sample values are required to produce the output. The instantaneous error Probability Density Functions (pdf) for estimated position and velocity profile by KF and EKF have been compared. The model of the eye movement and detailed estimation methods have been explained in the following sections. Subsequently, the alertness level detected by the proposed algorithm is compared with the subjective assessment for experiment III .

## 6.2 Estimation of Saccadic Velocity

Estimation of saccadic velocity includes three basic components. These are as follows-

1. Determining initial estimate using model of eye saccade
2. Detection of pupil (or iris) center relative to its corners as initial state
3. Estimation of velocity at current instant

## 6.2.1   Eye Saccade Model

Several investigations related to modeling saccadic motion have been reported [150]. All these attempts consider the eye along with ocular muscles as oculomotor plants. The relation between the velocity signal generated in brain and the position of pupil in 3-D space is found to be non-linear in nature. These models are useful for understanding the interaction of brain with the outer world through visual system in human subject. However, measuring the saccade from the sequence of frames is the inverse solution to the oculomotor plant model. Here, the velocity is estimated from the temporal information of iris position. In addition to this, projective geometric transformation maps 3-D objects into their 2-D images. Thus an indirect measurement of saccadic parameter like SR may be feasible from 2-D eye image. Although extensive effort has been put on development and analysis of the oculomotor plant model, a few research articles exist on such measurements [54]. In [54] projective geometric parameters like distance between camera and the subject are exploited to get the iris center in 3-D space. However, these parameters may vary in large extent during driving, causing inaccuracies in measurement.

Real-time estimation demands the indirect model to be computationally inexpensive and preferably linear. Moreover, elimination of the projective geometric parameter may improve the required estimation. Therefore, a model considering pupil center as an object in motion with constant acceleration is used to track the pupil. Literature defines the eye saccade as the fast and simultaneous jump of eye gaze from one point of interest to the other. During the saccadic movement, horizontal and vertical directions are used out of the three degrees of freedom of eye co-ordinate system [150]. It is observed that horizontal saccade occurs more frequently than the vertical one during driving. In this research, the horizontal saccade has been considered for measurement of saccadic ratio. In rest of the chapter, the term saccade is used to imply horizontal saccade. The models for both horizontal and vertical direction are same for the pupil center, and corners with their respective co-ordinates.

In the proposed problem $x(t)$ is assumed to be the continuous time horizontal (or $y(t)$ for vertical) position function of the pupil center relative to the eye corners with first and second order derivatives $x'(t)$ and $x''(t)$ respectively. Let $\Delta t$ be the time interval between two consecutive video frames, so that for smaller value of it, the position and velocity vectors at $k^{th}$ instant $x_k = f(k\Delta t)$ and $x'_k = x'(k\Delta t)$ are governed by the equation (6.1).

$$\begin{aligned} x(k) &= x(k-1) + x'(k-1)\Delta t + \tfrac{1}{2}x''(k-1)(\Delta t)^2 \\ x'(k) &= x'(k-1) + x''(k-1)\Delta t \end{aligned} \quad (6.1)$$

Where, $x''(k) = x''(k\Delta t)$ is the acceleration of iris at $k^{th}$ instant. The state space representation of this system is as shown in (6.2).

$$\mathbf{x}(k) = \mathbf{F}\mathbf{x}(k-1) + \mathbf{G}(k-1)\mathbf{u}(k-1) + \zeta(k-1) \quad (6.2)$$

Where, $\zeta(k)$ represents the uncertainty in system model at $k^{th}$ instant. $\mathbf{x}(k) = [x(k) \ x'(k)]^T$ represents the state of system . $\mathbf{u}(k)$ is the constant acceleration of magnitude unity and $\mathbf{G}(k) = [0.5(\Delta t)^2 \ \Delta t]^T$.

The observations can be modeled as shown in (6.3).

$$\begin{aligned} \mathbf{z}(k) &= \mathbf{H}\mathbf{x}(k-1) + \eta(k-1) \\ \mathbf{H} &= [1 \ 0] \end{aligned} \quad (6.3)$$

Where, $\eta(k)$ is the uncertainty in observed pupil position at $k^{th}$ instant.

### 6.2.2 Detection of Eye Center and Corner

The position of pupil center at any instant is estimated from its recent past values as can be seen from (6.1). Therefore initial value of the position is necessary at zeroth instant. In this work, the eye image is extracted using popular Haar-Like Feature (HLF) based object detection method. Subsequently the center of pupil is determined using FF based method as explained in Chapter 5. However, the co-ordinates are given in pixel position in absolute sense. The correspondence between this absolute position and that in 3-D space can be found through projective geometric parameters. However, the algorithm has higher computational complexity. In addition to this, inaccuracies may be introduced due to variation in these parameters. Hence, a relative measurement of pupil position with respect to other fixed reference point(s) may be appropriate to alleviate the problem. In eye images, corners might be considered as fixed reference points.

The common issues in image processing (like variation in illumination) also pose difficulty in finding the correct corner points. In literature, two modes of algorithms based on edge map and patch analysis are available to find corner points [151]. These are summarized in Table 6.1.

Table 6.1: Corner Description Algorithm [151]

| Category based on | Computational Methodology |
|---|---|
| Curvature of edge | 1. The curvature is computed as change in angular direction around an edge point.<br>2. The curvature is calculated as the changes in the image intensity along the edge.<br>3. Curvature is computed as minimum average change in image intensity along different directions around a pixel. |
| Patch analysis by local operator | 1. Salient feature is detected by scale invariant feature transform.<br>2. Saliency operator is used to find the robust and relevant feature in a local region based on computation of entropy. |

Detail analysis of these algorithms reveals that a pre-filter is required to remove the noise. However, pre-filtering by Gaussian operator may shift the true position of feature of interest such as edge, corner points etc. A multi-scale based method for such feature extraction may reduce error with increase in computational complexity.

An edge detection method by FF has been proposed as explained in Case Study I of Chapter 5. The advantage of this feature is that it does not require any pre-filter to remove the additive zero-mean i.i.d. additive noise. Instead, a blind estimation of STVR can be used to recover the FF. Empirical results revealed that the edge strength (inverse of square of FF) at corner points assumes a value close to 0.5. This indicates that the corner points correspond to square of ratio of average to peak pixel values. The ratio is actually the inverse square of crest factor. Thus the algorithm identifies the corner when Edge Strength Index (ESI) is close to 0.5 on the connected pixel forming the edge.

The algorithm searches two ends of the image for the corner points of the eye. The region of interest is decided empirically as the 2% of total eye image width in horizontal direction. The detected corners are assumed to span the range of the view angle ($\pm 60°$) with its mid-point as the zeroth angle. Thus the angular position of the eye center is computed with respect to the corners at every instant (shown in (6.4)) and it is used to estimate the saccadic velocity.

$$x(k) = \sqrt{\left(x_{cent}(k) - \frac{x_{cornL}(k) + x_{cornR}(k)}{2}\right)^2 + \left(y_{cent}(k) - \frac{y_{cornL}(k) + y_{cornR}(k)}{2}\right)^2} \tag{6.4}$$

Where, $\{x_{cent}(k), y_{cent}(k)\}$, $\{x_{cornL}(k), y_{cornL}(k)\}$, and $\{x_{cornR}(k), y_{cornR}(k)\}$ are the co-ordinate of center, left and right corners of eye respectively at particular instant '$k$'.

### 6.2.3 Real Time Estimation Methods

#### 6.2.3.1 Linear Kalman Filter

The first step of the KF is the prediction of states from the eye saccade model as shown in (6.5). Measurements such as saccadic position, can be predicted from an observation model following (6.7).

$$\hat{\mathbf{x}}(k/k-1) = \mathbf{F}(k-1)\hat{\mathbf{x}}(k-1/k-1) + \mathbf{G}(k-1)\mathbf{u}(k-1) \tag{6.5}$$

$$\hat{\mathbf{z}}(k/k-1) = \mathbf{H}(k-1)\hat{\mathbf{x}}(k-1/k-1) \tag{6.6}$$

It is found that the error in estimation of state at any instant ($k$) from the predicted state is proportional to the difference between predicted measurement and the current observation ($\mathbf{z}(k)$). Thus the state can be updated at any instant by (6.7).

$$\hat{\mathbf{x}}(k/k) = \hat{\mathbf{x}}(k/k-1) + \mathbf{K}(k)\left[\mathbf{z}(k) - \hat{\mathbf{z}}(k/k-1)\right] \tag{6.7}$$

where, $\mathbf{K}(k)$ is the Kalman gain at $k^{th}$ instant. This parameter can be computed from the state covariance matrix $(\mathbf{P}^f(0))$ as shown in (6.8).

$$\begin{aligned}
\mathbf{P}^f(0) &= Var(\mathbf{x}(0)) \\
\mathbf{P}^f(k/k-1) &= \mathbf{F}\mathbf{P}^f(k-1)\mathbf{F}^T + \mathbf{Q}(k-1) \\
\mathbf{K}(k) &= \mathbf{P}^f(k/k-1)\mathbf{H}^T \left(\mathbf{H}\mathbf{P}^f(k/k-1)\mathbf{H}^T + \mathbf{R}(k)\right)^{-1} \\
\mathbf{P}^f(k) &= (\mathbf{I} - \mathbf{K}(k)\mathbf{H})\,\mathbf{P}^f(k/k-1)
\end{aligned} \quad (6.8)$$

Where, $\mathbf{Q}(k)$ and $\mathbf{R}(k)$ are the covariances of state prediction and measurement error respectively.

In this subsection the saccadic eye movement is assumed to be linear. However, the generation of saccadic motion has been modeled as non-linear plant [152]. This has led to investigate the estimation of saccadic velocity by EKF.

### 6.2.3.2 Extended Kalman Filter

In EKF, the model is linearized near currently estimated state variable. The initial estimation is obtained with a prior guess of states ($\hat{\mathbf{x}}(0) = [0\ \ 0]^T$) and its error covariance ($\mathbf{P}^f(0)$). The current state and the covariance can be forecasted using (6.9). The unbiased estimate of the state is obtained from the predicted state, measured position ($\mathbf{z}(k)$), and the measurement uncertainties at $k^th$ instant by (6.10).

$$\begin{aligned}
\hat{\mathbf{x}}(k/k-1) &= \mathbf{f}(\hat{\mathbf{x}}(k-1/k-1)) \\
\mathbf{P}^f(k/k-1) &= \mathbf{J}_f(\hat{\mathbf{x}}(k-1/k-1))\mathbf{P}^f(k-1)\mathbf{J}_f^T(\hat{\mathbf{x}}(k-1/k-1)) + \mathbf{Q}(k-1)
\end{aligned} \quad (6.9)$$

$$\begin{aligned}
\hat{\mathbf{x}}(k/k) &= \hat{\mathbf{x}}(k/k-1) + \mathbf{K}(k)(\mathbf{z}(k) - \mathbf{h}(\hat{\mathbf{x}}(k/k-1))) \\
\mathbf{K}(k) &= \mathbf{P}^f(k/k-1)\mathbf{J}_h^T(\hat{\mathbf{x}}(k/k-1))\left(\mathbf{J}_h(\hat{\mathbf{x}}(k/k-1))\mathbf{P}^f(k/k-1)\mathbf{J}_h^T(\hat{\mathbf{x}}(k/k-1)) + \mathbf{R}(k)\right)^{-1} \\
\mathbf{P}^f(k) &= (\mathbf{I} - \mathbf{K}(k)\mathbf{J}_h(\hat{\mathbf{x}}(k/k-1)))\,\mathbf{P}^f(k/k-1)
\end{aligned}$$
$$(6.10)$$

Where, $\mathbf{J}_f$ and $\mathbf{J}_h$ are the first order jacobian of state and the measured variable.

## 6.3 Experimental Results

In this section the algorithm's performance is tested in three modes. First, synthetic pupil position with respect to eye corners is generated by modeling saccade as step change in the pupil position. The estimated velocity from this data by KF and EKF are compared by the estimation error pdf. Similarly the state estimation filters along with the pupil position detector based on FF is applied on eye saccade video captured at higher frame rate ($60 - 1000$ fps). The saccadic ratio is estimated by EKF method and the results are compared with assessed alertness level based on a questionnaire survey.

### 6.3.1 Saccadic Velocity Estimation for Simulated Eye Movement

Literature indicates that saccadic eye movement can be simulated by a step change of small duration ($\leq 20$ ms) [51–53]. Moreover, the time interval between two consecutive saccades must be at least 200 ms. Other types of eye movements include fixation and smooth pursuit in everyday working conditions such as vehicle driving. Fixation occurs when the subject aligns the fovea towards the point of interest. However, during fixation the eye remains in micro-oscillation around the mean position. On the other hand, smooth pursuit occurs if a moving object is followed by the subject smoothly. Here, the velocity of the eye is same as that of the moving object. It is evident that smooth pursuit can be simulated by a ramp function when the velocity of the moving object is constant. These naturally-occurring eye movements have been simulated using step and ramp functions of random magnitude and slope to generate pupil position data. The model as mentioned in Subsection 6.2.1 is used to estimate the positions and velocity by KF and EKF. The pdfs of the estimation errors given by KF and EKF and the simulated signal for eye movement and the estimation error pdf are shown in Figure 6.1. The pdf of mean estimation error is produced for thousand randomly generated saccadic movements.

The velocity profiles for each of these cases are also estimated along with the position estimation. The envelops of these estimated velocities are compared with that obtained using 5-tap and 7-tap FIR filter method [148] on the reference pupil position. Subsequently, the profile match error of the velocities given by EKF and FIR filter based method is computed along with that between KF and the FIR filter based method. The probability density mean error in profile match is computed after compensating the time delay between the velocities obtained from the methods. These are shown in the Figure 6.2.

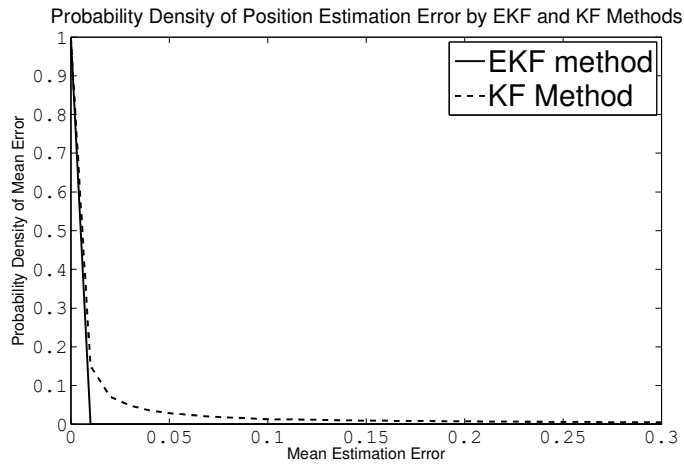

Figure 6.1: Probability density function of Mean Estimation Error in Pupil Position Obtained from KF and EKF Estimation on Simulated Eye Movement Sampled at 500 Hz

Figure 6.2 indicates that the motion of eye can be modeled as a nonlinear motion with fixed acceleration assumption. To support it further, the model has been tested on two eye movement videos produced from Experiment III.

### 6.3.2  Saccadic Velocity Estimation for Actual Eye Movement

A typical saccade of 30° lasts for at least 0.05 seconds. In order to get at least 5 data points in a 30° saccade, video should be captured from at least a 90 fps camera. In the present work, results are shown for two saccade videos recorded with 90 fps and 200 fps. The form factor based algorithm has been applied on both the videos to identify the position of the pupil in each frame. This position data was fed to the KF and EKF algorithms. The results of the estimated position and velocity data are shown for 90 fps in Figure 6.3(a). The velocity profile estimated for the video is shown in Figure 6.3(b). It is assumed that the eye corners span an angle of 30° from the center. Hence, any motion on the image is translated into an equivalent angular

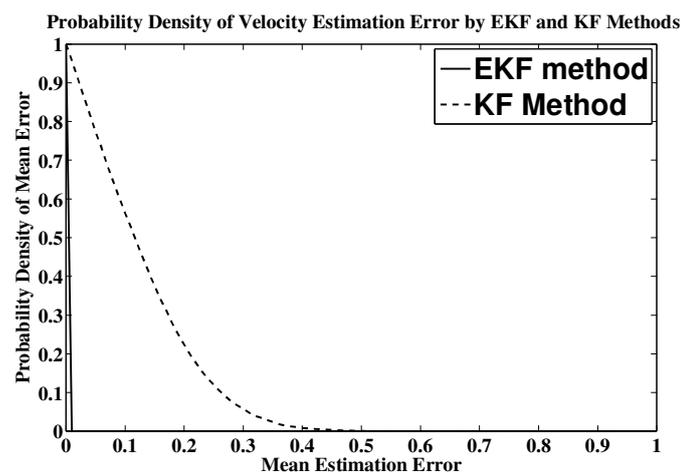

(a)

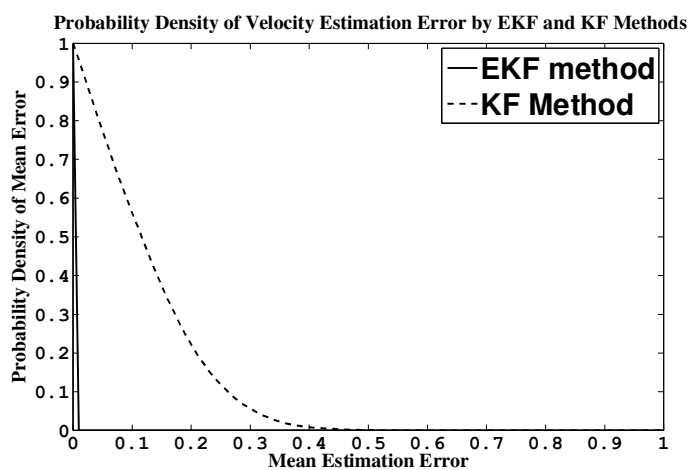

(b)

Figure 6.2: Probability density function of Mean Error in Velocity Profile Match between Reference Velocity Output of (a) 5-tap and (b) 7-tap FIR Filters and KF and EKF Based Estimations on Simulated Eye Movement Sampled at 500 Hz

motion by a suitable multiplying factor. It is to be noted that a change in this factor would not affect the velocity profile shape.

The saccade regions can easily be identified as those with sharp peak velocities and short duration. The first video covers two saccades, while the second video covers a single saccade. Data sets for reference position of pupil center and eye corners are generated by human scorers. Average of these data has been used to yield the relative pupil position over time. The results obtained from the proposed algorithm are compared with this reference set and the mean estimation error of pupil position and that in velocity profile match have been presented in Table. 6.2 and in Table. 6.3 respectively.

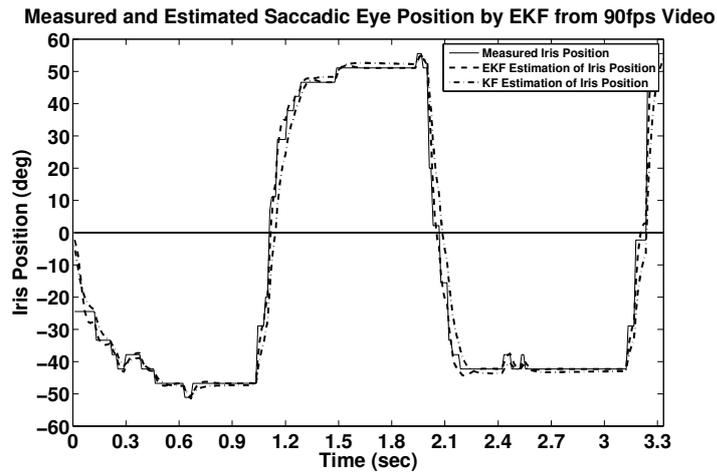

(a)

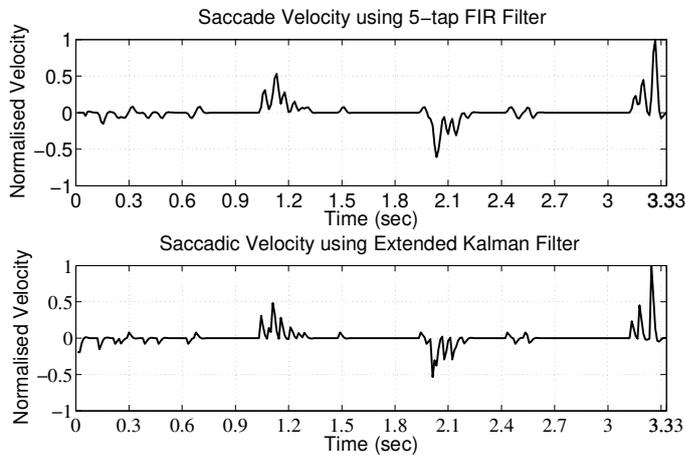

(b)

Figure 6.3: Results Showing (a) Saccadic Position, and (b) Saccadic Velocity

Table 6.2: Mean Pupil Position Estimation Error

| Resource of Pupil Position | Method of Estimation | |
|---|---|---|
| | EKF | KF |
| Video with 90 fps | −0.01614 | 0.08332 |
| Video with 200 fps | −0.00699 | −0.15882 |

Table 6.3: Mean Saccadic Velocity Profile Match Error

| Resource of Saccadic Velocity | FIR Filter used for Comparison | Method of Estimation | |
|---|---|---|---|
| | | EKF | KF |
| Video with 90 fps | 5-tap | 0.00639 | 0.03670 |
| | 7-tap | 0.00946 | 0.03977 |
| Video with 200 fps | 5-tap | 0.00458 | 0.30822 |
| | 7-tap | 0.00110 | 0.30473 |

### 6.3.3 Estimation of Saccadic Ratio and Alertness in Human Drivers

In this work, the relative position of pupil center with respect to two corner points are measured using the algorithm based on directional and local FF. This helps to avoid the object-camera geometry in the eye image analysis. SR is measured by measuring average of ratio between the peak velocities and the durations of each of the small saccades within a $5-10$ ms time interval. The measurement of SR has been carried out on three subjects and presented in the Table 6.4. The population is small due to the time and other constraints. However, the algorithm can be applied for large number of population. The output is compared with that of the subjective assessment of alertness levels. The subjective scores are computed as the weighted average of the responses to questionnaire explained in Chapter 2. Higher value of this score indicates higher level of alertness.

Table 6.4: Variation of Saccadic Ratio for Different State of Alertness

| Parameter | State of Alertness | |
|---|---|---|
| | High | Low |
| SR ($rad/sec^2$) | $20.30 \pm 6.295$ | $11.02 \pm 4.445$ |
| Subjective Score (0-10) | $7 \pm 1.821$ | $2 \pm 1.120$ |

Table 6.4 indicate that the measured SR for the subjects and the alertness scores obtained from the responses to the questionnaire are highly correlated. However, an extended experiment with larger population can be carried out to establish it further.

## 6.4 Conclusion

A robust algorithm based on DFF and LFF along with EKF based algorithm has been proposed to estimate SR for two states of alertness. The eye saccade model and the estimation method have been developed and analyzed for simulated and video data of saccadic eye movements. The overall experimental results for determination of alertness from saccadic ratio has been compared with questionnaire based assessment of alertness explained in Chapter 2. The high correlation between these two measurements indicates that the proposed algorithm can be applied on data set with larger population in future.

# Chapter 7

# Conclusion and Future Scope

In this thesis image based algorithms for determination of alertness level in human vehicle drivers have been developed. Two ocular indices such as PERCLOS and SR have been used to figure out the temporal characteristics of the eyelid and pupil (iris) motion under different level of alertness respectively. A correlation filter based technique through DCT domain synthesis is proposed to measure the eyelid motion. The pupil motion is detected applying a newly proposed feature called Form Factor. The image database are generated using a CCD video camera sensitive to visual light during daylight. During night an NIR sensitive CCD camera with passive NIR illuminator has been used to capture the faces. The proposed methods are evaluated on various internal and external database and the results are compared with the existing state-of-the-art in respective field. The degree of alertness level can be determined by SR when the PERCLOS value is lower than a threshold while at reduced state of alertness it is to be decided by PERCLOS only.

In the first section of this chapter an overview of chapter-wise contributions of the thesis and the conclusions are summarized. Topics for future research are pointed out in the second section of this chapter.

## 7.1 Conclusion

**Experiment Design, Database Generation and Collection**  Literature survey indicates that image sequence database for both day and night driving conditions are unavailable. Therefore, it was necessary to generate database that includes all the challenges in driving scenario. Three experiments have been conducted to generate image sequence database. This database considers all the challenges that exist in driving situation such as variation in illumination level, free movement of operator, and variation of alertness level by incremental sleep deprivation etc. Experiment I was a preliminary experiment and provides only daylight images. Experiment II has yielded an extensive facial image data set with bio-marker indices as the reference for alertness level in human subject. This database is suitable to measure slow eyelid closure using PERCLOS with the image sequence being captured with 30 fps. The measured index has been compared with the bio-indices for validation. On the other hand a database recorded with high frame rate ($60 - 1000$ fps) is generated for measuring the SR during Experiment III.

The video database contains both color and Near Infra-Red (NIR) monochrome images. Any image processing algorithm that requires images with varying illumination can be applied on these image data for examining their performance. Further, an attempt can be made to measure human alertness based on other facial features and to validate these measures. The overall database (that includes video data, blood samples, response to psychometric questionnaire etc.) can be used for multi-modal analysis of human fatigue or alertness.

**Detection of Face and Eye** In this chapter, detection accuracy and computational complexity of several object detection techniques have been investigated. Statistical analysis has been carried out to choose the optimally accurate face and eye detector. Mathematical formulation of bound on execution time and memory consumption by these algorithms are developed and compared with experimental outcome. Experimental results indicate that Haar-like feature based method with modification can effectively be used to detect face and subsequently eye compared to others. However, speeding up the execution time still remains a research issue.

**PERCLOS:Eyelid Motion Measurement** The PERCLOS can be thought of as the temporal variation of eyelid position over eye. Detection of such variation has been formulated as classification eye images into open, closed and partially closed eye. This facilitates to develop a new correlation filters based classification technique for the purpose. The synthesis and testing of such correlation filters has been proposed to be carried out through type-II DCT instead of DFT. A theoretical and statistical outcome shows the effectiveness of the proposed method over DFT based one. The algorithm is applied to both on gray scale and NIR monochrome images that shows the usability of the proposed method in both daylight and night driving condition.

A real time on-board non-intrusive driver's drowsiness monitoring system comprising of an innovative combination of gray scale CCD camera for imaging during daylight and an NIR ($700 - 1200$ nm) sensitive monochrome CCD camera combined with compact passive NIR illuminator for imaging during night has been developed.

**Form Factor: A New Image Feature for Pupil Motion Detection** A new feature called Form Factor has been introduced in this chapter. Directional and local FF has been defined and analyzed for its range. Recovery of FF under zero-mean i.i.d. additive noise has been formulated. This indicates that blind estimation of SNR or STVR can be used for the purpose. The effectiveness of blind estimation of STVR over that of SNR by a few methods is justified with mathematical formulation and experimental validation.

The FF in horizontal and vertical direction is proposed to find the center of pupil (iris) in partially occluded and fully open eye. The results are compared with that obtained from popular Circular Hough Transform based technique. A new method based on radial FF has been devised to measure the pupil diameter which may provide the state of the operator's level of alertness.

The local FF based method has been proposed to develop an edge detection and subsequent eye corner detection from the edge map. The proposed method and the popular Canny operator have been applied on standard data set and the results are compared. The experimental output shows the advantage of the proposed method over that given by Canny edge operator.

Two applications of the proposed feature have been demonstrated. More investigations on it may extend its utility in other context.

**Detection and Interpretation of Saccadic Eye Movement from Image Sequence Analysis** A simple eye saccadic model has been proposed that can be used to estimate the saccadic velocity from eye image sequence. An EKF based estimation method using the proposed model has been applied on the positional information of pupil center relative to its corners. This relative position obviates the need of 3-D object geometric information.

The estimated saccadic ratio is used to indicate the level of alertness and compared it with the level indicated by questionnaire based subjective analysis.

## 7.2 Future Scope

The results of this thesis show important directions for further research in the following cases:

- Investigation on fast face and eye detection algorithm
- Exploration of Form Factor feature in other context of image processing
- Development of extended database for SR estimation with larger number of alertness level
- Implementation of combined PERCLOS and SR estimation algorithm for real-time embedded platform.

# Appendix A

# Derivation of the expressions (5.5) and (5.6)

The equation (5.5) is derived for signal (set of pixel values) contaminated with zero-mean additive i.i.d. noise. By definition the Form Factor of an observed noisy set of pixels is as follows-

$$F_g = \sqrt{1 + \frac{\sigma_g^2}{\mu_g^2}} \qquad (A.1)$$

Where, $\mu_g$ is the mean of summation of true signal and the noise and $\sigma_g$ is the standard deviation of the observed signal.

The overall mean and s.d. can be expressed by those of true signal and the noise. The relationships are shown in the following equations.

$$\mu_g = \mu_s + \mu_n \qquad (A.2)$$

Where, $\mu_s$ and $\mu_n$ are the averages of true signal and the noise respectively.

$$\sigma_g^2 = \sigma_s^2 + \sigma_n^2 + 2\rho_{sn}\sigma_s\sigma_n \qquad (A.3)$$

Where, $\sigma_s$ and $\sigma_n$ are the s.d. of true signal and the noise respectively. $\rho_{sn}$ is the correlation coefficient of the true signal and noise.

Now, for zero-mean additive i.i.d. signal $\mu_n = 0$ and $\rho_{sn} = 0$. Therefore, replacing

$\mu_g$ and $\sigma_g$ from (A.2) and (A.3) into (A.1) the square of the FF can be expressed as follows-

$$F_g^2 = 1 + \frac{\sigma_s^2 + \sigma_n^2}{\mu_s^2} \tag{A.4}$$

Rearranging (A.4) the FF for true signal can be related with observed signal. So, we get-

$$\begin{aligned} F_g^2 &= 1 + \frac{\sigma_s^2}{\mu_s^2}\left(1 + \frac{\sigma_n^2}{\sigma_s^2}\right) \\ &= 1 + (F_s^2 - 1)(1 + SNR^{-1}) \end{aligned} \tag{A.5}$$

Where, $F_s = \sqrt{1 + \frac{\sigma_s^2}{\mu_s^2}}$ is FF of true signal and $\frac{\sigma_n^2}{\sigma_s^2}$ is the Signal-to-Noise Ratio (SNR).

Therefore, from (A.5) we get the relation in (5.5) that recovers FF of true signal from the observed FF and the SNR i.e.

$$F_s^2 = \frac{F_g^2 + SNR^{-1}}{1 + SNR^{-1}} \tag{A.6}$$

Again the equation (A.4) can also be rearranged in another way as follows-

$$\begin{aligned} F_g^2 &= 1 + \frac{\sigma_s^2}{\mu_s^2} + \frac{\sigma_n^2}{\mu_s^2} \\ &= 1 + F_s^2 + \frac{\sigma_g^2 - \sigma_s^2}{\mu_s^2} \\ &= 1 + F_s^2 + \frac{\sigma_g^2}{\mu_g^2}\left(1 - \frac{\sigma_s^2}{\sigma_g^2}\right) \\ &= 1 + F_s^2 + \left(F_g^2 - 1\right)(1 - STVR) \end{aligned} \tag{A.7}$$

Where, $\frac{\sigma_s^2}{\sigma_g^2}$ is the Signal-to-Total Variance Ratio (STVR).

Therefore, from (A.7) we get the relation in (5.6) that recovers FF of true signal from the observed FF and the STVR i.e.

$$F_s^2 = STVR(F_g^2 - 1) + 1 \tag{A.8}$$

# References

[1] R. Grace, V. E. Byrne, D. M. Bierman, J. M. Legrand, D. Gricourt, B. K. Davis, J. J. Staszewski, and B. Carnahan, "A drowsy driver detection system for heavy vehicles," in *Proc. AIAA/IEEE/SAE th DASC Digital Avionics Systems Conf.*, vol. 2, Oct.-Nov. 1998, pp. I36/1–I36/8.

[2] S. K. L. Lal and A. Craig, "A critical review of the psychophysiology of driver's fatigue," *Biological Psychology*, vol. 55, no. 3, pp. 173–194, Feb. 2001.

[3] S. K. L. Lal, "The physiology of driver fatigue/drowsiness: Electroencephalography, electro-oculogram, electrocardiogram and psychological effects," Ph.D. dissertation, Dept. of Health Science, Univ. of Technology, Sydney, 2001.

[4] A. W. MacLean, D. R. T. Davies, and K. Thiele, "The hazards and prevention of driving while sleepy," *Sleep Medicine Review*, vol. 7, no. 6, pp. 507–521, Jul. 2003.

[5] J. Jagnoor, "Road traffic injury prevention: A public health challenge," *Indian Journal of Community Medicine*, vol. 31, no. 3, pp. 129–131, Jul. 2006.

[6] N. B. Powell and J. K. M. Chau, "Sleepy driving," *Sleep Medicine Clinics, Elsevier*, vol. 6, no. 1, pp. 117–124, Mar. 2011.

[7] D. F. Dinges, M. M. Mallis, G. Maislin, and J. W. Powell, "Evaluation of techniques for ocular measurement as an index of fatigue and as the basis for alertness management," Tech. Rep. DOT HS 808 762, Apr. 1998.

[8] M. W. Johns, "A new method for measuring daytime sleepiness: The epworth sleepiness scale," *Sleep*, vol. 14, no. 6, pp. 540–545, 1991.

[9] T. Chalder, G. Berelowitz, T. Pawlikowska, L. Watts, S. Wessely, D. Wright, and E. Wallace, "Development of a fatiguescale," *Journal of Psychosomatic Research*, vol. 37, no. 2, pp. 147–153, Feb. 1993.

[10] N. Chakraborty, "Development of driver evaluation system in india," in *Int. Seminar on Highway Safety Management and Devices*, 1998.

[11] Y. Y. Pang, X. P. Li, K. Q. Shen, H. Zheng, W. Zhou, and E. P. V. Wilder-Smith, "An auditory vigilance task for mental fatigue detection," in $27^{th}$ *IEEE Int. Conf. on Engineering in Medicine and Biology, Shanghai, China*, Sep. 2005, pp. 5284–5286.

[12] D. R. Davies, "Skin conductance, alpha activity and vigilance," *Am. J. Psychol.*, vol. 78, no. 2, pp. 304–306, Jun. 1965.

[13] C. W. Erwin, M. R. Volow, and B. Gray, "Psychophysiologic indices of drowsiness," in *International Automotive Engg Congress of SAE,Detroit*, Jan. 1973.

[14] E. Grandjean, "Fatigue in industry," *Br. J. Ind. Med.*, vol. 36, no. 3, pp. 175–186, Aug. 1979.

[15] T. Ackerstedt, L. Torsvall, and M. Gillberg, "Sleepiness and shiftwork: field studies," *Sleep: Journal of Sleep Research & Sleep Medicine*, vol. 5, no. 2, pp. 95–106, 1982.

[16] T. C. Chieh, M. M. Marzuki, H. Aini, H. S. Farshad, and M. B. Yeop, "Development of vehicle driver drowsiness detection system using electrooculogram (eog)," in *Proc. 1st Int. Conf. Computers, Communications, & Signal Processing with Special Track Biomedical Engineering CCSP 2005*, Nov. 2005, pp. 165–168.

[17] E. Magosso, M. Ursino, F. Provini, and P. Montagna, "Wavelet analysis of electroencephalographic and electro-oculographic changes during the sleep onset period," in *Proc. 29th Annual Int. Conf. of the IEEE Engineering in Medicine and Biology Society EMBS 2007*, Aug. 2007, pp. 4006–4010.

[18] A. Picot, A. Caplier, and S. Charbonnier, "Comparison between eog and hhigh frame rate camera for drowsiness detection," in *Proc. Workshop Applications of Computer Vision (WACV)*, Dec. 2009, pp. 1–6.

[19] E. Park and S. G. Meek, "Fatigue compensation of electromyographic signal for prosthetic control and force estimation," *IEEE Trans. on Biomedical Engineering*, vol. 40, no. 10, pp. 1019–1023, Oct. 1993.

[20] H. Oka, "Estimation of muscle fatigue by using emg and muscle stiffness," in *Proc. 18th Annual Conference of the IEEE Engineering in Medicine and Biology Society*, vol. 4, Oct.-Nov. 1996, pp. 1449–1450.

[21] M. Knaflitz and F. Molinari, "Assessment of muscle fatigue during biking," *IEEE Trans. Neural System and Rehabilitation Engineering*, vol. 11, no. 1, pp. 17–23, Mar. 2003.

[22] R. Gwin, B. A. Gwin, and C. R. Hyde, "Driver alarm," U.S.A. Patent 5,585,785, Dec. 17, 1996.

[23] L. Leavitt, "Sleep-detecting driving gloves," U.S.A. Patent 6,016,103, Jan. 18, 2000.

[24] E. Rogado, J. L. Garcia, R. Barea, L. M. Bergasa, and E. Lopez, "Driver fatigue detection system," in *Proc. IEEE Int. Conf. Robotics and Biomimetics ROBIO 2008*, 2009, pp. 1105–1110.

[25] G. R. Harrison and J. A. Horne, "Sleep deprivation affects speech," *Sleep*, vol. 20, no. 10, pp. 871–877, 1997.

[26] J. Krajewski and B. Kroger, "Using prosodic and spectral characteristics for sleepiness

detection," in *Proc. Inter Speech*, 2007.

[27] J. E. Gangwisch, S. B. Heymsfield, B. Boden-Albala, R. M. Buijs, F. Kreier, T. G. Pickering, A. G. Rundle, G. K. Zammit, and D. Malaspina, "Sleep duration as a risk factor for diabetes incidence in a large us sample," *Sleep*, vol. 30, no. 12, pp. 1667–1673, Dec. 2007.

[28] M. Matousek and A. Petersn, "A method for assessing alertness fluctuations from eeg spectra," *Electroencephalography and Clinical Neurophysiology, Elsevier*, vol. 55, no. 1, pp. 108–113, Jan. 1983.

[29] M. M. Mallis and M. B. Russo, "Ocular measures of fatigue and extended wakefulness," *J Aviation, Space, Environmental Medicine*, vol. 76, 2005.

[30] W. Ronben, G. Lie, T. Bingliang, and J. Lisbeng, "Monitoring mouth movement for driver fatigue or distraction with one camera," in *Proc. IEEE $7^{th}$ Int. Conf. Intelligent Transportation Syst., Washington, D.C., USA*, Oct. 2004, pp. 314–319.

[31] R. Bittner, K. Hna, L. Pouek, P. Smrcka, and P. Vysok, "Detecting of fatigue states of a car driver," in *Proc. Medical Data Analysis: First Int. Symposium, ISMDA 2000, Frankfurt, Germany*, Sep. 2000, pp. 260–273.

[32] K. Fukuda, J. A. Stern, T. B. Brown, and M. B. Russo, "Cognition, blinks, eye-movements, and pupilary, movements during performance of a running memory task," *Aviation, Space, and Environmental Medicine*, vol. 76, no. 7, pp. C75–C85, Jul. 2005.

[33] J. A. Stern and T. Ranney, "Ocular based measures of driver alertness," in *Ocular Measures of Driver Alertness: Technical Conference Proceedings, Herndon, Virginia*, no. FHW A-MC-99-136, Apr. 1999, pp. 4–23.

[34] E. J. Sirevaag and J. A. Stern, *Ocular Measures of Fatigue and Cognitive Factors*. Lawrence Erlbaum Associates, NJ, 2000, pp. 269–287.

[35] Q. Wang, J. Yang, M. Ren, and Y. Zheng, "Driver fatigue detection: A survey," in *Proc. $6^{th}$ World Congress on Intell. Control and Automation, WCICA, 2006*, vol. 2, Jun. 2006, pp. 8587–8591.

[36] R. J. Caroll, "Ocular measures of driver alertness: Technical conference proceedings," in *FHWA Technical Report, Washington DC, Federal Highway Administration Office of Motor Carrier and Highway Safety*, no. FHWA-MC-99-136, Apr. 1999.

[37] M. Al-Abdulmunem and S. T. Briggs, "Spontaneous blink rate of a normal population sample," *International Contact Lens Clinic, Elsevier*, vol. 26, no. 2, pp. 29–32, Mar. 1999.

[38] L. Hartley, T. Horberry, N. Mabbott, and G. P. Krueger, "Review of fatigue detection and prediction technologies," Tech. Rep., Sep. 2000.

[39] R. Grace, "Drowsy driver monitor and warning system," in *in Int. Driving Symp. on Human Factors in Driver Assessment, Training and Vehicle Design*, 2001.

[40] P. Smith, M. Shah, and N. D. V. Lobo, "Determining driver visual attention with one camera," *IEEE Trans. Intelligent Transportation System*, vol. 4, no. 4, pp. 205–218, Dec. 2003.

[41] Q. Ji, Z. Zhu, and P. Lan, "Real-time nonobtrusive monitoring and prediction of driver fatigue," *IEEE Trans. Vehicular Technology*, vol. 53, no. 4, pp. 1052–1068, Jul. 2004.

[42] L. M. Bergasa, J. Nuevo, M. A. Sotelo, R. Barea, and M. E. Lopez, "Real-time system for monitoring driver vigilance," *IEEE Trans. Intelligent Transportation System*, vol. 7, no. 1, pp. 63–77, Mar. 2006.

[43] L. Barr, S. Popkin, and H. Howarth, "An evaluation of emerging driver fatigue detection measures and technologies," Tech. Rep. FMCSA-RRR-09-005, Jun. 2009.

[44] M. M. Mallis and D. F. Dinges, *Monitoring Alertness by Eyelid Closure*. CRC Press, 2005, pp. 25/1–25/6.

[45] A. Ueno, Y. Ota, M. Takase, and H. Minamitani, "Relationship between vigilance levels and characteristics of saccadic eye movement," in *Proc. IEEE 17th Annual Conf. Engineering in Medicine and Biology Society*, vol. 2, 1995, pp. 1445–1446.

[46] A. Ueno, T. Tateyama, M. Takase, and H. Minamitani, "Dynamics of saccadic eye movement depending on diurnal variation in human alertness," *IEICE Transactions on Information and Systems*, vol. J83-D-2, no. 4, pp. 1172–1179, 2000.

[47] A. Ueno, Y. Ota, M. Takase, and H. Minamitani, "Parametric analysis of saccadic eye movement depending on vigilance states," in *Proc. 18th Annual Int. Conf. of the IEEE Bridging Disciplines for Biomedicine Engineering in Medicine and Biology Society*, vol. 5, 1996, pp. 1782–1783.

[48] A. Ueno and Y. Uchikawa, "Relation between human alertness, velocity wave profile of saccade, and performance of visual activities," in *Proc. 26th Annual Int. Conf. of the IEEE Engineering in Medicine and Biology Society IEMBS '04*, vol. 1, 2004, pp. 933–935.

[49] A. Ueno, S. Tei, T. Nonomura, and Y. Inoue, "An analysis of saccadic eye movements and facial images for assessing vigilance levels during simulated driving," in *Engineering Psychology and Cognitive Ergonomics*, ser. Lecture Notes in Computer Science, D. Harris, Ed. Springer Berlin / Heidelberg, 2009, vol. 5639, pp. 451–460.

[50] T. Nonomura, Y. Inoue, and A. Ueno, "Investigation of intraindividual difference in vigilance index of saccade based on rated facial sleepiness," in *World Congress on Medical Physics and Biomedical Engineering, September 7 - 12, 2009, Munich, Germany*, ser. IFMBE Proceedings, R. Magjarevic, J. H. Nagel, O. Dössel, and W. C. Schlegel, Eds. Springer Berlin Heidelberg, 2009, vol. 25/12, pp. 356–359. [Online]. Available: http://dx.doi.org/10.1007/978-3-642-03893-8_102

[51] A. T. Bahill, M. R. Clark, and L. Stark, "The main sequence, a tool for studying human eye movements," *Math. Bio-sci.*, vol. 24, pp. 191–197, 1975.

[52] C. M. Harris, J. Wallman, and C. A. Scudder, "Fourier analysis of saccades in monkeys and humans," *J. Neurophysiol.*, vol. 63, no. 4, pp. 877–886, 1990.

[53] R. Wierts, M. J. A. Janssen, and H. Kingma, "Measuring saccade peak velocity using a low frequency sampling rate of 50hz," *IEEE trans. on Biomed. Engg.*, vol. 55, 2008.

[54] A. Ueno, Y. Otani, and Y. Uchikawa, "A noncontact measurement of saccadic eye movement with two high-speed cameras," in *Proc. 28th Annual Int. Conf. of the IEEE Engineering in Medicine and Biology Society EMBS '06*, Sep. 2006, pp. 5583–5586.

[55] T. Hayami, K. Matsunaga, K. Shidoji, and Y. Matsuki, "Detecting drowsiness while driving by measuring eye movement - a pilot study," in *IEEE 5th Int. Conf. on Intelligent Transportation Systems, Singapore*, 2002.

[56] K. Kozak, R. Curry, J. Greenberg, B. Artz, M. Blommer, and L. Cathey, "Leading indicators of drowsiness in simulated driving," in *Proc. The Human Factors and Ergonomics Society Annual Meeting*, vol. 49, no. 22, Sep. 2005, pp. 1917–1921.

[57] R. Martin, "Noise power spectral density estimation based on optimal smoothing and minimum statistics," *IEEE Trans. Acoust., Speech, Signal Processing*, vol. 9, no. 5, pp. 504–512, Jul. 2001.

[58] M. B. Stegmann, B. K. Ersbøll, and R. Larsen, "Fame - a flexible appearance modelling environment," *IEEE Trans. on Medical Imaging*, vol. 22, no. 10, pp. 1319–1331, 2003.

[59] O. Jesorsky, K. J. Kirchberg, and R. W. Frischholz, "Robust face detection using the hausdorff distance," in *Audio- and Video-Based Biometric Person Authentication*, ser. Lecture notes in Comput. Sci. Springer Berlin/Heidelberg, Jun. 2001, pp. 90–95.

[60] S. Z. Li, R. Chu, S. Liao, and L. Zhang, "Illumination invariant face recognition using near-infrared images," *IEEE Trans. on Pattern Analysis and Machine Intelligence*, vol. 29, no. 4, pp. 627–639, 2007.

[61] H. Proenca and L. A. Alexandre, "Ubiris: A noisy iris image database," in *Proc. of ICIAP 2005 - Intern. Confer. on Image Analysis and Processing*, vol. 1, 2005, pp. 970–977.

[62] T. Thongkamwitoon, S. Aramvith, and T. H. Chalidabhongse, "An adaptive realtime background subtraction and moving shadows detection," in *Proc. IEEE Int. Conf. Multimedia and Expo ICME '04*, vol. 2, 2004, pp. 1459–1462.

[63] D. Hong and W. Woo, "A background subtraction for a vision-based user interface," in *Proc. Joint Conf. of the Fourth Int. Conf. and the Fourth Pacific Rim Conf Information, Communications and Signal Processing Multimedia*, vol. 1, 2003, pp. 263–267.

[64] M. Heikkilä and M. Pietikainen, "A texture-based method for modellig the back-

ground and detecting moving objects," *IEEE Trans. Pattern Analysis and Machine Intelligence*, vol. 28, no. 4, pp. 657–662, Apr. 2006.

[65] A. Elgammal, R. Duraiswami, D. Harwood, and L. S. Davis, "Background and foreground modeling using nonparametric kernel density estimation for visual surveillance," in *Proc. IEEE*, vol. 90, no. 7, Jul. 2002, pp. 1151–1163.

[66] K. Kim, T. H. Chalidabhongse, D. Harwood, and L. Davis, "Real-time foreground-background segmentation using codebook model," *Real-Time Imaging, Elsivier*, vol. 11, no. 3, pp. 172–185, Jun. 2005.

[67] P. Viola and M. Jones, "Rapid object detection using a boosted cascade of simple features," in *Proc. Int. conf. of IEEE Computer Vision and Pattern Recognition*, vol. 1, Dec. 2001, pp. 511–518.

[68] R. Lienhart and J. Maydt, "An extended set of haar-like features for rapid object detection," in *Proc. IEEE Conf. Image Processing*, vol. 1, Sep. 2002, pp. I–900–I–903.

[69] W. Hu, T. Tan, L. Wang, and S. Maybank, "A survey on visual surveillance of object motion and behaviors," *IEEE Trans. Syst., Man, and Cybern., Part C: Applications and Reviews*, vol. 34, no. 3, pp. 334–352, Aug. 2004.

[70] Q. Z. Wu and B. S. Jeng, "Background subtraction based on logarithimic intensities," *Pattern Recognition Letts.*, vol. 23, no. 13, pp. 1529–1536, Nov. 2002.

[71] P. Sinha, "Object recognition via image invariants: A case study," *Int. J. Investigative Ophthalmology and Visual Science*, vol. 35, pp. 1735–1740, May. 1994.

[72] P. Sinha, B. Balas, Y. Ostrovsky, and R. Russell, "Face recognition by humans: Nineteen results all computer vision researchers should know about," in *Proc. of the IEEE Biometric: Algorithms and Applications*, vol. 94, no. 11, Nov. 2006, pp. 1948–1962.

[73] M. Oren, C. Papageorgiou, P. Sinha, E. Osuna, and T. Poggio, "Pedestrian detection using wavelet templates," in *Proc. Computer Vision and Pattern Recognition (CVPR '97)*, Jun. 1997, pp. 193–199.

[74] H. A. Rowley, S. Baluja, and T. Kanade, "Neural network-based face detection," *IEEE Patt. Anal. Mach. Intell.*, vol. 20, no. 1, pp. 23–38, Jan. 1998.

[75] C. H. Messom and A. Barczak, "Fast and efficient rotated haar-like features using rotated integral images," in *Proc. Australian Conf. on Robotics and Automation (ACRA2006)*, 2006.

[76] N. Seo, "Tutorial: Opencv haartraining (rapid object detection with a cascade of boosted classifiers based on haar-like features)," accessed on $15^{th}$ August 2011. [Online]. Available: http://note.sonots.com/SciSoftware/haartraining.html

[77] T. Fawcett, "Roc graphs: Notes and practical considerations for researchers," Tech. Rep. HPL-2003-4, Jan. 2003.

[78] T. H. Cormen, C. E. Leiserson, R. L. Rivest, and C. Stein, *Introduction to Algorithms*,

2nd ed. The MIT press, Cambridge, Massachusetts, 2001.

[79] M. M. Mallis, G. Maislin, N. Konowal, V. Bryne, D. Bierman, R. Davis, R. Grace, and D. F. Dinges, "Biobehavioral responses to drowsy driving alarms and alerting stimuli," Tech. Rep. DOT HS 809 202, 2000.

[80] P. Viola and M. Jones, "Robust real-time object detection," *Int. J. Computer Vision*, vol. 57, no. 2, pp. 137–154, Jul. 2004.

[81] M. Turk and A. Pentland, "Eigenfaces for recognition," *Journal of Cognitive Neuroscience*, vol. 3, no. 1, pp. 71–86, 1991.

[82] B. V. K. V. Kumar, A. Mahalanobis, and R. Juday, *Correlation Pattern Recognition*, 1st ed. Cambridge, U. K.: Cambridge Univ. Press, 2005.

[83] V. Britanak, P. Yip, and K. R. Rao, *Discrete Cosine and Sine Transforms: General Properties, Fast Algorithms and Integer Approximations*, 1st ed. Academic Press, 2006.

[84] J. Yeo, "Driver's drowsiness detection method of drowsy driving warning system," U.S.A. Patent 6,243,015, Jun. 5, 2001.

[85] J. A. Home and L. A. Reyner, "Sleep related vehicle accidents," *British Medical Journal*, 1995.

[86] W. H. Chen, C. Smith, and S. C. Fralick, "A fast computational algorithm for the discrete cosine transform," *IEEE Trans. Commun.*, vol. COM 25, no. 9, pp. 1004–1008, Sep. 1977.

[87] M. E. Rizkalla, M. Ei-Sharkawy, P. Salama, and B. Dukel, "Implementation of floating point fast discrete cosine transform," in *Proc. IEEE International Symposium on Circuits and Systems*, vol. 2, Aug. 2002, pp. 17–20.

[88] H. Zhou and T. Chao, "Mach filter synthesizing for detecting targets in cluttered environment for gray scale optical correlator," in *Proc. SPIE, Optical Pattern Recognition*, vol. 3715, 1999, pp. 394–398.

[89] Y.-H. Chan and W. Siu, "A new convolution structure for the realisation of the discrete cosine transform," in *Proc. IEEE International Symposium on Circuits and Systems*, vol. 3, May 1990, pp. 2373–2376.

[90] W. H. Chen and S. C. Fralick, "Image enhancement using cosine transform filtering," in *Image Science Math. Symposium, Monterey, CA*, 1976.

[91] B. Chitprasert and K. R. Rao, "Discrete cosine transform filtering," in *Proc. IEEE Int. Conf. on Acoustics, Speech, and Signal Processing*, vol. 3, Apr. 1990, pp. 1281–1284.

[92] C. K. Ng, M. Savvides, and P. K. Khosla, "Real time face verification system on a cell phone using advanced correlation filters," in *IEEE Workshop on Automatic Identification Advanced Technologies*, 2005.

[93] R. M. Gray, *Entropy and Information Theory*, 2nd ed. New York: Springer Verlag,

Dec. 2010.

[94] A. K. Jain, *Fundamentals of Digital Image Processing*. Prentice Hall of India Private Limited.

[95] P. Rasmussen, N. H. Secher, and N. T. Petersen, "Understanding central fatigue: where to go?" *Exp. Physiol.*, vol. 92, pp. 369–370, 2007.

[96] K. Luthra, *Basic Concept of Clinical Biochemistry*. National Science Digital Library at NISCAIR, India, 2008. [Online]. Available: http://nsdl.niscair.res.in/handle/123456789/691

[97] S. A. Marcus, *The Hungry Brain: The Nutrition/Cognition Connection*. Corwin Press, 2007.

[98] R. D. Nowak, "Wavelet-based rician noise removal for magnetic resonance imaging," *IEEE Trans. on Image Processing*, vol. 8, no. 10, pp. 1408–1419, Oct. 1999.

[99] S. Aja-Fernández and G. Vegas-Sánchez-Ferrero, "Automatic noise estimation in image using local statistics. additive and multiplicative cases," *Image and Vision Computing, Elsevier*, vol. 27, no. 6, pp. 756–770, May 2009.

[100] S. Dan and G. Lindong, "On the blind SNR estimation for IF signals," in *Proc. 1st Int. Conf. on Innovative, Computing, Information and Control*, Aug.-Sep. 2006, pp. 374–378.

[101] K. S. Sim, M. A. Lai, C. P. Tso, and C. C. Teo, "Single image signal-to-noise ratio estimation for magnetic resonance images," *Int. J. Med. Syst.*, vol. 35, no. 1, pp. 39–48, Jul. 2009.

[102] Z. Wang and A. C. Bovik, "A universal image quality index," *IEEE Trans. Signal Process. Letts.*, vol. 9, no. 3, pp. 81–84, Mar. 2002.

[103] Z. Wang, A. C. Bovik, H. R. Sheikh, and E. P. Simoncelli, "Image quality assessment: From error visibility to structural similarity," *IEEE trans. Image Processing*, vol. 13, no. 4, pp. 600–612, Apr. 2004.

[104] M. A. Saad, A. C. Bovik, and C. Charrier, "A dct statistics-based blind image quality index," *IEEE Trans. Signal Process. Letts.*, vol. 17, no. 6, pp. 583–586, Jun. 2010.

[105] H. Proenca, S. Filipe, R. Santos, J. Oliveira, and L. Alexandre, "The UBIRIS.v2: A database of visible wavelength images captured on-the-move and at-a-distance," *IEEE Trans. Pattern Analysis and Machine Intelligence*, vol. 32, no. 8, pp. 1529–1535, Aug. 2010.

[106] C. Liu, W. T. Freeman, R. Szeliski, and S. B. Kang, "Noise estimation from a single image," in *Proc. IEEE Computer Society Conf. Computer Vision and Pattern Recognition*, vol. 1, Jun. 2006, pp. 901–908.

[107] C. Liu, R. Szeliski, S. B. Kang, C. L. Zitnick, and W. T. Freeman, "Automatic estimation and removal of noise from a single image," *IEEE Trans. Pattern Analysis*

*and Machine Intelligence*, vol. 30, no. 2, pp. 299–314, Feb. 2008.

[108] M. Sonka, V. Halvac, and R. Boyle, *Image Processing, Analysis, and Machine Vision*, 2nd ed.   Singapore: Thomson Learning Asia Pvt. Ltd., 1998.

[109] R. C. Gonzalez and R. E. Woods, *Digital Image Processing*, 3rd ed.   Singapore: Pearson Education Singapore Pvt. Ltd., 2008, vol. 2.

[110] F. A. Pellegrino, W. Vanzella, and V. Torre, "Edge detection revisited," *IEEE Trans. Syst., Man, and Cybern.-Part B: Cybern.*, vol. 34, no. 3, pp. 1500–1518, Jun. 2004.

[111] R. R. Rakesh, P. Chaudhuri, and C. A. Murthy, "Threshonding in edge detection: A statistical approach," *IEEE Trans. Image Process.*, vol. 13, no. 7, pp. 927–936, Jul. 2004.

[112] M. C. Morrone and R. A. Owens, "Feature detection from local energy," *Pattern Recognition Letts.*, vol. 6, no. 5, pp. 303–313, Dec. 1987.

[113] M. C. Morrone and D. C. Burr, "Feature detection in human vision: A phase dependent energy model," in *Proc. Roy. Soc. of London, Series B, Biological Sci.*, vol. 235, no. 1280, Dec. 1988, pp. 221–245.

[114] J. Canny, "A computational approach to edge detection," *IEEE Trans. Pattern Anal. and Mach. Intell.*, vol. PAMI 8, no. 6, pp. 679–698, Nov. 1987.

[115] S. Sarkar and K. L. Boyer, "Optimal infinite impulse response zero crossing based edge detectors," *CVGIP: Image Understanding*, vol. 54, no. 2, pp. 224–243, Sep. 1991.

[116] D. Marr and E. Hildreth, "Theory of edge detection," in *Proc. Roy. Soc. of London, Series B, Biological Sci.*, vol. 207, no. 1167, Feb. 1980, pp. 187–217.

[117] I. Abraham, R. Abraham, A. Desolneux, and S. T. L. Thiao, "Significant edges in the case of non-stationary gaussian noise," *Pattern Recognition, Elsevier*, vol. 40, no. 11, pp. 3277–3291, Nov. 2007.

[118] P. Kovesi, "Invariant measures of image features from phase information," Ph.D. dissertation, Dept. Psychology, Western Australia Univ., 1996.

[119] ——, "Edges are not just steps," in *Proc. 5th Asian Conf. on Comput. Vision, Melbourne, Australia*, Jan. 2002, pp. 822–827.

[120] ——, "Phase congruency detects corners and edges," in *Proc. 7th Int. Conf. Digital Image Computing: Techniques and Applicat., Sydney, Australia*, Dec. 2003, pp. 309–318.

[121] S. M. Smith and J. M. Brady, "Susan-a new approach to low level image processing," *Int. J. Comput. Vision*, vol. 23, no. 1, pp. 45–78, 1997.

[122] I. N. Bankman and E. W. Rogala, "Corner detection for identification of man made objects in noisy aerial images," in *Proc. Int. Conf. Soc. of Photographic Instrumentation Engineers*, vol. 4726, 2002, pp. 304–309.

[123] S. Yi, D. Labate, G. R. Easley, and H. Krim, "A shearlet approach to edge analysis

and detection," *IEEE Trans. Image Process.*, vol. 18, no. 5, pp. 929–941, May. 2009.

[124] S. Konishi, A. L. Yuille, J. M. Coughlan, and S. C. Zhu, "Statistical edge detection: Learning and evaluating edge cues," *IEEE Trans. Pattern Anal. and Mach. Intell.*, vol. 25, no. 1, pp. 57–74, Jan. 2003.

[125] S. Konishi, J. M. Coughlan, and A. L. Yuille, "Statistical approach to multi scale edge detection," *Image Vision and Comput.*, vol. 21, no. 1, pp. 37–48, Jan. 2003.

[126] J. A. Baddeley, "An error metric for binary images," in *Proc. IEEE Workshop on Robust Comput. Vision*, 1992, pp. 59–78.

[127] D. Mukherjee and M. V. Ratnaparkhi, "On the functional relationship between entropy and variance with related application," *Commun. in Stat. - Theory and Methods*, vol. 15, no. 1, pp. 291–311, 1986.

[128] D. Martin, C. Fowlkes, D. Tal, and J. Malik, "A database of human segmented natural images and its application to evaluating segmentation algorithms and measuring ecological statistics," in *Proc. 8th Int. Conf. Comput. Vision*, vol. 2, Jul. 2001, pp. 416–423.

[129] C. L. Molina, H. Bustince, J. Fernandez, E. Barrenechea, P. Couto, and B. D. Baets, "A t norm based approach to edge detection," in *Bio-Inspired Systems: Computational and Ambient Intell.*, ser. Lecture notes in Comput. Sci. Springer Berlin / Heidelberg, 2009, vol. 5517, pp. 302–309.

[130] C. L. Molina, H. Bustince, J. Fernandez, P. Couto, and B. D. Baets, "A gravitational approach to edge detection based on triangular norms," *Pattern Recognition, Elsivier*, vol. 43, no. 11, pp. 3730–3741, Nov. 2010.

[131] S. Venkatesh and L. J. Kitchen, "Edge evaluation using necessary components," *Comput. Vision, Graph. and Image Process.*, vol. 54, no. 1, pp. 23–30, Jan. 1992.

[132] D. Demigny, F. G. Lorca, and L. Kessal, "Evaluation of edge detectors performances with a discrete expression of canny's criteria," in *Proc. Int. Conf. Image Process.*, vol. 2, Oct. 1995, pp. 169–172.

[133] W. K. Pratt, *Digital Image Processing*, 3rd ed. Singapore: John Willey & Sons Inc., 2003.

[134] K. Bowyer, C. Kranenburg, and S. Dougherty, "Edge detector evaluation using empirical roc curves," *Comput. Vision and Image Understanding, Elsevier*, vol. 84, no. 1, pp. 77–103, Oct. 2001.

[135] R. Medina-Carnicer, A. Carmona-Poyatoand, R. Munoz-Salinas, and F. Madrid-Cuevas, "Determining hysteresis thresholds for edge detection by combining the advantages and disadvantages of thresholding methods," *IEEE Trans. Image Process.*, vol. 19, no. 1, pp. 165–173, Jan. 2010.

[136] N. Fernández-Gracía, A. Carmona-Poyato, R. Medina-Carnicer, and F. Madrid-

Cuevas, "Automatic generation of consensus ground truth for the comparison of edge detection techniques," *Image and Vision Computing*, vol. 26, no. 4, pp. 496–511, Apr. 2008.

[137] T. L. Morrisa and J. C. Millerb, "Electrooculographic and performance indices of fatigue during simulated flight," *Biol. Psychol.*, vol. 42, no. 3, pp. 343–360, 1996.

[138] M. Russo, M. Thomas, D. Thorne, H. Sing, D. Redmond, L. Rowland, D. Johnson, S. Hall, J. Krichmar, and T. Balkin, "Oculomotor impairment during chronic partial sleep deprivation," *Clinical Neurophysiology*, vol. 114, no. 4, pp. 723–736, 2003.

[139] J. Huang and H. Wechsler, "Eye detection using optimal wavelet packets and radial basis functions (rbfs)," *Int. J. Pattern Recognition. Artificial Intell.*, vol. 13, pp. 1009–1025, 1999.

[140] S. Sirohey and A. Rosenfeld, "Eye detection in a face image using linear and nonlinear filters," *Pattern Recognition*, vol. 34, no. 7, pp. 1367–1391, 2001.

[141] T. Kawaguchi and M. Rizon, "Iris detection using intensity and edge information," *Pattern Recognition*, vol. 36, pp. 549–562, 2003.

[142] Z. Xingming and Z. Huangyuan, "An illumination independent eye detection algorithm," in *18th Int. Conf. on Pattern Recognition (ICPR)*, vol. 1, 2006, pp. 392–395.

[143] S. Zhao and R.-R. Grigat, "Robust eye detection under active infrared illumination," in *18th Int. Conf. on Pattern Recognition (ICPR)*, vol. 4, 2006, pp. 481–484.

[144] J. Y. Z. Zheng and L. Yang, "A robust method for eye features extraction on color image," *Pattern Recognition Lett.*, vol. 26, no. 14, pp. 2252–2261, 2005.

[145] M. L. T. D'Orazio and A. Distante, "Eye detection in face images for a driver vigilance system," in *IEEE Intelligent Vehicles Symposium*, Jun. 2004, pp. 95–98.

[146] W. M. K. W. M. Khairosfaizal and A. J. Nor'aini, "Eyes detection in facial images using circular hough transform," in *5th Int. Colloquium on Signal Processing and Its Applications (CSPA)*, Mar. 2009, pp. 238–242.

[147] S. Jainta, M. Vernet, Q. Yang, and Z. Kapoula, "The pupil reflects motor preparation for saccades even before the eye starts to move," *Front. Hum. Neurosci.*, vol. 5, pp. 97/1–97/10, Sep. 2011.

[148] A. T. Duchowski, *Eye Tracking Methodology: Theory and Practice*, 2nd ed. Springer-Verlag, New York, 2007.

[149] J. D. Enderle, *Models of Horizontal EyeMovements, Part I: Early Models of Saccades and Smooth Pursuit*, ser. Synthesis Lectures in Biomedical Engineering. Morgan & Claypool Publishers, 2010.

[150] T. Vilis, "Physiology of three-dimensional eye movements: Saccade and vergence," in *Three-Dimensional Kinamatics of Eye, Head and Limb Movements*, 1st ed., M. Fetter, T. Haslwanter, H. Misslisch, and D. Tweed, Eds. Harwood Academic Publisher,

Netherlands, 1997, pp. 59–72.

[151] M. S. Nixon and A. S. Aguado, *Feature Extraction & Image Processing*, 2nd ed. Oxford, U. K.: Academic Press, 2008.

[152] M. Fetter, T. Haslwanter, H. Misslisch, and D. Tweed, *Three-Dimensional Kinamatics of Eye, Head and Limb Movements*, 1st ed. Harwood Academic Publisher, Netherlands, 1997.

# Author's Publications

(a) **Patent:**
 1. S. Gupta, and A. Routray, Real-time On Board Driver Drowsiness Monitoring System, *filed to Patent Office, Govt. of India, Jun. 2010, Ref. No. 660/Kol/2010.*

(b) **Referred Journals:**
 1. S. Gupta, A. Routray, and A. Mukherjee, A New Method for Edge Extraction in Images using Local Form Factors, *Int. J. Comput. Application (IJCA)* vol. 21, no. 2, pp. 15-22, May 2011
 2. S. Gupta, A. Routray, and A. Mukherjee, No Reference Estimation of Image Quality from Single Frame, *revised and submitted to Signal, Image and Video Processing, Springer*, Dec. 2011
 3. V. Ramanathan, S. Gupta, and A. Routray, Robust Eye Pupil Detection and Tracking, *Int. J. Comput. Application (IJCA)* (accepted)

(c) **Conference Proceedings:**
 1. S. Gupta, A. Dasgupta, and A. Routray, Analysis of Training Parameters for Classifiers Based on Haar-like Features to Detect Human Faces, *in IEEE Int. Conf. on Image Information Processing, Shimla, India*, Nov. 2011.
 2. S. Gupta, R. Vignesh, and A. Routray, Form Factor: A New Technique for automated center of pupil detection, *in 1st Int. Conf. on Emerging Trends in Signal Processing and VLSI Design, Hyderabad, India*, Jun. 2010.
 3. S. Gupta, A. Routray, and A. Mukherjee, Correlation Filter Based Method for Measurement of Eyelid Movement, *in 2nd Int. Conf. on Multimedia and Content Based Information Retrieval, Bangalore, India*, Jul., 2010.
 4. S. Gupta, S. Kar, S. Gupta, and A. Routray, Fatigue in Human Drivers: A Study Using Ocular, Psychometric, Physiological Signals, *in IEEE symposium, Techsym, Kharagpur, India*, Apr. 2010.
 5. S. Dhar, T. Pradhan, S. Gupta, and A. Routray, Implementation of Real Time Visual Attention Monitoring Algorithm of Human Drivers on an Embedded Platform, *in IEEE symposium, Techsym, Kharagpur, India*, Apr. 2010. (Best Paper Award)
 6. A. N. Bagaria, S. Gupta, and A. Routray, Detection of Slow Eyelid Closure Using Eigen-eyes, *in $5^{th}$ Int. Conference on Industrial and Information System, Mangalore, India*, Jul-Aug. 2010.
 7. S. Gupta, A. Mukherjee, A. Routray, and S. Sharma, A Comparative Assessment of Background Subtraction Methods for Face Extraction under Variable Illumination, *in National Conference on Information Science and Security, Chennai, India*, Mar. 2010.
 8. S. Gupta, S. Kar, S. Naik, A. Routray, and A. Mukherjee, Multidimensional Approach for Detection of Human Fatigue, *in 1st Indo Japan Conference on Science and Technology of Facial Expression Analysis, CDAC, Kolkata, India*, Mar. 2009.

# Author's Biography

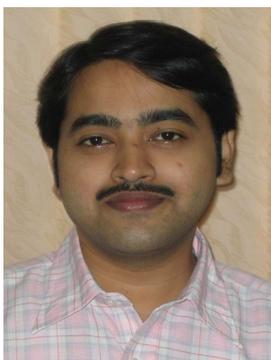

Supratim Gupta was born in Berhampore, Murshidabad, West Bengal, India, on 24[th] December, 1979. After finishing his schooling in 1998, he obtained Bachelor of Electrical Engineering (B.E.E.) with Honours from Jadavpur University, Kolkata, India, in 2002. He obtained Masters of Electrical Engineering (M.E.E.) with specialization in Measurement and Instrumentation from Jadavpur University, Kolkata, India, in 2004. He served as an R&D Officer in Megatherm Electronics Pvt. Ltd., Kolkata, India for two years just after his M.E.E. Since July, 2006 to June, 2012 he was an Institute Research Scholar in the Department of Electrical Engineering, Indian Institute of Technology, Kharagpur[1], India. Presently he is serving in the Department of Electrical Engineering, NIT, Rourkela, India as an Assistant Professor. His current area of research includes Image Processing, Computer Vision, and Embedded System Design and Progamming. His detail contact address is given below:

**Permanent Address:**
F-8, New Garia Co-operative Hsg. Dev. Soc.
P.O.: Panchasayar
Kolkata-700094
West Bengal
India.

**Present Address:**
Department of Electrical Engineering
National Institute of Technology,
Rourkela
Odisha-769008.

**E-mail Address:**    guptasu@nitrkl.ac.in
                      sgupta.iitkgp@gmail.com
**Mobile Number:**    +91-9439740117

---

[1] http://www.iitkgp.ac.in/